\documentclass{article}

\usepackage{amsmath}
\usepackage{amsfonts}
\usepackage{amssymb}
\usepackage{commath}
\usepackage[ansinew]{inputenc}
\usepackage{graphicx,color}
\usepackage{url}
\usepackage{latexsym,enumerate}
\usepackage{booktabs}
\usepackage{framed}
\usepackage{subfigure}
\usepackage{multirow} 
\usepackage{color}
\usepackage{colortbl}

\usepackage{algorithm}
\usepackage{algorithmic}

\definecolor{gray1}{rgb}{0.8,0.8,0.8}
\definecolor{gray2}{rgb}{0.95,0.95,0.95}

\newcommand{\RE}{\,{\rm Re}}

\newcommand{\argmin}{\mathop{\rm argmin}}

\newcommand{\Shrink}{\mathop{\rm Shrink}}

\newcommand{\CST}{\mathop{\rm CST}}

\newcommand{\cC}{{\mathcal C}}

\newcommand{\cI}{{\mathcal I}}
\newcommand{\Bf}{{\mathbf f}}
\newcommand{\Bu}{{\mathbf u}}

\newcommand{\Br}{{\mathbf r}}
\newcommand{\Bv}{{\mathbf v}}

\newcommand{\Bx}{{\mathbf x}}

\newcommand{\Bw}{{\mathbf w}}

\newcommand{\Bg}{{\mathbf g}}

\newcommand{\Bt}{{\mathbf t}}

\newcommand{\Beps}{{\boldsymbol \epsilon}}

\newcommand{\Bome}{{\boldsymbol \omega}}
\newcommand{\Bomeo}{{\boldsymbol{\omega_1}}}
\newcommand{\Bomet}{{\boldsymbol{\omega_2}}}

\newcommand{\Bk}{{\mathbf k}}

\newcommand{\BDm}{{\mathbf{D_m}}}
\newcommand{\BDmT}{{\mathbf{D}_{\mathbf m}^{\text T}}}
\newcommand{\BDn}{{\mathbf{D_n}}}
\newcommand{\BDnT}{{\mathbf{D}_{\mathbf n}^{\text T}}}

\newcommand{\Bz}{{\mathbf z}}
\newcommand{\Bzo}{{\mathbf{z_1}}}
\newcommand{\Bzt}{{\mathbf{z_2}}}
\newcommand{\BzoI}{{\mathbf z}_{\mathbf 1}^{-1}}
\newcommand{\BztI}{{\mathbf z}_{\mathbf 2}^{-1}}

\begin{document}

\title{Directional Global Three-part Image Decomposition}

\author{D.H. Thai\thanks{Statistical and Applied Mathematical Science Institute (SAMSI), USA.
Email: dhthai@samsi.info}
\ and C. Gottschlich\thanks{Institute for Mathematical Stochastics, 
University of Goettingen,
Goldschmidtstr. 7, 37077 G\"ottingen, 
%Goettingen, Germany.
Email:~gottschlich@math.uni-goettingen.de}}

\maketitle

\begin{abstract}
We consider the task of image decomposition and we introduce a new model coined directional global three-part decomposition (DG3PD) for solving it. As key ingredients of the DG3PD model, we introduce a discrete multi-directional total variation norm and a discrete multi-directional G-norm. Using these novel norms, the proposed discrete DG3PD model can decompose an image into two parts or into three parts. Existing models for image decomposition by Vese and Osher, by Aujol and Chambolle, by Starck et al., and by Thai and Gottschlich are included as special cases in the new model. Decomposition of an image by DG3PD results in a cartoon image, a texture image and a residual image. Advantages of the DG3PD model over existing ones lie in the properties enforced on the cartoon and texture images. The geometric objects in the cartoon image have a very smooth surface and sharp edges. The texture image yields oscillating patterns on a defined scale which is both smooth and sparse. Moreover, the DG3PD method achieves the goal of perfect reconstruction by summation of all components better than the other considered methods. Relevant applications of DG3PD are a novel way of image compression as well as feature extraction for applications such as latent fingerprint processing and optical character recognition.
\end{abstract}

\section*{Keywords}

Image decomposition, variational calculus, cartoon image, texture image, image compression, feature extraction,
latent fingerprint image processing, optical character recognition, fingerprint recognition.

\section{Introduction} \label{sec:Introduction}

\begin{figure*}
\begin{center}
  \subfigure[{\tiny Simulated}]{\includegraphics[width=0.15\textwidth]{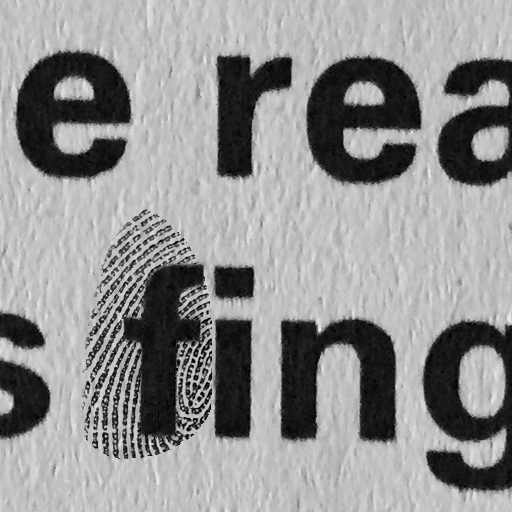}}
	\subfigure[{\tiny Document}]{\includegraphics[width=0.15\textwidth]{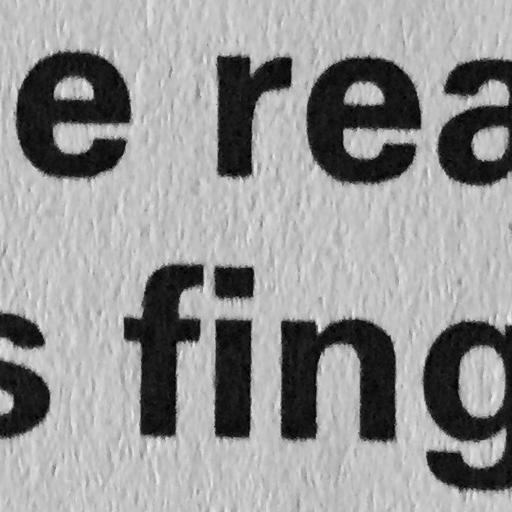}}
	\subfigure[{\tiny Fingerprint}]{\includegraphics[width=0.15\textwidth]{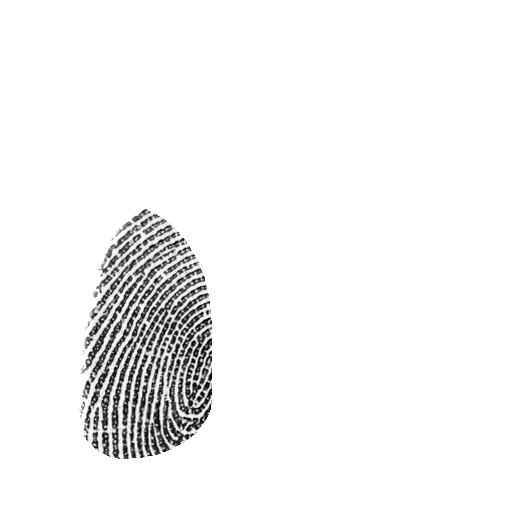}} 
	\subfigure[$\Bu$]{\includegraphics[width=0.15\textwidth]{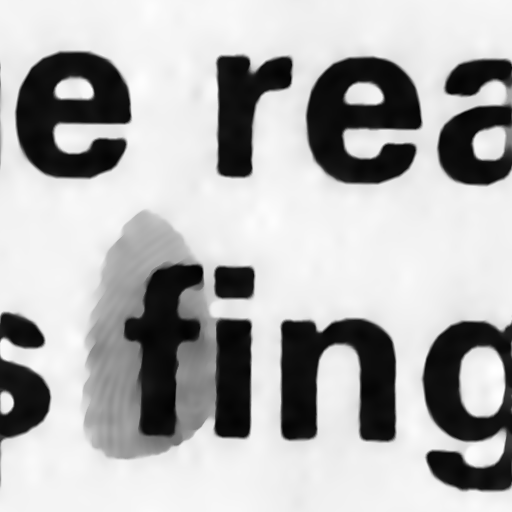}}
	\subfigure[$\Bv$]{\includegraphics[width=0.15\textwidth]{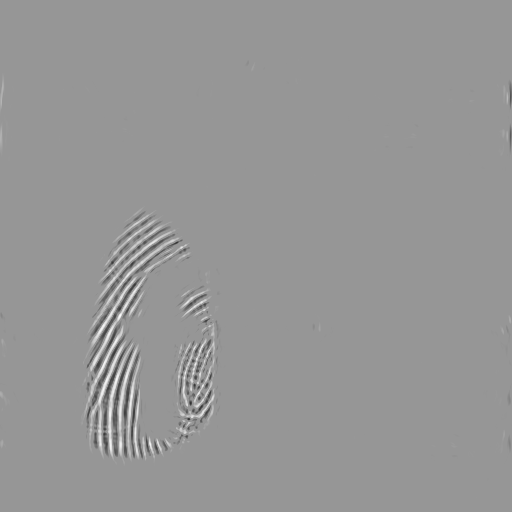}}
	\subfigure[$\Bv_\text{bin}$]{\includegraphics[width=0.15\textwidth]{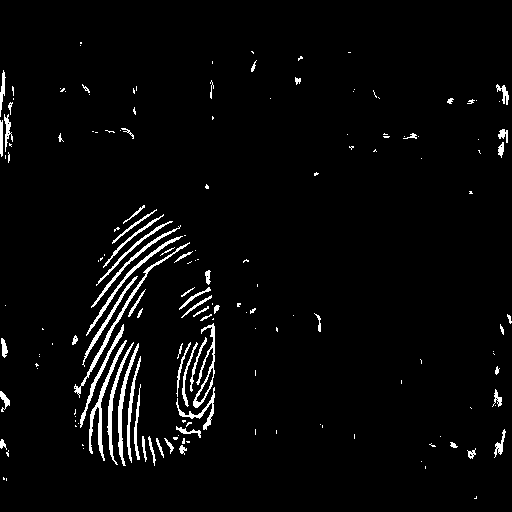}} \\
	
	\subfigure[{\tiny ROI}]{\includegraphics[width=0.15\textwidth]{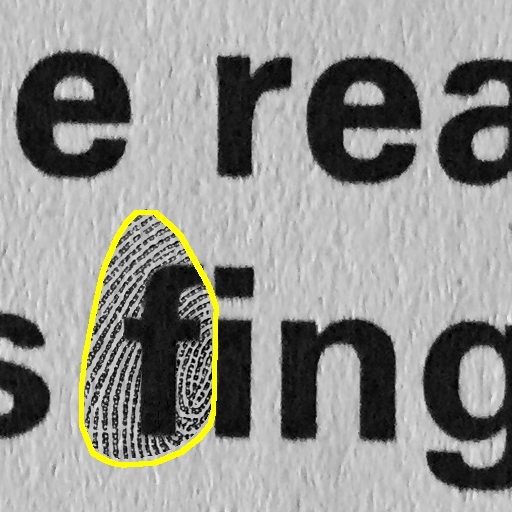}}
	\subfigure[{\tiny Brodatz D15}]{\includegraphics[width=0.15\textwidth]{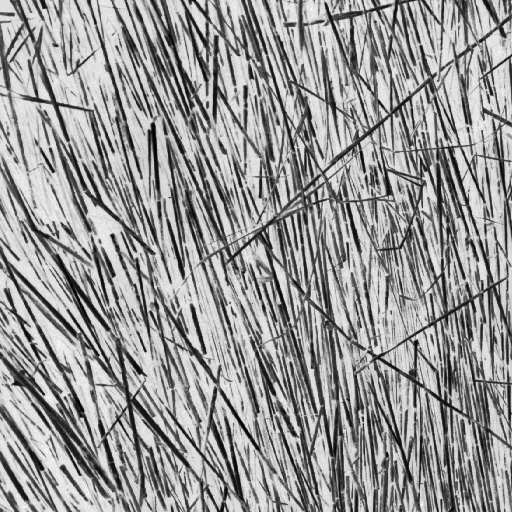}}
	\subfigure[{\tiny Simulated}]{\includegraphics[width=0.15\textwidth]{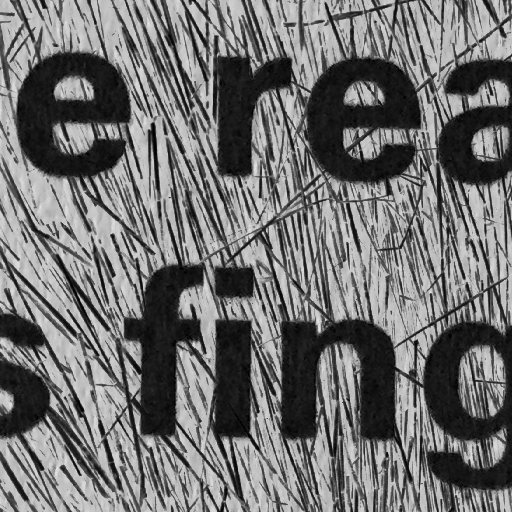}}
	\subfigure[$\Bu$]{\includegraphics[width=0.15\textwidth]{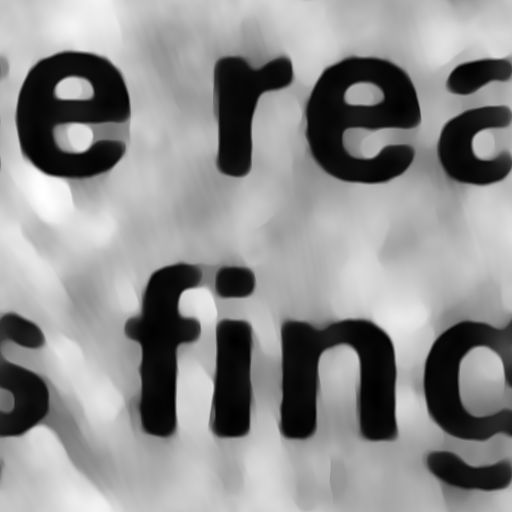}}
	\subfigure[$\Bv$]{\includegraphics[width=0.15\textwidth]{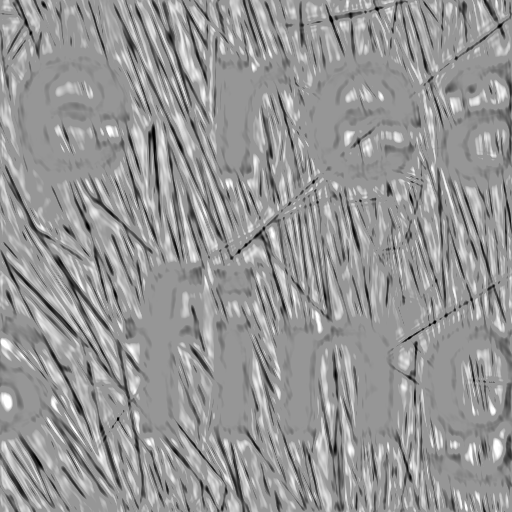}}
	\subfigure[$\Beps$]{\includegraphics[width=0.15\textwidth]{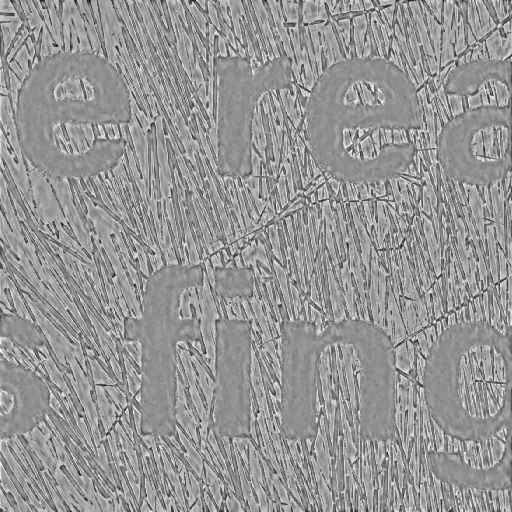}}
	
\end{center}
\caption{The simulated latent fingerprint (a) is composed by adding the fingerprint (c) 
         to image (b) which is a detail from a photo of a printed document.
				 DG3PD decomposition of (a) obtains a smooth cartoon image $\Bu$ (d) 
				 and a smooth and sparse texture image $\Bv$ (e). 
				 All positive coefficients of $\Bv$ are visualized as white pixels in (f).
				 The region-of-interest shown in (g) is estimated from $\Bv_\text{bin}$ using morphological operations \cite{ThaiHuckemannGottschlich2015}.
				 Image (i) is composed from (b) and (h) and decomposed by DG3PD 
				 into cartoon (j), texture (k) and residual (l).
				 }
\label{fig:exampleSimulatedLatent}
\end{figure*}

Feature extraction, denoising and image compression are key issues in computer vision and image processing.
We address these main tasks based on the paradigm
that an image can be regarded as the addition or montage of several meaningful components.
Image decomposition methods attempt to model these components by their properties
and to recover the individual components using an algorithm. 
Relevant component images include geometrical objects which have piece-wise constant values or a smooth surface
like e.g. the characters in Figure \ref{fig:exampleSimulatedLatent} (b)
or components which are filled with an oscillating pattern like e.g. the fingerprint in \ref{fig:exampleSimulatedLatent} (c).

Based on these observations, we define the following goals:

\textbf{Goal 1: the cartoon component $\mathbf u$ contains only geometrical objects with a very smooth surface, sharp boundaries and no texture.}

\textbf{Goal 2: the texture component $\mathbf v$ contains only geometrical objects with oscillating patterns 
and $\mathbf v$ shall be both smooth and sparse.}

\textbf{Goal 3: three-part decomposition and reconstruction $\mathbf f = \mathbf u + \mathbf v + \boldsymbol \epsilon$.}

How does achieving these goals serve the tasks of feature extraction,
denoising and compression?

Extremely efficient representations of the cartoon image $\mathbf u$ and texture image $\mathbf v$ exist.
These two component images are highly compressible as discussed with full details in Section \ref{sec:compression}.
Depending on the application, $\mathbf u$ or $\mathbf v$, or both can be considered as feature images.
For the application to latent fingerprints, we are especially interested in the texture image $\mathbf v$ 
as a feature for fingerprint segmentation and all subsequent processing steps.
Example results for the very challenging task of latent fingerprint segmentation 
are given in Section \ref{sec:featextract}. 
In optical character recognition (OCR), pre-processing includes the removal of 
complex background and the isolation of characters.
After three-part decomposition and depending on the scale of the characters,
the cartoon image $\mathbf u$ contains the information of interest for OCR (see Figure \ref{fig:exampleSimulatedLatent} (j)),
and the background is fractionized into $\mathbf v$ and $\boldsymbol \epsilon$
simultaneously in the minimization procedure.
As a consequence from the requirements imposed on $\mathbf u$ and $\mathbf v$, 
noise and small scale objects are driven into the residual image $\boldsymbol \epsilon$ 
during the dismantling of $\mathbf f$.
Therefore, the image $\mathbf u + \mathbf v$ can be regarded as a denoised version of $\mathbf f$
and the degree of denoising can be steered by the choice of parameters.

The paper is organized as follows.
In Section \ref{sec:notation}, we begin by describing notation and preliminaries.
After having established these prerequisites, we define the DG3PD model in Section \ref{sec:DG3PD}
and in Section \ref{sec:relatedWork},
we explain its relation to existing models in the literature for two-part and three-part decomposition.
In Section \ref{sec:solutionDG3PD}, we describe an iterative, numerical algorithm
which solves the DG3PD model for practical applications to discrete, 2-dimensional images.
In Section \ref{sec:comparisonPriorArt}, we perform a detailed comparison of the DG3PD method
to state-of-the-art decomposition approaches.
Applications of DG3PD, especially feature extraction and image compression, are the topic of Section \ref{sec:applications}.
Discussion and conclusion are given in Section~\ref{sec:conclusion}.
An overview over the algorithm and an additional comparison is given in the Appendix.

\section{Notation and Preliminaries} \label{sec:notation}

For simplification, we use a bold symbol to denote the coordinates of a two-dimensional signal, e.g. 
$\Bx = (x_1 \,, x_2) \,, \Bk = [k_1 \,, k_2] \,, \Bome = (\omega_1 \,, \omega_2)$ and $e^{j \Bome} = [ e^{j \omega_1} \,, e^{j \omega_2} ]$.
A two-dimensional image $f[\Bk] : \Omega \rightarrow \mathbb R_+$, the discretization of the continuous version $f(\Bx)$
(i.e. $f[\Bk] = f(\Bx) \mid_{\Bx = \Bk \in \Omega}$), is specified on the lattice:
\begin{equation*}
 \Omega = \Big\{ (k_1 \,, k_2) \in \{ 1 \,, m \} \times \{ 1 \,, n \} \subset \mathbb N^2 \Big\}.
\end{equation*}

Let $X$ be the Euclidean space whose dimension is given by the size of the lattice $\Omega$, 
i.e. $X = \mathbb R^{\abs{\Omega}}$. 
The 2D discrete Fourier transform $\mathcal F$ acting on $f[\Bk]$ is
\begin{equation*}
 f[\Bk] ~\stackrel{\mathcal F}{\longleftrightarrow}~ F(e^{j \Bome}) 
 = \sum_{\Bk \in \Omega} f[\Bk] \cdot e^{-j \langle \Bk \,, \Bome \rangle_{\ell_2}},
\end{equation*}
where $\Bome$ is defined on the lattice:
\begin{equation*}
 \cI = \Big\{ (\omega_1 \,, \omega_2) = \Big( \frac{2 \pi n'}{n} \,, \frac{2 \pi m'}{m} \Big) 
 \mid (n' \,, m') = \Big[ - \frac{n}{2} \,, \frac{n}{2} \Big] \times \Big[ - \frac{m}{2} \,, \frac{m}{2} \Big] \subset \mathbb Z^2
 \Big\} \,,
\end{equation*}
i.e. $\Bome \in [-\pi \,, \pi]^2.$

\paragraph{Forward and backward difference operators:}
Given the matrix
\begin{equation*}
 \BDm = \begin{pmatrix}
         -1            & 1             &0            &\hdots           &0              \\
          0            &-1             &1            &\hdots           &0              \\
          \vdots       &\vdots         &\vdots       &\ddots           &\vdots         \\
          0            &0              &0            &\hdots           &1              \\
          1            &0              &0            &\hdots            &-1             \\             
        \end{pmatrix}
  \in \mathbb R^{m \times m},
\end{equation*}
the forward and backward difference operators with periodic boundary condition in convolution and matrix forms and their Fourier transform
are explained in Table~\ref{tab:ForwardBackwardDifference}.
   
\paragraph{Discrete directional derivative:}

{\renewcommand{\arraystretch}{1.6} 
\begin{table}
\begin{center}
  \begin{tabular}{|c|c|c|}
    \hline
                  & Forward difference                                                      & Backward difference		
    \\
    \hline
    Convolution   & $\partial^{+}_{x} f[\Bk] = f[k_1, k_2+1] - f [k_1, k_2]$                & $\partial^{-}_{x} f[\Bk] = f[k_1, k_2] - f [k_1, k_2-1]$
    \\
    form          & $\partial^{+}_{y} f[\Bk] = f[k_1+1, k_2] - f [k_1, k_2]$                & $\partial^{-}_{y} f[\Bk] = f[k_1, k_2] - f [k_1-1, k_2]$
    \\
    \hline
    Matrix        & $\Big[ \partial^{+}_{x} f[\Bk] \Big]_{\Bk \in \Omega} = \Bf \, \BDnT$   & $\Big[ \partial^{-}_{x} f[\Bk] \Big]_{\Bk \in \Omega} = - \Bf \, \BDn$		
    \\
    form          & $\Big[ \partial^{+}_{y} f[\Bk] \Big]_{\Bk \in \Omega} = \BDm \Bf$       & $\Big[ \partial^{-}_{y} f[\Bk] \Big]_{\Bk \in \Omega} = - \BDmT \Bf$
    \\
    \hline
    Fourier       & $\big( e^{j \omega_2} - 1 \big) F \big( e^{j \Bome} \big)$              & $- \big( e^{-j\omega_2} - 1 \big) F \big( e^{j \Bome} \big)$
    \\ 
    transform     & $\big( e^{j\omega_1} - 1 \big) F \big( e^{j \Bome} \big)$               & $- \big( e^{-j \omega_1} - 1 \big) F \big( e^{j \Bome} \big)$
    \\
    \hline
  \end{tabular} 
  \vspace*{8pt}
  \caption{Forward/backward difference with periodic boundary condition in convolution/matrix form and their Fourier transform.
	         $\BDnT$ and $\BDmT$ are the transposed matrices of $\BDn$ and $\BDm$, respectively.
           \label{tab:ForwardBackwardDifference}}
\end{center}
\end{table}  
}

Let $\nabla^+ = \big[ \partial_x^+ \,, \partial_y^+ \big]$ be the discrete forward gradient operator with 
$\partial_x^+$ and $\partial_y^+$ are defined in Table \ref{tab:ForwardBackwardDifference}.
The discrete derivative operator following the direction $\overrightarrow{d} = \Big[ \cos \frac{\pi l}{L} \,, \sin \frac{\pi l}{L} \Big]^\text{T}$ 
with $l = 0 \,, \ldots \,, L-1$ is defined as   
\begin{align*}
 \partial_l^+ &= \Big \langle \overrightarrow{d} \,, \nabla^+ \Big \rangle 
 = \cos (\frac{\pi l}{L}) \partial_x^+ ~+~ \sin (\frac{\pi l}{L}) \partial_y^+.
\end{align*}
Thus, the discrete directional gradient operator is
\begin{equation} \label{eq:disdirGrad}
 \nabla_L^+ = \Big[ \partial_l^+ \Big]_{l \in [0, L-1]}.
\end{equation}

\subsection{Discrete directional TV-norm} \label{sec:DTVnorm}

The continuous total variation norm has been defined in \cite{RudinOsherFatemi1992}.
Due to the discrete nature of images, we define its discrete version with forward difference operators as 
\begin{equation*}
 \norm{\nabla^+ \Bu}_{\ell_1} = \sum_{\Bk \in \Omega} \sqrt{ \big(\partial_x^+ u[\Bk]\big)^2 + \big(\partial_y^+ u[\Bk]\big)^2 }.
\end{equation*}

We extend it into multi-direction $L$ with the discrete directional gradient operator (\ref{eq:disdirGrad}):
\begin{equation*}
 \norm{\nabla_L^+ \Bu}_{\ell_1} = \sum_{\Bk \in \Omega} \sqrt{\sum_{l=0}^{L-1} (\partial_l^+ u[\Bk])^2}
 = \norm{ \sqrt{\sum_{l=0}^{L-1} \big(\cos \frac{\pi l}{L} \Bu \BDnT + \sin \frac{\pi l}{L} \BDm \Bu \big)^2} }_{\ell_1}.
\end{equation*}
The discrete anisotropic total variation norm in a matrix form is
\begin{equation} \label{eq:disdirTVnorm}
 \norm{\nabla_L^+ \Bu}_{\ell_1} = \sum_{l=0}^{L-1} \norm{ \cos \frac{\pi l}{L} \Bu \BDnT + \sin \frac{\pi l}{L} \BDm \Bu }_{\ell_1}.
\end{equation}

\subsection{Discrete directional G-norm} \label{sec:DGnorm}

\paragraph{Discrete G-norm:}

Meyer \cite{Meyer2001} has proposed a space $G$ of continuous functions to measure oscillating functions (texture and noise).
The discrete version of the G-norm has been introduced by Aujol and Chambolle in \cite{AujolChambolle2005}. 
We rewrite it with the matrix form of the forward difference operators as
\begin{equation}  \label{eq:disGnorm}
 \norm{\Bv}_G = \inf \Bigg\{ \norm{ \sqrt{ \Bg_1^2 + \Bg_2^2 } }_{\ell_\infty} \,,~ 
 \Bv = \Bg_1 \BDnT + \BDm \Bg_2 
 \,,~ \Big[ \Bg_l \Big]_{l \in [1,2]} \in X^2 
 \Bigg\}.
\end{equation}

\paragraph{Discrete Directional G-norm:} 

We extend (\ref{eq:disGnorm}) into multi-directions $S \in \mathbb N_+$ with the directional difference operator to obtain 
the discrete directional G-norm as
\begin{equation}  \label{eq:disdirGnorm}
 \norm{\Bv}_{G_S} = \inf \Bigg\{ \norm{ \sqrt{ \sum_{s=0}^{S-1} \Bg_s^2 } }_{\ell_\infty} \,,~ 
 \underbrace{ \Bv = \sum_{s=0}^{S-1} \Big[ \cos \frac{\pi s}{S} \Bg_s \BDnT + \sin \frac{\pi s}{S} \BDm \Bg_s \Big]
 }_{\displaystyle \Leftrightarrow~ v[\Bk] = \sum_{s=0}^{S-1} \partial_s^+ g_s [\Bk] \,, \forall \Bk \in \Omega }
 \,,~ \Big[ \Bg_s \Big]_{s = 0}^{S-1} \in X^S 
 \Bigg\}.
\end{equation}

\section{DG3PD} \label{sec:DG3PD}

We define the DG3PD model 
for discrete directional three-part decomposition of an image into cartoon, texture and residual parts as
 \begin{equation} \label{eq:DTV-DG3part}
  \min_{(\Bu\,,\Bv\,,\Beps) \in X^3}
  \Big\{ \norm{\nabla_L^+ \Bu}_{\ell_1} + \mu_1 \norm{\Bv}_{G_S} + \mu_2 \norm{\Bv}_{\ell_1}
  \text{ s.t. } 
  \sup_{(i, l, k) \in \mathcal K} \abs{\mathcal C_{i,l} \big\{ \Beps \big\} [\Bk]} \leq \delta
  \,,~ \Bf = \Bu + \Bv + \Beps \Big\},
 \end{equation}
where $\mathcal C$ is the curvelet transform \cite{CandesDemanetDonohoYing2006,MaPlonka2010} with the index set $\mathcal K$.
Please note that by setting the parameter $\delta = 0$ in (\ref{eq:DTV-DG3part}),
we obtain a two-part decomposition 
which can be considered as a special case of the DG3PD model.
Next, we discuss how the DG3PD model relates to existing decomposition models.

\section{Related Work} \label{sec:relatedWork}

In this section, we give an overview over related work in 
two-part and three-part image decomposition in chronological order.

\subsection{Mumford and Shah}

In 1989, Mumford and Shah \cite{MumfordShah1989} have proposed piece-wise smooth (for image restoration)
and piece-wise constant (for image segmentation) models by minimizing the energy functional.
This approach can be considered as a precursor for subsequent image decomposition models with texture components.
However, due to the Hausdorff 1-dimensional measure $\mathcal H^1$ in $\mathbb R^2$,
it poses a challenging or even NP-hard problem in optimization 
to minimize the Mumford and Shah functional.
Later, based on the Mumford and Shah model, 
Chan and Vese \cite{ChanVese2001} have proposed an active contour for image segmentation 
which is solved by a level set method \cite{OsherSethian1988}.

\subsection{Rudin, Osher, and Fatemi}

In 1992, Rudin, Osher, and Fatemi \cite{RudinOsherFatemi1992} 
pioneered image decomposition with a two-part model for denoising.

\subsection{Meyer}

The model defined by Meyer in 2001 \cite{Meyer2001} for two-part decomposition in the continuous setting
is comprised in the DG3PD model for the special case of $L = S = 2$ and $\mu_2 = \delta = 0$
in the discrete domain.

\subsection{Vese and Osher}

In 2003,
Vese and Osher \cite{VeseOsher2003} solved Meyer's model
for two-part decomposition in the continuous setting
and
they proposed to approximate the $L_\infty$-norm in the G-norm by the $L_1$-norm
and to apply the penalty method for reformulating the constraint.
For practical application to images, they discretized their solution.
This approach is extended in \cite{VeseOsher2004}.

\subsection{Aujol and Chambolle}

Aujol and Chambolle in 2005 \cite{AujolChambolle2005} 
adapted the work by Meyer for discrete two-part decomposition (see (5.49) in \cite{AujolChambolle2005})
and they used the penalty method for the constraint.
Their model is included in the DG3PD model with parameters $L = S = 2$, $\mu_1 = \mu_2 = 0$
and applying the supremum norm to the wavelet coefficients of the oscillating pattern, i.e. $\norm{\mathcal W \{ \cdot \}}_{\ell_\infty}$,
instead of the curvelet coefficients $\norm{\mathcal C \{ \cdot \}}_{\ell_\infty}$ as in the DG3PD model.

Moreover, Aujol and Chambolle 
proposed a model for discrete three-part decomposition (see (6.59) in \cite{AujolChambolle2005})
which measures texture by the G-norm and noise by the supremum norm of wavelet coefficients,
and the penalty method for the constraint.
Different from Vese and Osher as well as our approach explained later, 
they describe the G-norm for capturing texture using the indicator function defined on a convex set 
and they obtain the solution by Chambolle's projection onto this convex set.
Their model is included in the DG3PD model for parameters $L = S = 2$, $\mu_2 = 0$
and using wavelets instead of curvelets for the residual as before.

\subsection{Starck et al.}

Starck \textit{et al.} \cite{StarckEladDonoho2005} introduced a model for two-part decomposition
based on a dictionary approach. 
Their basic idea is to choose one appropriate dictionary for piecewise smooth objects (cartoon)
and another suitable dictionary for capturing texture parts.

\subsection{Aujol et al.}

In 2006, Aujol \textit{et al.} \cite{AujolGilboaChanOsher2006}
proposed a two-part decomposition of an image into a structure component and a texture component
using Gabor functions for the texture part.

\subsection{Gilles}

Gilles \cite{Gilles2007} has proposed a three-part image decomposition in 2007 which is similar to the Aujol-Chambolle model \cite{AujolChambolle2005},
but  G-norm is used as a measurement of the residual (or noise) instead of Besov space $\dot{B}^\infty_{-1, \infty}$
with a local adaptability property. Their argument is that the more a function is oscillatory, the smaller is the G norm.
Then, they propose a new ``merged-algorithm'' with a combination of a local adaptivity behavior and Besov space.

\subsection{Maragos and Evangelopoulos}

In 2007, Maragos and Evangelopoulos \cite{MaragosEvangelopoulos2007} have proposed 
a two-part decomposition model which relies on energy responses of a bank of 2D Gabor filters
for measuring the texture component. 
They discuss the connection between Meyer's oscillating functions \cite{Meyer2001}, Gabor filters \cite{Gottschlich2012}
and AM-FM image modeling \cite{HavlicekHardingBovik2000,LarkinFletcher2007}.

\subsection{Buades et al.}

In 2010, Buades \textit{et al.} \cite{BuadesLeMorelVese2010} derived a nonlinear filter pair 
for two-part decomposition into cartoon and texture parts.
Further models for two-part decomposition are listed 
in Table 1 of \cite{BuadesLeMorelVese2010}.

\subsection{Maurel et al. and Chikkerur et al.}

In 2011, Maurel \textit{et al.} \cite{MaurelAujolPeyre2011}
proposed a decomposition approach which models the texture component
by local Fourier atoms. 
For fingerprint textures, Chikkerur \textit{et al.} \cite{ChikkerurCartwrightGovindaraju2007}
proposed in 2007 the application of local Fourier analysis (or short time Fourier transform, STFT)
for image enhancement. 
However, the usefulness of local Fourier analysis for capturing texture information depends on and is limited 
by the level of noise in the corresponding local window, see Figure 2c in \cite{GottschlichSchoenlieb2012}
for an example in which STFT enhances some regions successfully and fails in other regions.

\subsection{G3PD}

A model for discrete three-part decomposition of fingerprint images has recently been proposed by
Thai and Gottschlich in 2015 \cite{ThaiGottschlich2015G3PD} with the aim of obtaining 
a texture image $\mathbf v$ which serves as a useful feature for estimating the region-of-interest (ROI).
The G3PD model is included in the DG3PD model by choosing $L = 2$ 
and replacing the directional G-norm in the DG3PD model by the $\ell_1$-norm of curvelet coefficients 
(multi-scale and multi-orientation decomposition) to capture texture.
However, a disadvantage of the $\ell_1$-norm of curvelet coefficients
is a tendency to generate the halo effect on the boundary of the texture region 
due to the scaling factor in curvelet decomposition 
(see Figure 3(d) in \cite{ThaiGottschlich2015G3PD}),
whereas the directional G-norm in the DG3PD model is capable to capture oscillating patterns (cf. \cite{Meyer2001})
without the halo effect.

\subsection{Directional Total Variation and G-Norm}

In Section \ref{sec:DTVnorm}, we introduced the discrete directional total variation norm
and in Section \ref{sec:DGnorm}, we introduced the discrete directional G-norm. 
Please note the aspect of summation over multiple directions in Equations (\ref{eq:disdirTVnorm}) and (\ref{eq:disdirGnorm}).

The term 'directional total variation'
has previously been used by Bayram and Kamasak \cite{BayramKamasak2012EUSIPCO,BayramKamasak2012}
for defining and computing the TV-norm in only one specific direction. 
They have treated the special case of images with one globally dominant direction 
and addressed those by two-part decomposition and for the purpose denoising. 
Zhang and Wang \cite{ZhangWang2013} proposed an extension of the work by Bayram and Kamasak 
for denoising images with more than one dominant direction.

\section{Solution of the DG3PD Model} \label{sec:solutionDG3PD}

Now, we present a numerical algorithm
for obtaining the solution of the DG3PD model stated in (\ref{eq:DTV-DG3part}).
Given $\delta > 0$, denote $G^*\big( \frac{\Beps}{\delta} \big)$ as the indicator function on 
 the feasible convex set $A(\delta)$ of (\ref{eq:DTV-DG3part}), i.e.
 \begin{equation*}
  A(\delta) ~=~ \Big\{ \Beps \in X \,:~ \norm{C \big\{ \Beps \big\}}_{\ell_\infty} \leq \delta \Big\} 
  \text{  and  }
  G^*\big( \frac{\Beps}{\delta} \big) ~=~ \begin{cases} 0 \,, &\Beps \in A(\delta)  \\  \pm\infty \,, &\Beps \notin A(\delta)
                              \end{cases}.
 \end{equation*}
 
By analogy with the work of Vese and Osher, we consider the approximation of G-norm with
 $\ell_1$ norm and the anisotropic version of directional total variation norm.
 The minimization problem in (\ref{eq:DTV-DG3part}) is rewritten as
 \begin{align} \label{eq:DTV-DG3part_rewrite}
  &\big( \Bu^* \,, \Bv^* \,, \Beps^* \,, \big[ \mathbf{g}_s^* \big]_{s=0}^{S-1} \big) 
  = \argmin_{\big( \Bu \,, \Bv \,, \Beps \,, \big[ \mathbf{g}_s \big]_{s=0}^{S-1} \big) \in X^{S+3} }
  \Bigg\{ \sum_{l=0}^{L-1} \norm{ \cos\big( \frac{\pi l}{L} \big) \Bu \BDnT + \sin\big( \frac{\pi l}{L} \big) \BDm \Bu }_{\ell_1}     \notag
  \\&
  + \mu_1 \sum_{s=0}^{S-1} \norm{\mathbf{g}_s}_{\ell_1} + \mu_2 \norm{\Bv}_{\ell_1} + G^* \big( \frac{\Beps}{\delta} \big)
  ~\text{s.t.}~ 
  \begin{cases}
   \Bf = \Bu + \Bv + \Beps  \\
   \Bv = \displaystyle \sum_{s=0}^{S-1} \Big[ \cos\big( \frac{\pi s}{S} \big) \Bg_s \BDnT + \sin\big( \frac{\pi s}{S} \big) \BDm \Bg_s \Big]   \\ 
  \end{cases}
  \Bigg\} .
 \end{align}

To simplify the calculation, we introduce two new variables:
\begin{equation*}
 \begin{cases}
  \Br_b = \cos\big( \frac{\pi b}{L} \big) \Bu \BDnT + \sin\big( \frac{\pi b}{L} \big) \BDm \Bu \,,~ b = 0 \,, \ldots \,, L-1 \,, \\
  \Bw_a = \Bg_a \,, a = 0 \,, \ldots \,, S-1 \,.
 \end{cases}
\end{equation*}
Equation (\ref{eq:DTV-DG3part_rewrite}) is a constrained minimization problem. 
The augmented Lagrangian method (ALM) is applied to turn 
(\ref{eq:DTV-DG3part_rewrite}) into an unconstrained one as
\begin{equation} \label{eq:DTV-DG3part_ALM} 
 \min_{\big( \Bu \,, \Bv \,, \Beps \,, \big[ \Br_l \big]_{l=0}^{L-1} \,,
 \big[ \Bw_s \big]_{s=0}^{S-1} \,, \big[ \Bg_s \big]_{s=0}^{S-1} \big) \in X^{L+2S+3} }
 \mathcal L \Big( \Bu \,, \Bv \,, \Beps \,, \big[ \Br_l \big]_{l=0}^{L-1} \,,
 \big[ \Bw_s \big]_{s=0}^{S-1} \,, \big[ \Bg_s \big]_{s=0}^{S-1} \;; \big[ \boldsymbol{\lambda}_{\boldsymbol{1} l} \big]_{l=0}^{L-1} \,,
 \big[ \boldsymbol{\lambda}_{\boldsymbol{2}s} \big]_{s=0}^{S-1} \,, \boldsymbol{\lambda_3} \,, \boldsymbol{\lambda_4} \Big) \,,
\end{equation}
where the Lagrange function is
\begin{align*}
 &\mathcal L ( \cdot \;; \cdot ) ~=~
 \sum_{l=0}^{L-1} \norm{ \Br_l }_{\ell_1} 
 + \mu_1 \sum_{s=0}^{S-1} \norm{\Bw_s}_{\ell_1} + \mu_2 \norm{\Bv}_{\ell_1} + G^* \big( \frac{\Beps}{\delta} \big)
 \\&
 + \frac{\beta_1}{2} \sum_{l=0}^{L-1} \norm{ \Br_l - \cos\big( \frac{\pi l}{L} \big) \Bu \BDnT - \sin\big( \frac{\pi l}{L} \big) \BDm \Bu 
 + \frac{\boldsymbol{\lambda}_{\boldsymbol{1} l}}{\beta_1} }^2_{\ell_2}
 + \frac{\beta_2}{2} \sum_{s=0}^{S-1} \norm{\Bw_s - \Bg_s + \frac{\boldsymbol{\lambda}_{\boldsymbol{2}s}}{\beta_2}}^2_{\ell_2}
 \\&
 + \frac{\beta_3}{2} \norm{ \Bv - \sum_{s=0}^{S-1} \Big[ \cos\big( \frac{\pi s}{S} \big) \Bg_s \BDnT 
 + \sin\big( \frac{\pi s}{S} \big) \BDm \Bg_s \Big] + \frac{\boldsymbol{\lambda_3}}{\beta_3} }^2_{\ell_2}
 + \frac{\beta_4}{2} \norm{ \Bf - \Bu - \Bv - \Beps + \frac{\boldsymbol{\lambda_4}}{\beta_4} }_{\ell_2}^2 .
\end{align*}

Due to the minimization problem with multi-variables, we apply  
the alternating directional method of multipliers to solve (\ref{eq:DTV-DG3part_ALM}).
Its minimizer is numerically computed through iterations $t = 1 \,, 2 \,, \ldots$
\begin{align} \label{eq:DTV-DG3part_ALM_numerical} 
 &{\Big( \Bu^{(t)} \,, \Bv^{(t)} \,, \Beps^{(t)} \,, \big[ \Br_l^{(t)} \big]_{l=0}^{L-1} \,,
 \big[ \Bw_s^{(t)} \big]_{s=0}^{S-1} \,, \big[ \Bg_s^{(t)} \big]_{s=0}^{S-1} \Big) }                     \notag
 ~=~
 \\
 &\argmin ~ \mathcal L \Big( \Bu \,, \Bv \,, \Beps \,, \big[ \Br_l \big]_{l=0}^{L-1} \,,
 \big[ \Bw_s \big]_{s=0}^{S-1} \,, \big[ \Bg_s \big]_{s=0}^{S-1} \;;~ \big[ \boldsymbol{\lambda}_{\boldsymbol{1} l}^{(t-1)} \big]_{l=0}^{L-1} \,,
 \big[ \boldsymbol{\lambda}_{\boldsymbol{2}s}^{(t-1)} \big]_{s=0}^{S-1} \,, \boldsymbol{\lambda}_{\mathbf 3}^{(t-1)} \,, \boldsymbol{\lambda}_{\mathbf 4}^{(t-1)} \Big)
\end{align}

and the Lagrange multipliers are updated after every step $t$ with a rate $\gamma$.
We initialize 
$\Bu^{(0)} = \Bf \,, \Bv^{(0)} = \Beps^{(0)} = \big[ \Br_l^{(0)} \big]_{l=0}^{L-1} = \big[ \Bw_s^{(0)} \big]_{s=0}^{S-1} = \big[ \Bg_s^{(0)} \big]_{s=0}^{S-1} 
= \big[\boldsymbol{\lambda}_{\mathbf{1}l}^{(0)}\big]_{l=0}^{L-1} = \big[\boldsymbol{\lambda}_{\mathbf{2}a}^{(0)}\big]_{a=0}^{S-1} 
= \boldsymbol{\lambda}_{\mathbf 3}^{(0)} = \boldsymbol{\lambda}_{\mathbf 4}^{(0)} =\boldsymbol 0$.
In each iteration, we first solve the following six subproblems in the listed order
and then we update the four Lagrange multipliers:

% Step 1
% r problem:
{\bfseries The ``$\big[ \Br_l \big]_{l=0}^{L-1}$-problem'':} Fix $\Bu$, $\Bv$, $\Beps$, $\big[ \Bw_s \big]_{s=0}^{S-1}$, $\big[ \Bg_s \big]_{s=0}^{S-1}$ and
\begin{equation} \label{eq:sub:r}
 \min_{ \big[ \Br_l \big]_{l=0}^{L-1} \in X^{L} }
 \Bigg\{ \sum_{l=0}^{L-1} \norm{ \Br_l }_{\ell_1} 
 + \frac{\beta_1}{2} \sum_{l=0}^{L-1} \norm{ \Br_l - \cos\big( \frac{\pi l}{L} \big) \Bu \BDnT - \sin\big( \frac{\pi l}{L} \big) \BDm \Bu 
 + \frac{\boldsymbol{\lambda}_{\boldsymbol{1} l}}{\beta_1} }^2_{\ell_2}
 \Bigg\}
\end{equation}
Due to its separability, we consider the problem at $b = 0 \,, \ldots \,, L-1$. 
The solution of (\ref{eq:sub:r}) is
\begin{equation*}
 \Br_b^* ~=~ \Shrink \Big( \cos\big( \frac{\pi b}{L} \big) \Bu \BDnT + \sin\big( \frac{\pi b}{L} \big) \BDm \Bu 
 - \frac{\boldsymbol{\lambda}_{\boldsymbol{1} b}}{\beta_1} \,, \frac{1}{\beta_1} \Big).
\end{equation*}
The operator $\Shrink(\cdot\,, \cdot)$ is defined in \cite{ThaiGottschlich2015G3PD}. 

% Step 2
% w_s problem
{\bfseries The ``$\big[ \Bw_s \big]_{s=0}^{S-1}$-problem'':} Fix $\Bu$, $\Bv$, $\Beps$, $\big[ \Br_l \big]_{l=0}^{L-1}$, $\big[ \Bg_s \big]_{s=0}^{S-1}$ and
\begin{equation}  \label{eq:sub:w}
 \min_{ \big[ \Bw_s \big]_{s=0}^{S-1} \in X^{S} }
 \Bigg\{  
 \mu_1 \sum_{s=0}^{S-1} \norm{\Bw_s}_{\ell_1} 
 + \frac{\beta_2}{2} \sum_{s=0}^{S-1} \norm{\Bw_s - \Bg_s + \frac{\boldsymbol{\lambda}_{\boldsymbol{2}s}}{\beta_2}}^2_{\ell_2}
 \Bigg\}
\end{equation}

Similarly, the solution of (\ref{eq:sub:w}) for each separable problem $a = 0 \,, \ldots \,, S-1$ is
\begin{equation} \label{eq:sub:solution:wa}
 \Bw_a^* ~=~
 \Shrink \Big( \underbrace{ \Bg_a - \frac{\boldsymbol{\lambda}_{\boldsymbol{2}a}}{\beta_2} }_{:=~ \Bt_{\Bw_a}}
 \,, \frac{\mu_1}{\beta_2} \Big).
\end{equation}

% Step 3
% g_s problem:
{\bfseries The ``$\big[ \Bg_s \big]_{s=0}^{S-1}$-problem'':} Fix $\Bu$, $\Bv$, $\Beps$, $\big[ \Br_l \big]_{l=0}^{L-1}$, $\big[ \Bw_s \big]_{s=0}^{S-1}$ and
\begin{equation} \label{eq:sub:g}
 \min_{\big[\Bg_s \big]_{s=0}^{S-1} \in X^S}
 \Bigg\{ \frac{\beta_2}{2} \sum_{s=0}^{S-1} \norm{\Bw_s - \Bg_s + \frac{\boldsymbol{\lambda}_{\boldsymbol{2}s}}{\beta_2}}_{\ell_2}^2 
%  \\& \qquad \qquad
 + \frac{\beta_3}{2} \norm{ \Bv - \sum_{s=0}^{S-1} \Big[ \cos\big( \frac{\pi s}{S} \big) \Bg_s \BDnT 
 + \sin\big( \frac{\pi s}{S} \big) \BDm \Bg_s \Big] + \frac{\boldsymbol{\lambda_3}}{\beta_3} }_{\ell_2}^2
 \Bigg\}
\end{equation}

For the discrete finite frequency coordinates $\boldsymbol \omega = [\boldsymbol \omega_1 \,, \boldsymbol \omega_2] \in \cI$,
let be $\Bz = [\Bzo \,, \Bzt] = \big[ e^{j\Bomeo} \,, e^{j\Bomet} \big]$.
We denote by
$W_a(\Bz) \,, \Lambda_{2a}(\Bz) \,, V(\Bz) \,, G_s(\Bz)$ and $\Lambda_3(\Bz)$
the discrete Fourier transforms of 
$w_a[\boldsymbol k] \,, \lambda_{2a}[\boldsymbol k] \,, v[\boldsymbol k] \,, g_s[\boldsymbol k] $
and $\lambda_3[\boldsymbol k]$, respectively.
Due to the separability, the solution of (\ref{eq:sub:g}) is obtained for $a = 0 \,, \ldots \,, S-1$ as

\begin{equation} \label{eq:sub:solution:ga}
 \Bg_a^* ~=~ \RE \Big[ \mathcal F^{-1} \big\{ \mathcal A(\Bz) \cdot \mathcal B (\Bz) \big\} \Big] 
\end{equation}
with
\begin{align*}
 &\mathcal A(\Bz) ~=~ \Bigg[ \beta_2 + \beta_3 \Big[ \sin \frac{\pi a}{S} (\BzoI - 1) + \cos \frac{\pi a}{S} (\BztI - 1) \Big]
 \Big[ \sin \frac{\pi a}{S} (\Bzo - 1) + \cos \frac{\pi a}{S} (\Bzt - 1) \Big] 
 \Bigg]^{-1}
 \,,
 \\
 &\mathcal B(\Bz) ~=~ 
 \beta_2 \Big[ W_a(\Bz)  + \frac{\Lambda_{2a}(\Bz) }{\beta_2} \Big]
 ~+~ \beta_3 \Big[ \sin\big( \frac{\pi a}{S} \big) (\BzoI -1) + \cos\big( \frac{\pi a}{S} \big) (\BztI-1) \Big] \times
 \\&
 \bigg[ V(\Bz) - \sum_{s=[0\,,S-1] \backslash \{a\} } 
 \Big[ \cos\big( \frac{\pi s}{S} \big) (\Bzt-1) 
 + \sin\big( \frac{\pi s}{S} \big) (\Bzo-1) \Big] G_s(\Bz)
 + \frac{\Lambda_3(\Bz)}{\beta_3} \bigg] \,.   
\end{align*}

% Step 4
% v problem
{\bfseries The ``$\Bv$-problem'':} Fix $\Bu$, $\Beps$, $\big[ \Br_l \big]_{l=0}^{L-1}$, $\big[ \Bw_s \big]_{s=0}^{S-1}$, $\big[ \Bg_s \big]_{s=0}^{S-1}$ and
\begin{align} \label{eq:sub:v}
 \min_{\Bv \in X} \Big\{ & 
 \mu_2 \norm{\Bv}_{\ell_1} 
 + \frac{\beta_3}{2} \norm{ \Bv - \bigg( \sum_{s=0}^{S-1} \Big[ \cos\big( \frac{\pi s}{S} \big) \Bg_s \BDnT 
 + \sin\big( \frac{\pi s}{S} \big) \BDm \Bg_s \Big] - \frac{\boldsymbol{\lambda_3}}{\beta_3} \bigg) }^2_{\ell_2} \notag
 \\& \qquad \qquad \qquad 
 + \frac{\beta_4}{2} \norm{ \Bv - \bigg( \Bf - \Bu - \Beps + \frac{\boldsymbol{\lambda_4}}{\beta_4} \bigg) }_{\ell_2}^2
 \Big\}
\end{align}
The solution of (\ref{eq:sub:v}) is defined as %(cf Proposition \ref{prop:DTV-DG3part_vprob})
\begin{equation} \label{eq:sub:solution:v}
 \Bv^* ~=~ \Shrink \Big( \Bt_{\Bv} \,,~ \frac{\mu_2}{\beta_3 + \beta_4} \Big),
\end{equation}
with
\begin{equation} \label{eq:sub:solution:tv}
 \Bt_{\Bv} ~:=~ \frac{\beta_3}{\beta_3 + \beta_4} \bigg( \sum_{s=0}^{S-1} \Big[ \cos\big( \frac{\pi s}{S} \big) \Bg_s \BDnT 
 + \sin\big( \frac{\pi s}{S} \big) \BDm \Bg_s \Big] - \frac{\boldsymbol{\lambda_3}}{\beta_3} \bigg)
 + \frac{\beta_4}{\beta_3 + \beta_4} \bigg( \Bf - \Bu - \Beps + \frac{\boldsymbol{\lambda_4}}{\beta_4} \bigg).
\end{equation}

% Step 5
% u problem
{\bfseries The ``$\Bu$-problem'':} Fix $\Bv$, $\Beps$, $\big[ \Br_l \big]_{l=0}^{L-1}$, $\big[ \Bw_s \big]_{s=0}^{S-1}$, $\big[ \Bg_s \big]_{s=0}^{S-1}$ and
\begin{equation} \label{eq:sub:u}
 \min_{\Bu \in X} \Big\{
 \frac{\beta_1}{2} \sum_{l=0}^{L-1} \norm{ \Br_l - \cos\big( \frac{\pi l}{L} \big) \Bu \BDnT - \sin\big( \frac{\pi l}{L} \big) \BDm \Bu 
 + \frac{\boldsymbol{\lambda}_{\boldsymbol{1} l}}{\beta_1} }^2_{\ell_2}
 + \frac{\beta_4}{2} \norm{ \Bf - \Bu - \Bv - \Beps + \frac{\boldsymbol{\lambda_4}}{\beta_4} }_{\ell_2}^2
 \Big\}
\end{equation}

We denote 
$F(\Bz) \,, \mathcal{E}(\Bz) \,, \Lambda_4(\Bz) \,, R_l(\Bz)$ and $\Lambda_{1l}(\Bz)$
as the discrete Fourier transforms of 
$f[\boldsymbol k] \,, \epsilon[\boldsymbol k] \,, \lambda_4[\boldsymbol k] \,, r_l[\boldsymbol k] $
and $\lambda_{1l}[\boldsymbol k]$, respectively.
This (\ref{eq:sub:u}) is solved in the Fourier domain by 

\begin{equation} \label{eq:sub:solution:u}
 \Bu^* ~=~ \RE \Big[ \mathcal F^{-1} \big\{ \mathcal X(\Bz) \cdot \mathcal Y(\Bz) \big\} \Big] 
\end{equation}
with
\begin{align*}
 \mathcal X(\Bz) &= 
 \Bigg[ \beta_4 + \beta_1 \sum_{l=0}^{L-1} \Big[ \sin\big( \frac{\pi l}{L} \big) (\BzoI - 1)  
 + \cos\big( \frac{\pi l}{L} \big) (\BztI -1) \Big] 
 \Big[ \sin\big( \frac{\pi l}{L} \big) (\Bzo -1) + \cos\big( \frac{\pi l}{L} \big) (\Bzt -1) \Big] 
 \Bigg]^{-1} \,,
 \\
 \mathcal Y(\Bz) &= 
 \beta_4 \Big[ F(\Bz) - V(\Bz) - \mathcal{E}(\Bz) + \frac{\Lambda_4(\Bz)}{\beta_4} \Big]
 + \beta_1 \sum_{l=0}^{L-1} \Big[ \sin \big( \frac{\pi l}{L} \big) (\BzoI-1) + \cos \big( \frac{\pi l}{L} \big) (\BztI -1) \Big]
 \Big[ R_l(\Bz) + \frac{\Lambda_{1l}(\Bz)}{\beta_1} \Big] .
\end{align*}

% Step 6
% Epsilon problem
{\bfseries The ``$\Beps$-problem'':} Fix $\Bu$, $\Bv$, $\big[ \Br_l \big]_{l=0}^{L-1}$, $\big[ \Bw_s \big]_{s=0}^{S-1}$, $\big[ \Bg_s \big]_{s=0}^{S-1}$ and
\begin{align} \label{eq:sub:e}
 \min_{\Beps \in X} \Big\{ 
 G^* \big( \frac{\Beps}{\delta} \big) + \frac{\beta_4}{2} \norm{\Beps - \big( \Bf - \Bu - \Bv + \frac{\boldsymbol{\lambda_4}}{\beta_4} \big)}^2_{\ell_2}
 \Big\}
\end{align}
Let $\mathcal C^*$ be the inverse curvelet transform \cite{CandesDemanetDonohoYing2006}.
The minimization of (\ref{eq:sub:e}) is solved by (cf. \cite{AujolChambolle2005})
\begin{equation*}
 \Beps^* = \Big( \Bf - \Bu - \Bv + \frac{\boldsymbol{\lambda_4}}{\beta_4} \Big) ~-~
 \underbrace{ \mathcal{C}^* \Big\{ \Shrink \big( \mathcal C \big\{ \Bf - \Bu - \Bv + \frac{\boldsymbol{\lambda_4}}{\beta_4} \big\} \,, \delta \big) \Big\} }
 _{ :=~ \CST \big( \Bf - \Bu - \Bv + \frac{\boldsymbol{\lambda_4}}{\beta_4} \,, \delta \big) }   
\end{equation*}
or by the projection method with the component-wise operators 
\begin{equation*}
 \Beps^* =
 \cC^* \Bigg\{ \frac{\delta \, \cC \big\{ \Bf - \Bu - \Bv + \frac{\boldsymbol{\lambda_4}}{\beta_4} \big\}}
 { \max\big( \delta \,,~ \abs{\cC \big\{ \Bf - \Bu - \Bv + \frac{\boldsymbol{\lambda_4}}{\beta_4} \big\}} \big) } \Bigg\}.
\end{equation*}

% Step 7
{\bfseries Update Lagrange Multipliers}
  $\Big( \big[\boldsymbol{\lambda}_{\mathbf{1}l}\big]_{l=0}^{L-1} \,, \big[\boldsymbol{\lambda}_{\mathbf{2}a}\big]_{a=0}^{S-1} 
  \,, \boldsymbol{\lambda_3}\,,  \boldsymbol{\lambda_4} \Big) \in X^{L+S+2}$: 
  \begin{align*}
   \boldsymbol{\lambda}_{\mathbf{1}b}^{(t)} &~=~ \boldsymbol{\lambda}_{\mathbf{1}b}^{(t-1)} 
   ~+~ \gamma \beta_1 \Big( \mathbf{r}_b - \cos\big( \frac{\pi b}{L} \big) \Bu \BDnT - \sin\big( \frac{\pi b}{L} \big) \BDm \Bu  \Big)  
   \,, \quad b = 0 \,, \ldots \,, L-1
   \\
   \boldsymbol{\lambda}_{\mathbf{2}a}^{(t)} &~=~ \boldsymbol{\lambda}_{\mathbf{2}a}^{(t-1)} 
   ~+~ \gamma \beta_2 \Big( \mathbf{w}_a - \mathbf{g}_a \Big)  
   \,, \quad a = 0 \,, \ldots \,, S-1
   \\
   \boldsymbol{\lambda}_{\mathbf{3}}^{(t)} &~=~ \boldsymbol{\lambda}_{\mathbf{3}}^{(t-1)} 
   ~+~ \gamma \beta_3 \Big( \Bv - \sum_{s=0}^{S-1} \big[ \cos\big(\frac{\pi s}{S} \big) \mathbf{g}_s \BDnT + \sin\big( \frac{\pi s}{S} \big) \BDm \mathbf{g}_s \big] \Big)  
   \\
   \boldsymbol{\lambda}_{\mathbf{4}}^{(t)} &~=~ \boldsymbol{\lambda}_{\mathbf{4}}^{(t-1)} 
   ~+~ \gamma \beta_4 \big( \Bf - \Bu - \Bv - \Beps \big)  
  \end{align*}

{\bfseries Choice of Parameters}

Due to the $\ell_1$-norms in the minimization problem (\ref{eq:DTV-DG3part_ALM}) 
which corresponds to the shrinkage operator
with parameters $\mu_1$ and $\mu_2$, these are defined as 
\begin{align}
 \mu_1 = c_{\mu_1} \beta_2 \cdot \max_{\Bk \in \Omega} \big( \abs{t_{\Bw_a}[\Bk]} \big)
 \quad \text{and} \quad
 \mu_2 = c_{\mu_2} (\beta_3 + \beta_4) \cdot \max_{\Bk \in \Omega} \big( \abs{t_\Bv[\Bk]} \big),
\end{align}
where $t_{\Bw_a}[\Bk]$ and $t_\Bv[\Bk]$ are defined in (\ref{eq:sub:solution:wa}) and (\ref{eq:sub:solution:tv}), respectively.
Note that the choice of $c_{\mu_1}$ and $c_{\mu_2}$ is adapted to specific images.

In order to balance between the smoothing terms and the updated terms for the solutions of the
$\Bg$-problem in (\ref{eq:sub:solution:ga}), the $\Bv$-problem in (\ref{eq:sub:solution:v})
and the $\Bu$-problem in (\ref{eq:sub:solution:u}), we choose 
$$
\displaystyle \beta_2 = c_2 \beta_3 \,, \beta_3 = \frac{\theta}{1 - \theta} \beta_4\,,\theta \in (0,1)
\text{ and } \beta_1 = c_1 \beta_4.
$$

\begin{figure}
\begin{center}    
  % delta = 0:
  \subfigure[$\delta = 0 \,, \Bu$]{\includegraphics[width=0.244\textwidth]{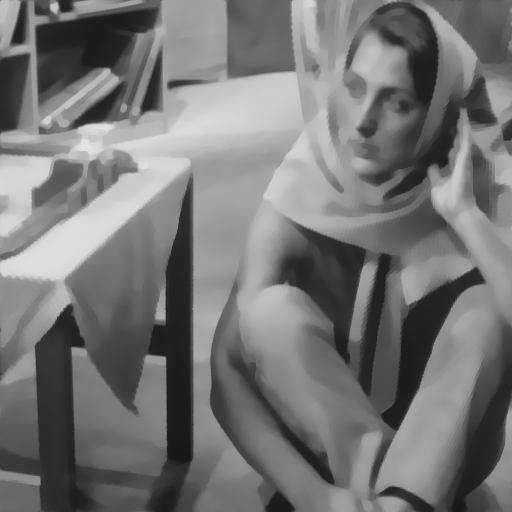}}
  \subfigure[$\Bv$]{\includegraphics[width=0.244\textwidth]{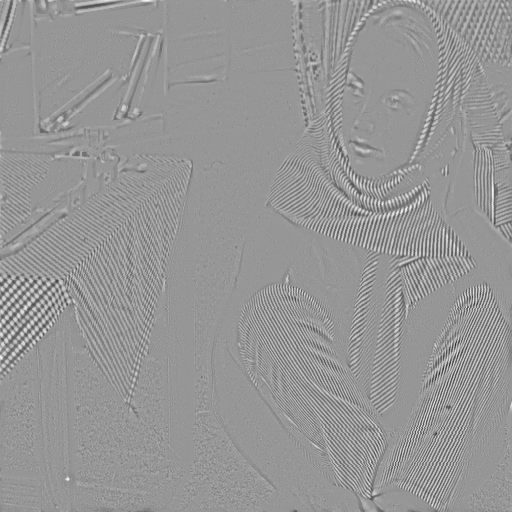}}
  \subfigure[$\Bv_\text{bin}$]{\includegraphics[width=0.244\textwidth]{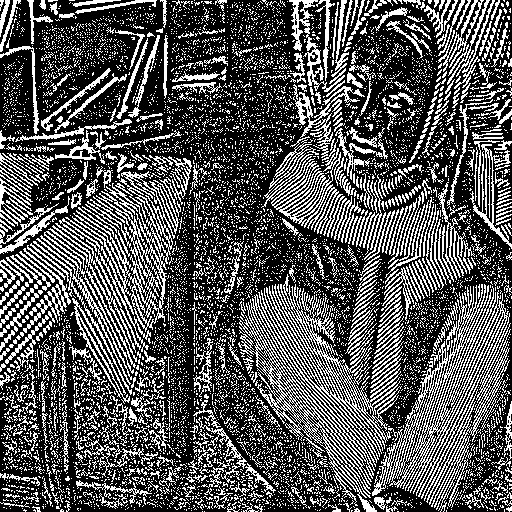}}
  \subfigure[$\Beps = \mathbf 0$]{\includegraphics[width=0.244\textwidth]{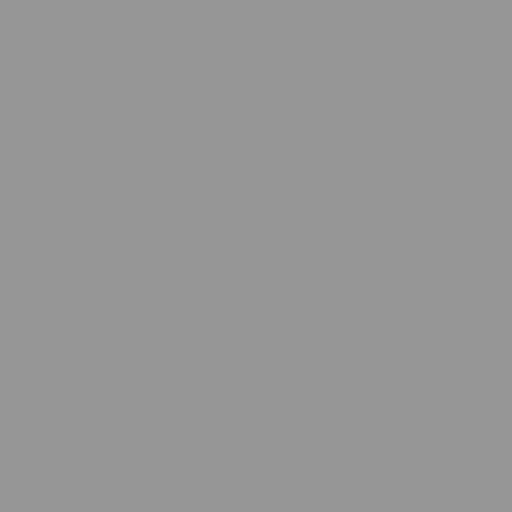}}  
  % delta = 10:
  \subfigure[$\delta = 10 \,, \Bu$]{\includegraphics[width=0.244\textwidth]{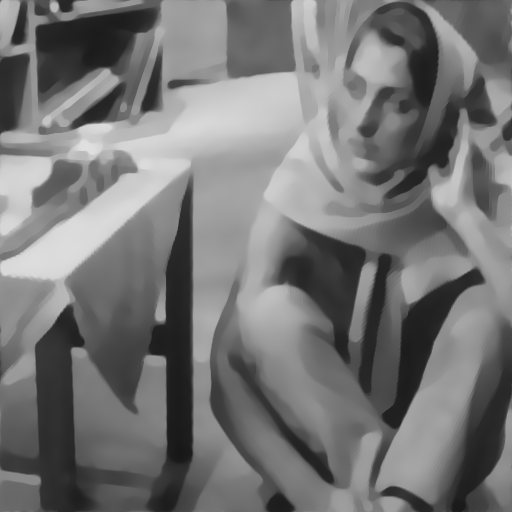}}
  \subfigure[$\Bv$]{\includegraphics[width=0.244\textwidth]{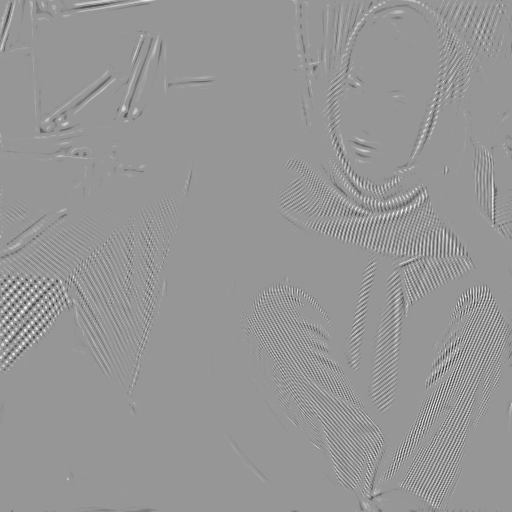}}
  \subfigure[$\Bv_\text{bin}$]{\includegraphics[width=0.244\textwidth]{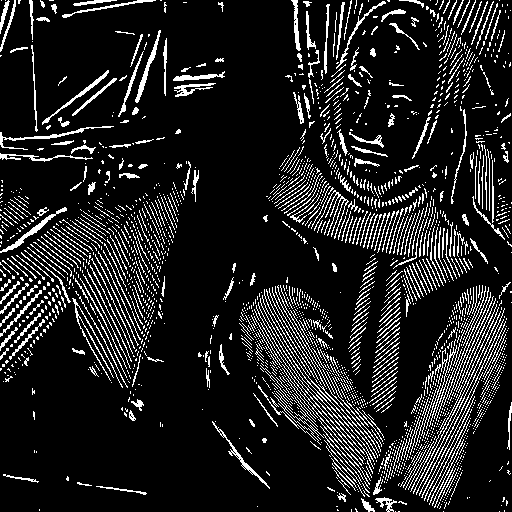}}
  \subfigure[$\Beps$]{\includegraphics[width=0.244\textwidth]{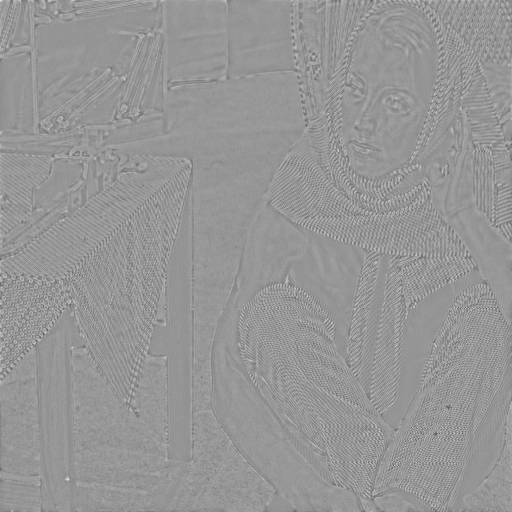}} 
  % Convergence, residual = f - u - v - epsilon: 
  \subfigure[$\delta = 10$]
  {\includegraphics[width=0.302\textwidth]{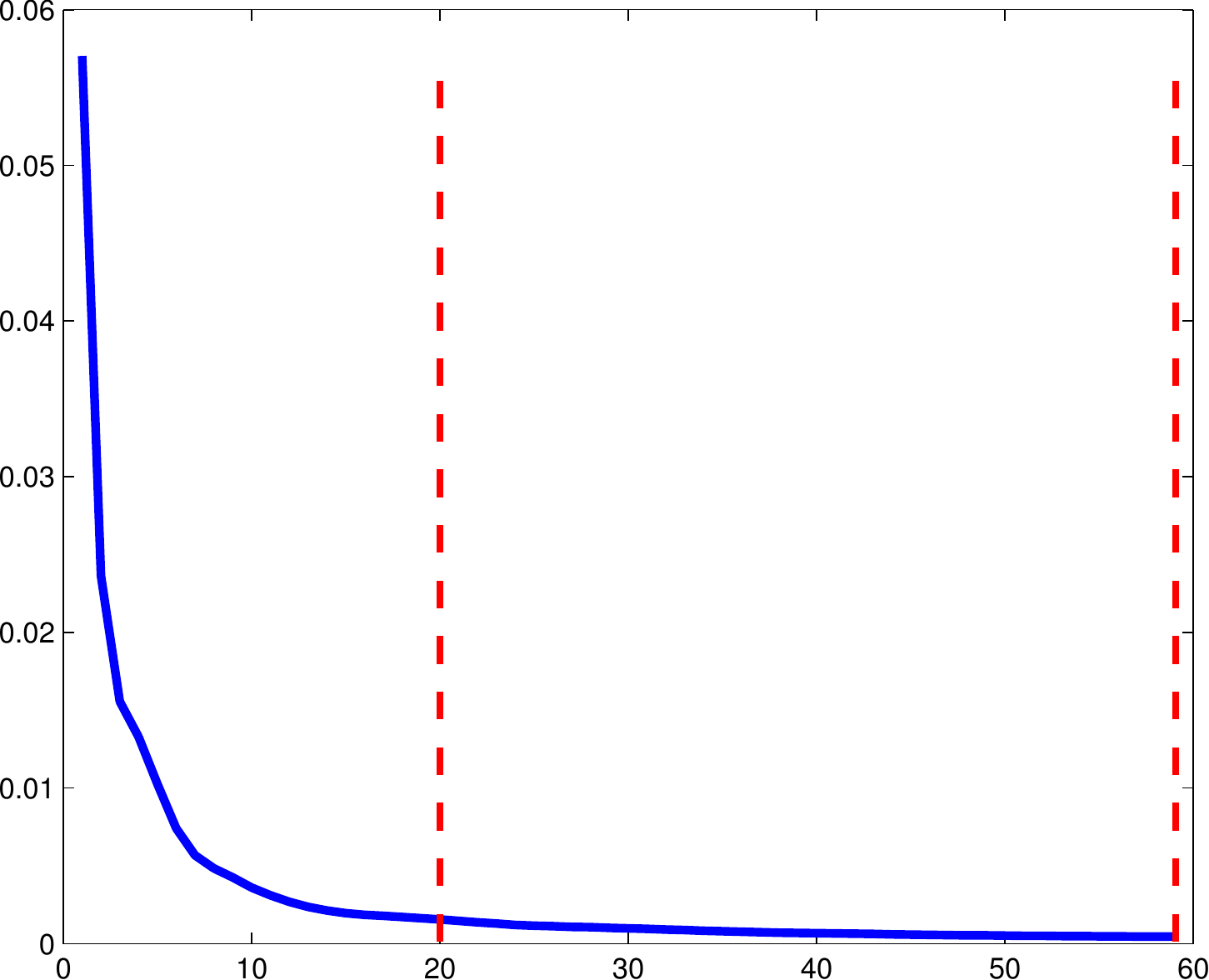}}  
  \subfigure[$150 + \Bf - \Bu - \Bv - \Beps$]{\includegraphics[width=0.244\textwidth]{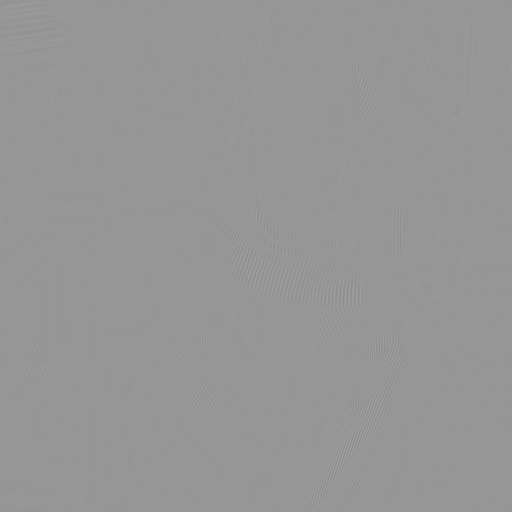}}
  \subfigure[$150 + \Bf - \Bu - \Bv - \Beps$]{\includegraphics[width=0.244\textwidth]{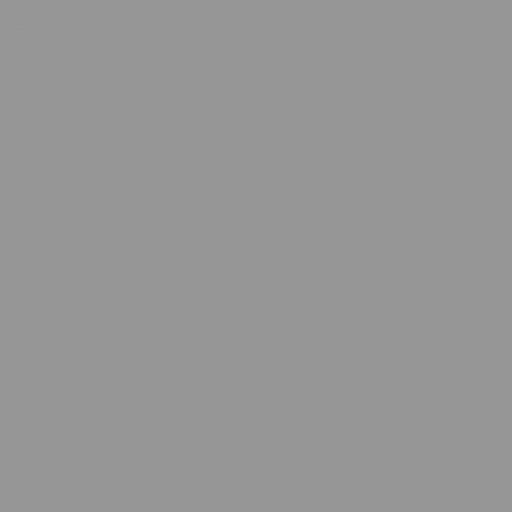}}  
\caption{Visualization of the decomposition results by DG3PD 
         with $\delta = 0$ (a-d)
         and $\delta = 10$ (e-h) after 20 iterations.
         The convergence rates for these decompositions
         with $\delta = 0$ and $\delta = 10$ are depicted 
         in Figure \ref{fig:comparison:Barbara:penaltyALM} (m)
         and in (i), respectively, 
				 which plots
				 the relative error (y-axis) as defined in \cite{ThaiGottschlich2015G3PD} 
				 versus the number of iterations (x-axis).				
         The parameters are $\beta_4 = 0.04 \,, \theta = 0.9 \,, c_1 = 1 \,, c_2 = 1.3 \,,
         c_{\mu_1} = c_{\mu_2} = 0.03 \,, \gamma = 1 \,, S = L = 9$.
				 Error images are illustrated in (j) after 20 iterations 
				 and in (k) after 60 iterations with $\delta = 10$.
        }
\label{fig:DG3PD:barbara}
\end{center}
\end{figure}

The choice of $\delta$ mainly impacts the smoothness and sparsity of the texture $\Bv$. 
The first row of Figure \ref{fig:DG3PD:barbara} 
shows the effect of selecting the threshold $\delta = 0$
which corresponds to a two-part decomposition, 
i.e. the residual image $\Beps = \mathbf 0$ in (d).
This case also demonstrates the limitation of all two-part decomposition approaches:
for this choice of $\delta$, very small scale objects are assigned to the texture image $\Bv$ 
in Figure \ref{fig:DG3PD:barbara} (b)
which is obvious in its binarization $\Bv_\text{bin}$ shown in (c).
In order to remove these and to yield a smoother and sparser texture $\Bv$, 
one can increase the value of $\delta$,
say e.g. by choosing $\delta = 10$.
The effect of this choice can be seen in the binarized version $\Bv_\text{bin}$ in Figure \ref{fig:DG3PD:barbara} (g)
and small scale objects are moved to the residual image $\Beps$ in (h). 
Therefore, the value of $\delta$ defines the level of the residual~$\Beps$.

\section{Comparison of DG3PD with Prior Art} \label{sec:comparisonPriorArt}

As stated before, the main objective of the DG3PD model is to achieve the following three goals (cf. Section \ref{sec:Introduction}):
\begin{itemize}
 \item Goal 1: $\Bu$ contains only geometrical objects with a very smooth surface, sharp boundaries and no texture.
 \item Goal 2: $\Bv$ contains only objects with sparse oscillating patterns 
               and $\Bv$ shall be both smooth and sparse.
 \item Goal 3: Perfect reconstruction of $\Bf$,  i.e. $\Bf = \Bu + \Bv + \Beps$.
\end{itemize}

\begin{figure}
\begin{center}  
	\subfigure[]{\includegraphics[height=0.2\textheight]{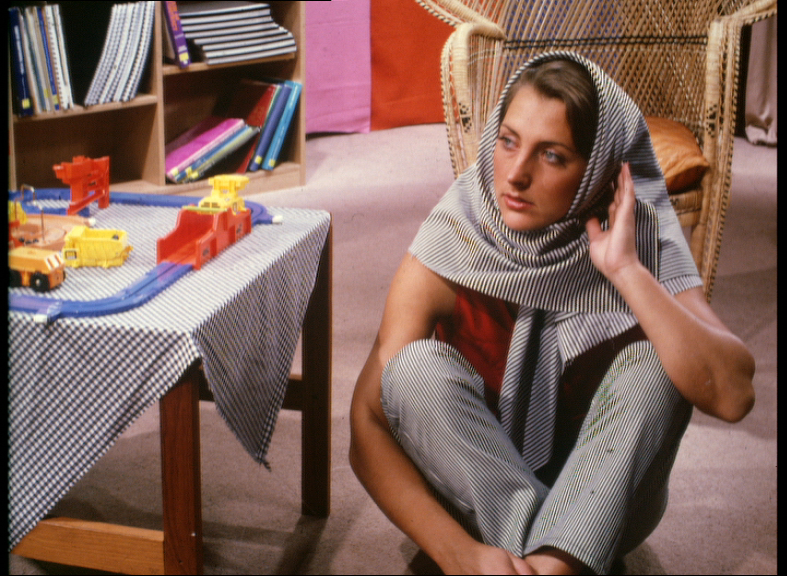}}
  \subfigure[]{\includegraphics[height=0.2\textheight]{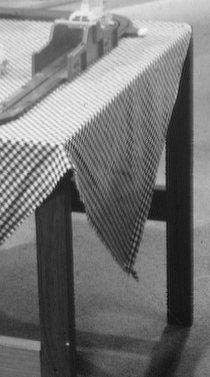}}
  \subfigure[]{\includegraphics[height=0.2\textheight]{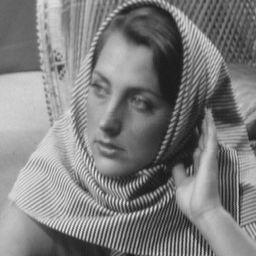}}
\caption{The original Barbara image (a) and highlighted details (b, c).}
\label{fig:OriginalImage}
\end{center}
\end{figure}

Based on these criteria 
we compare the proposed DG3PD model in this section with the state-of-the-art methods
using the original Barbara image.
We highlight selected regions for an improved conspicuousness of the differences 
between the considered methods, cf. Figure \ref{fig:OriginalImage}: 
\begin{itemize}
 \item Images without noise: 
       Rudin, Osher, and Fatemi (ROF) \cite{RudinOsherFatemi1992}, 
			 Vese and Osher (VO) \cite{VeseOsher2003}, 
			 Starck, Elad, and Donoho (SED) \cite{StarckEladDonoho2005} 
       and TV-Gabor (TVG) by Aujol \textit{et al.} \cite{AujolGilboaChanOsher2006} models.
 \item Images suffering from i.i.d. Gaussian noise $\mathcal N(0 \,, \sigma)$:
       the Aujol and Chambolle (AC) model \cite{AujolChambolle2005}.
\end{itemize}

% Original Barbara:

\begin{figure}
\begin{center}  
  
  % ROF, VeseOsher, StarckEladDonoho, A2BC:
  % u:
  \subfigure[ROF: $\Bu$]{\includegraphics[width=0.22\textwidth,width=0.22\textwidth]{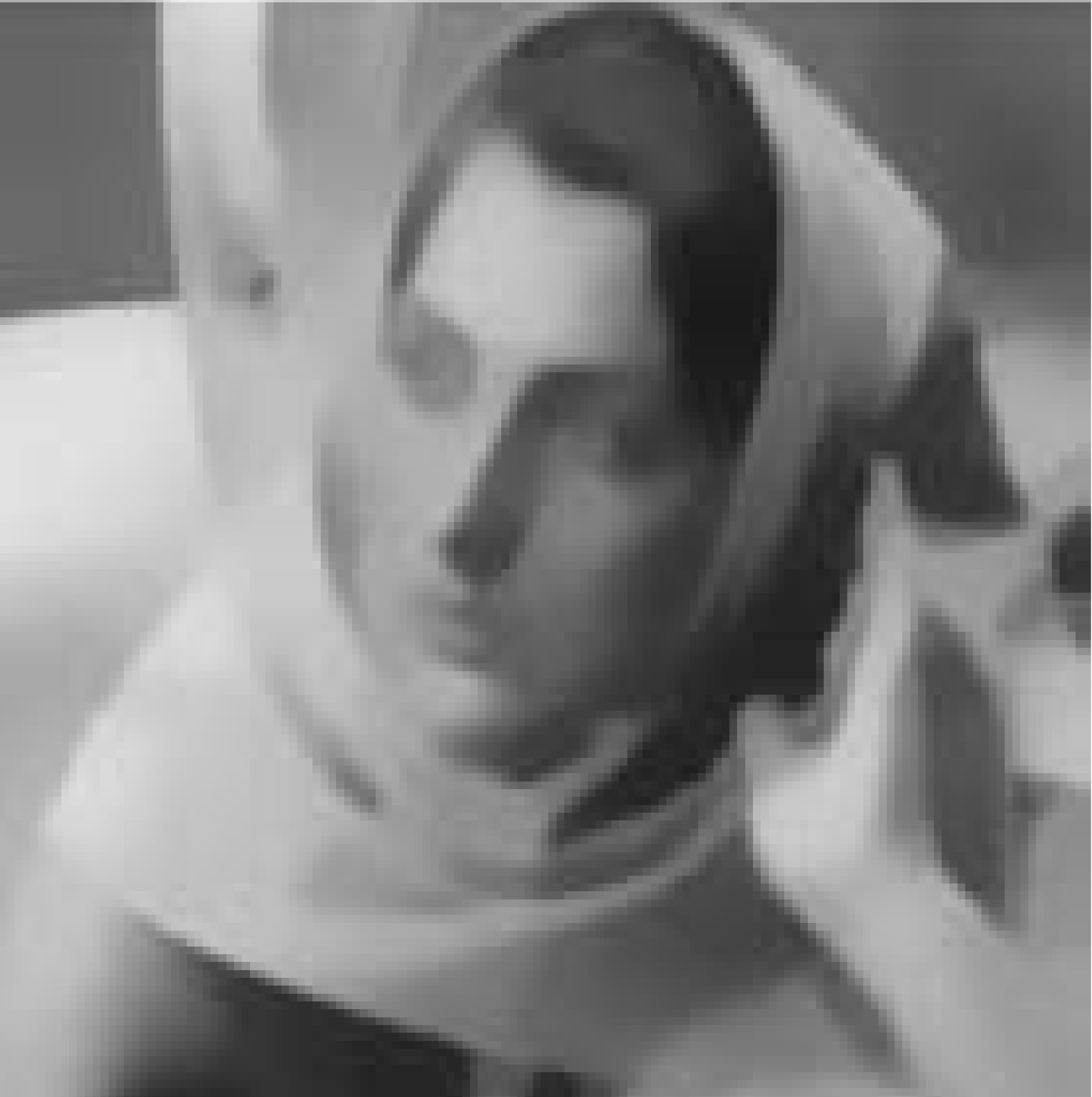}}
  \subfigure[VO: $\Bu$]{\includegraphics[width=0.22\textwidth,width=0.22\textwidth]{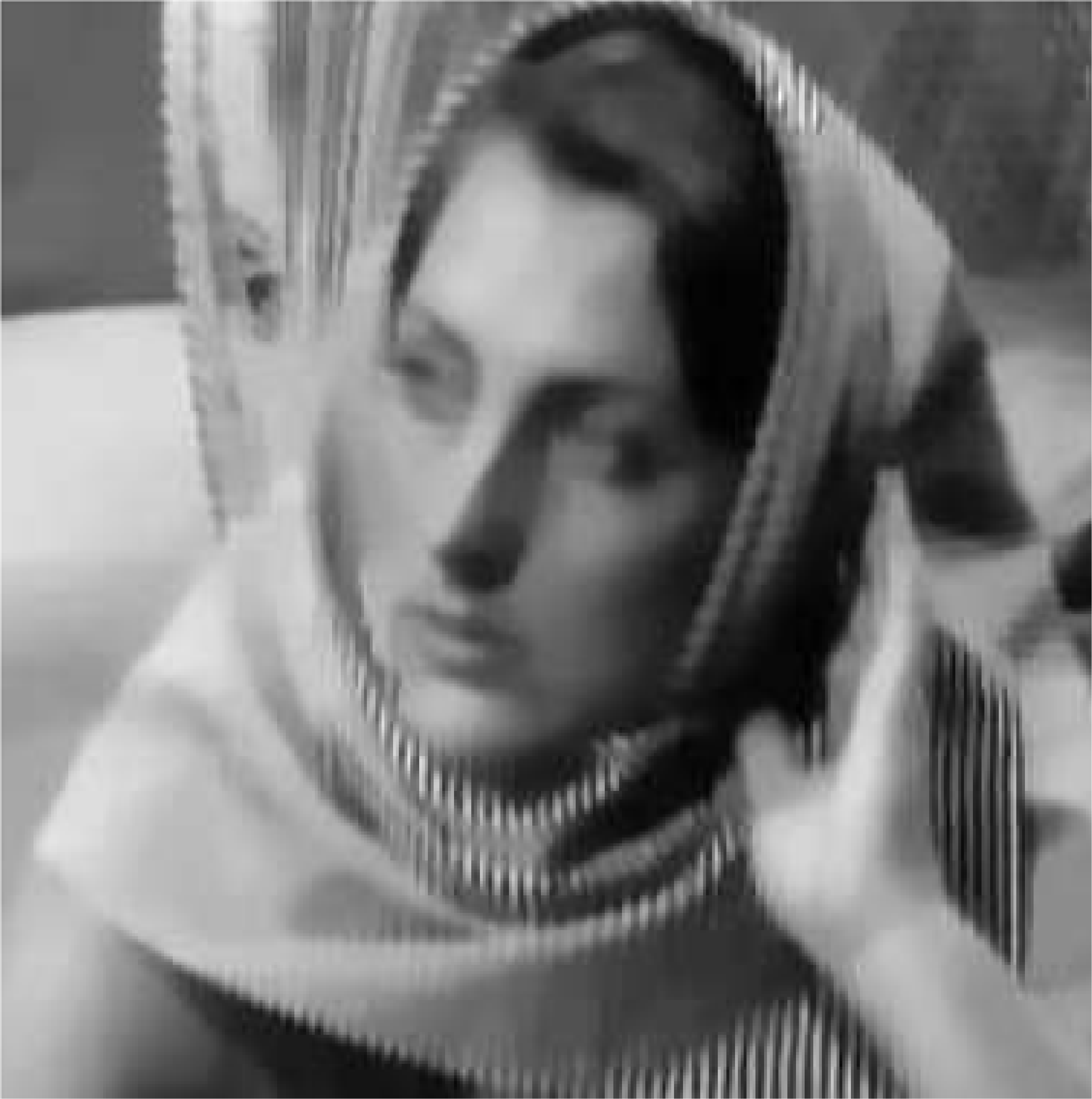}}
  \subfigure[SED: $\Bu$]{\includegraphics[width=0.22\textwidth,width=0.22\textwidth]{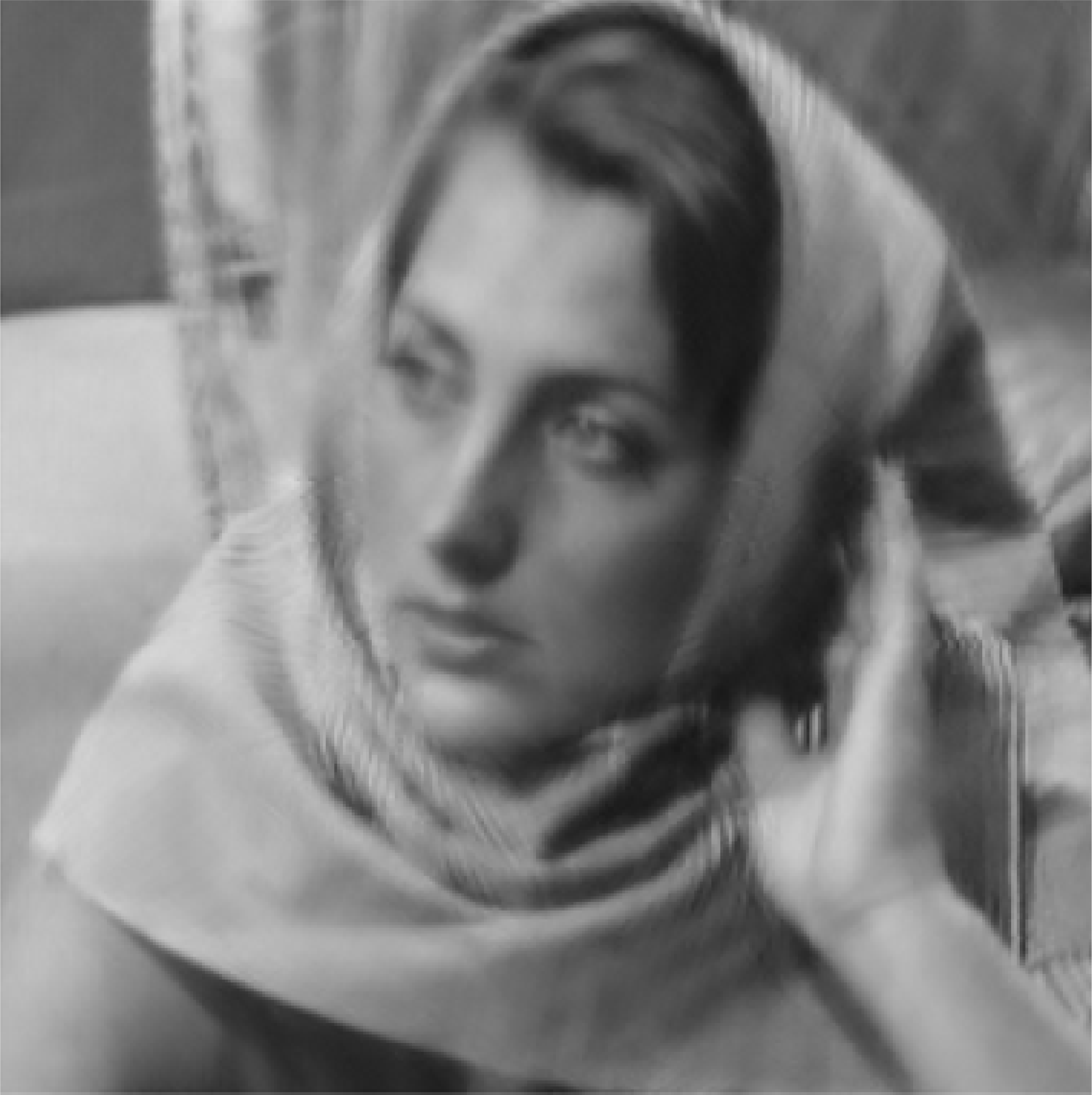}}
  \subfigure[TVG: $\Bu$]{\includegraphics[width=0.22\textwidth,width=0.22\textwidth]{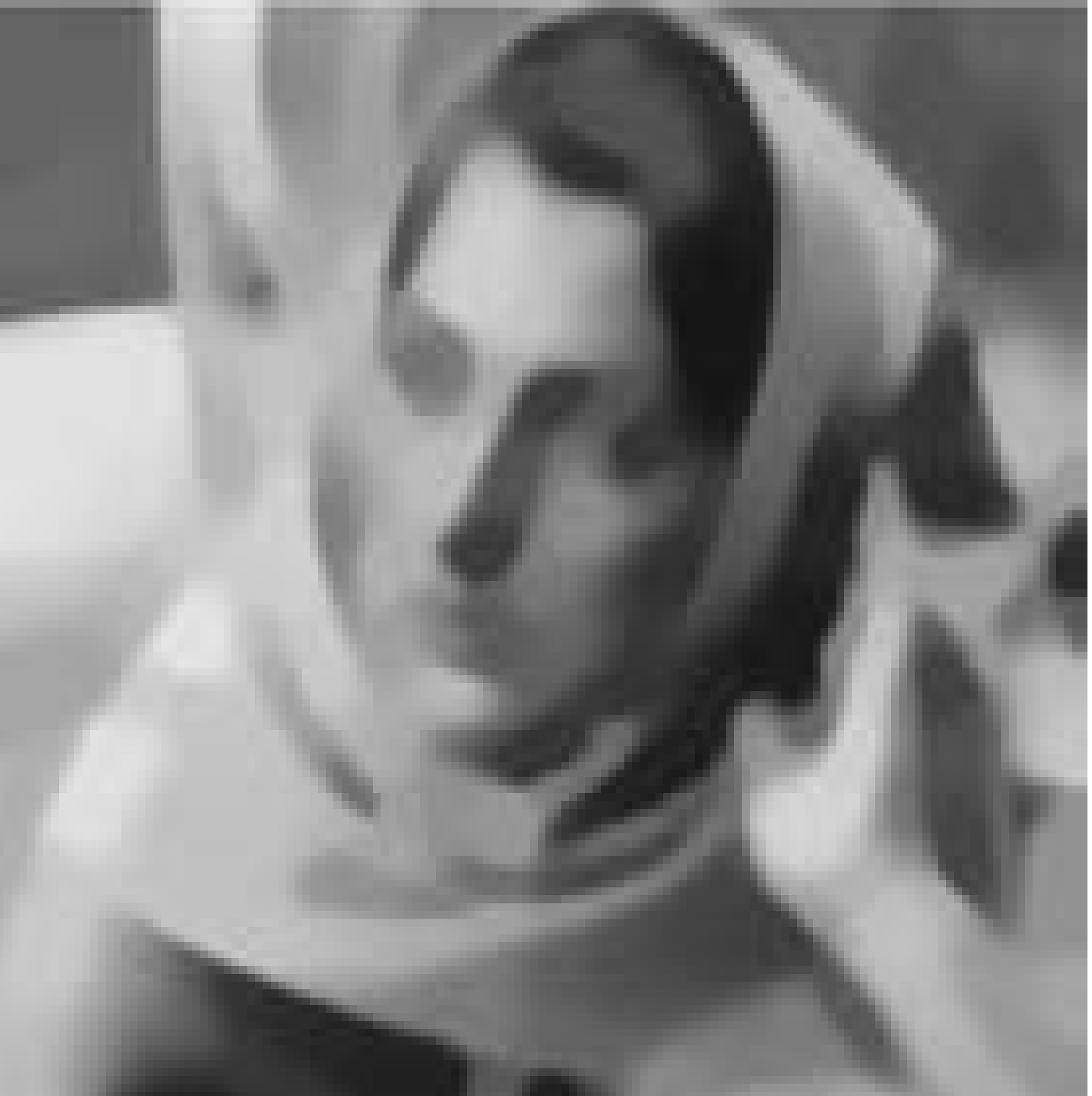}}
  
  % v:
  \subfigure[ROF: $\Bv$]{\includegraphics[width=0.22\textwidth,width=0.22\textwidth]{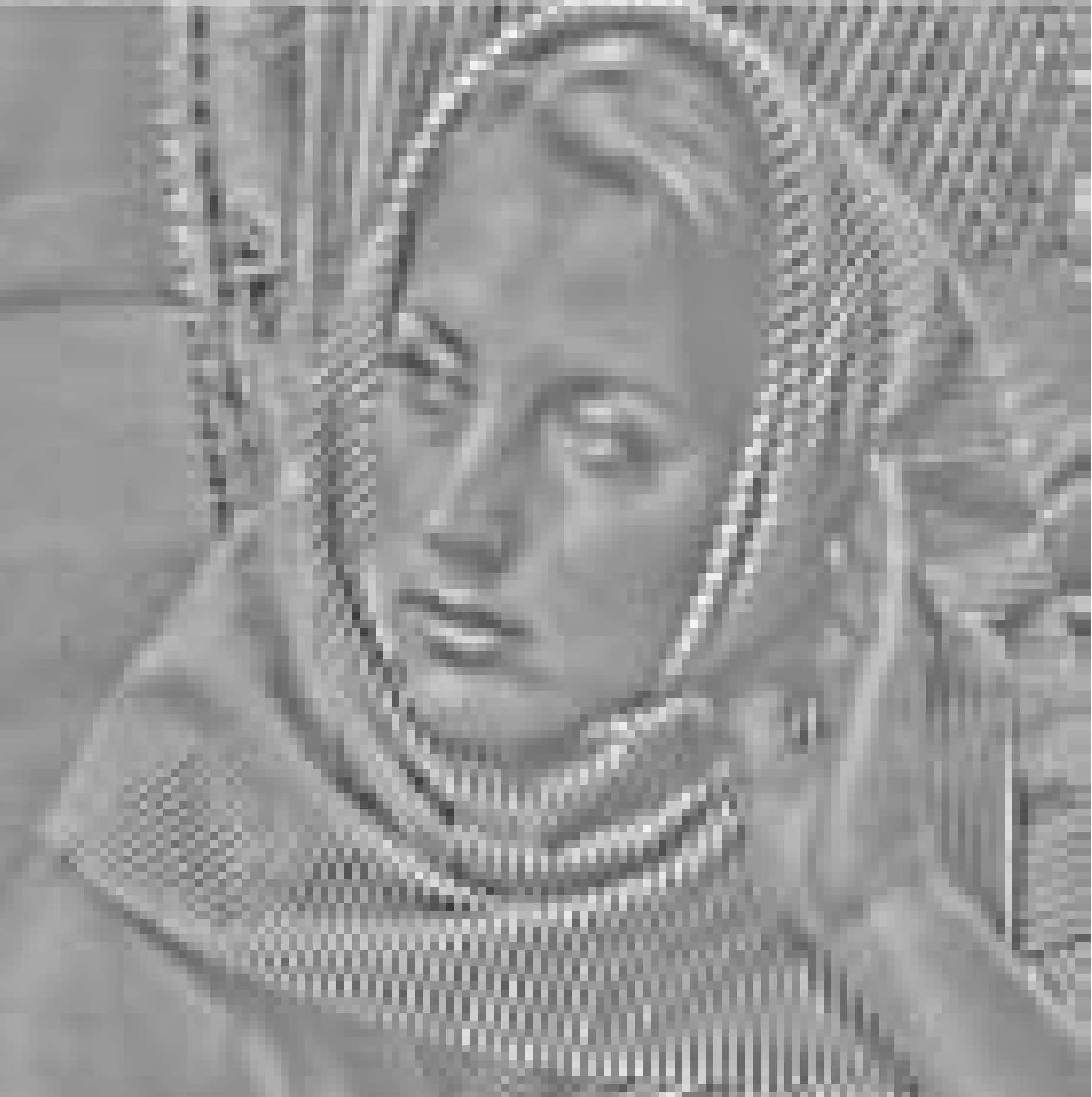}}  
  \subfigure[VO: $\Bv$]{\includegraphics[width=0.22\textwidth,width=0.22\textwidth]{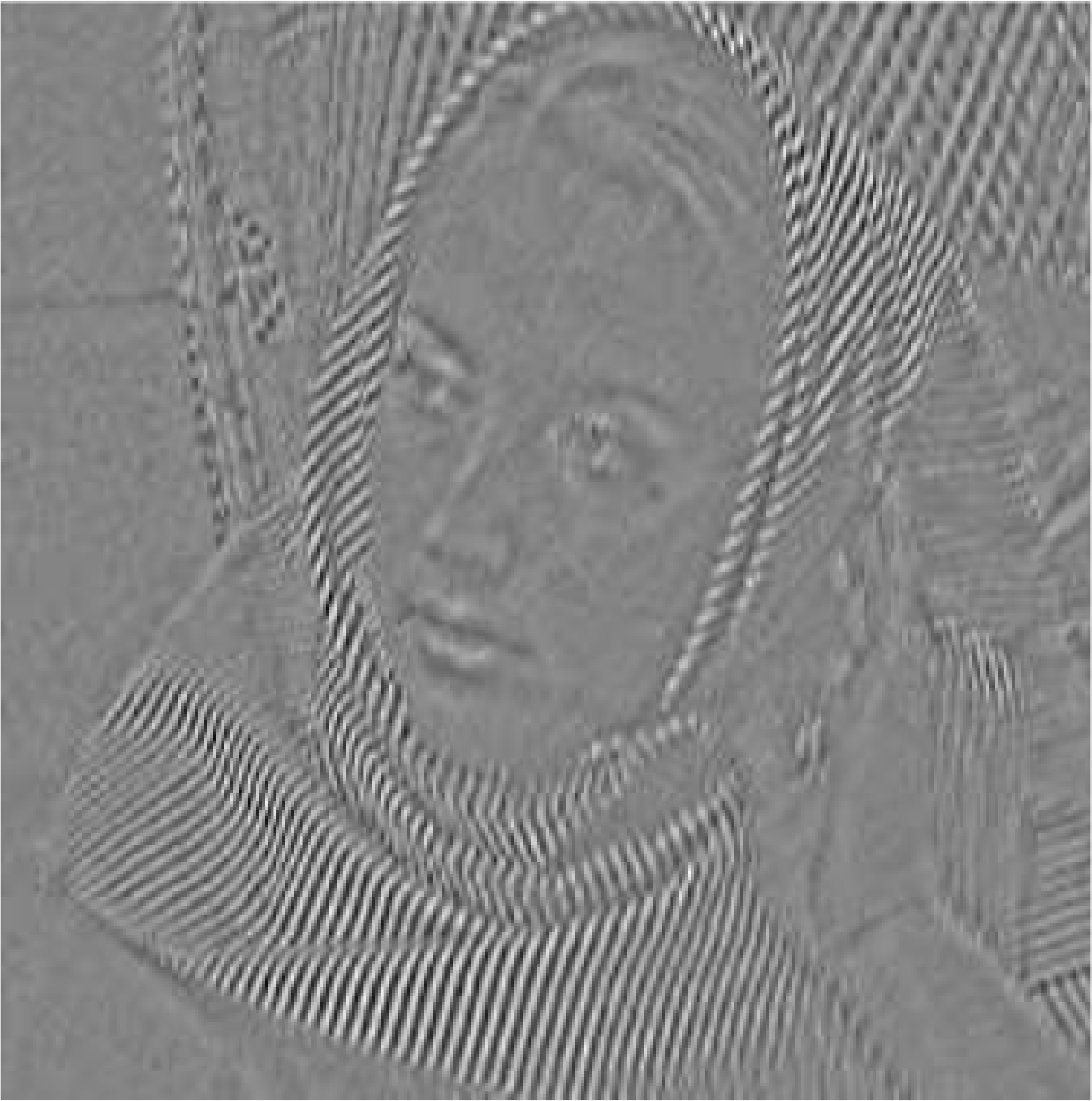}}
  \subfigure[SED: $\Bv$]{\includegraphics[width=0.22\textwidth,width=0.22\textwidth]{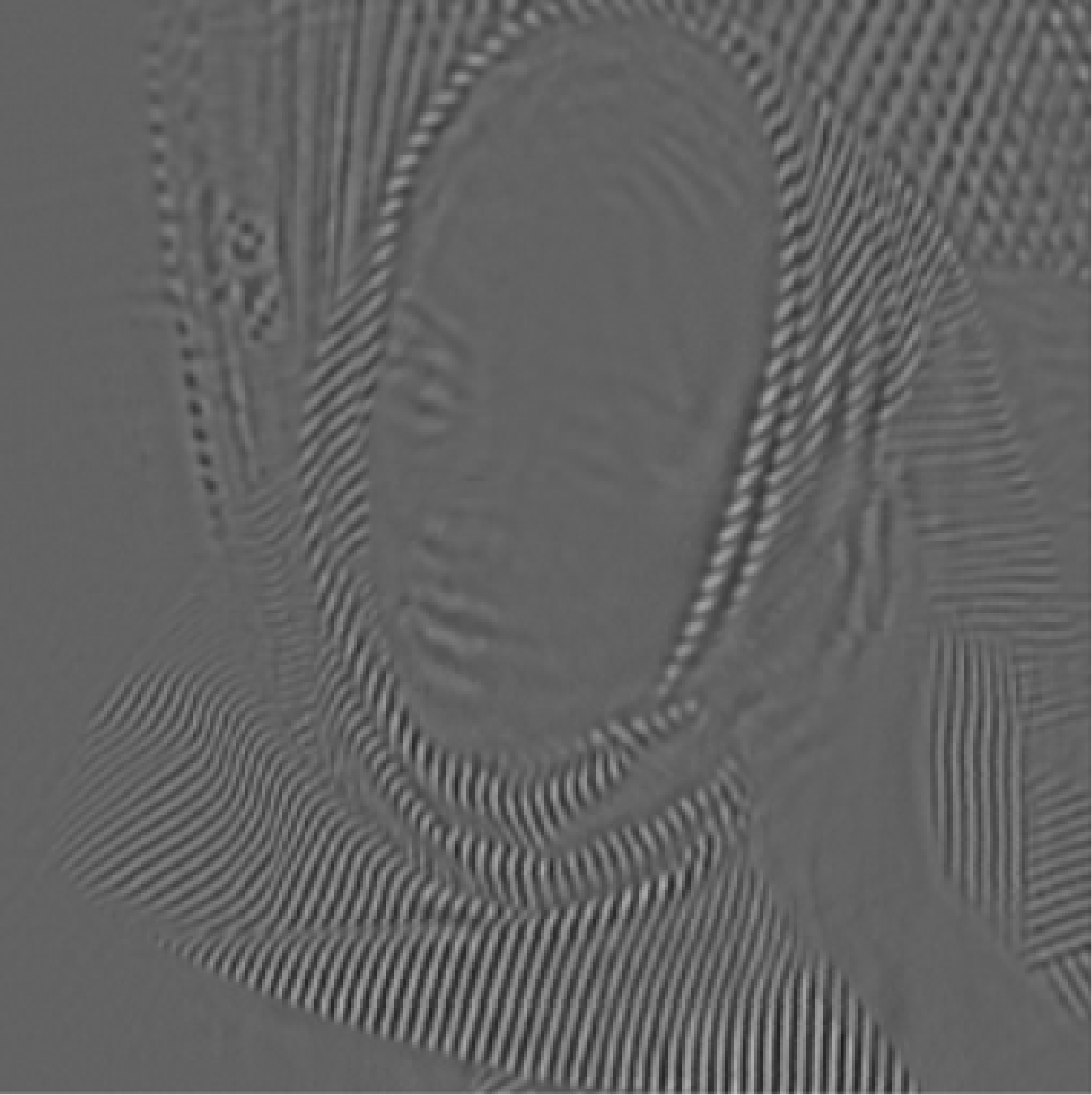}}
  \subfigure[TVG: $\Bv$]{\includegraphics[width=0.22\textwidth,width=0.22\textwidth]{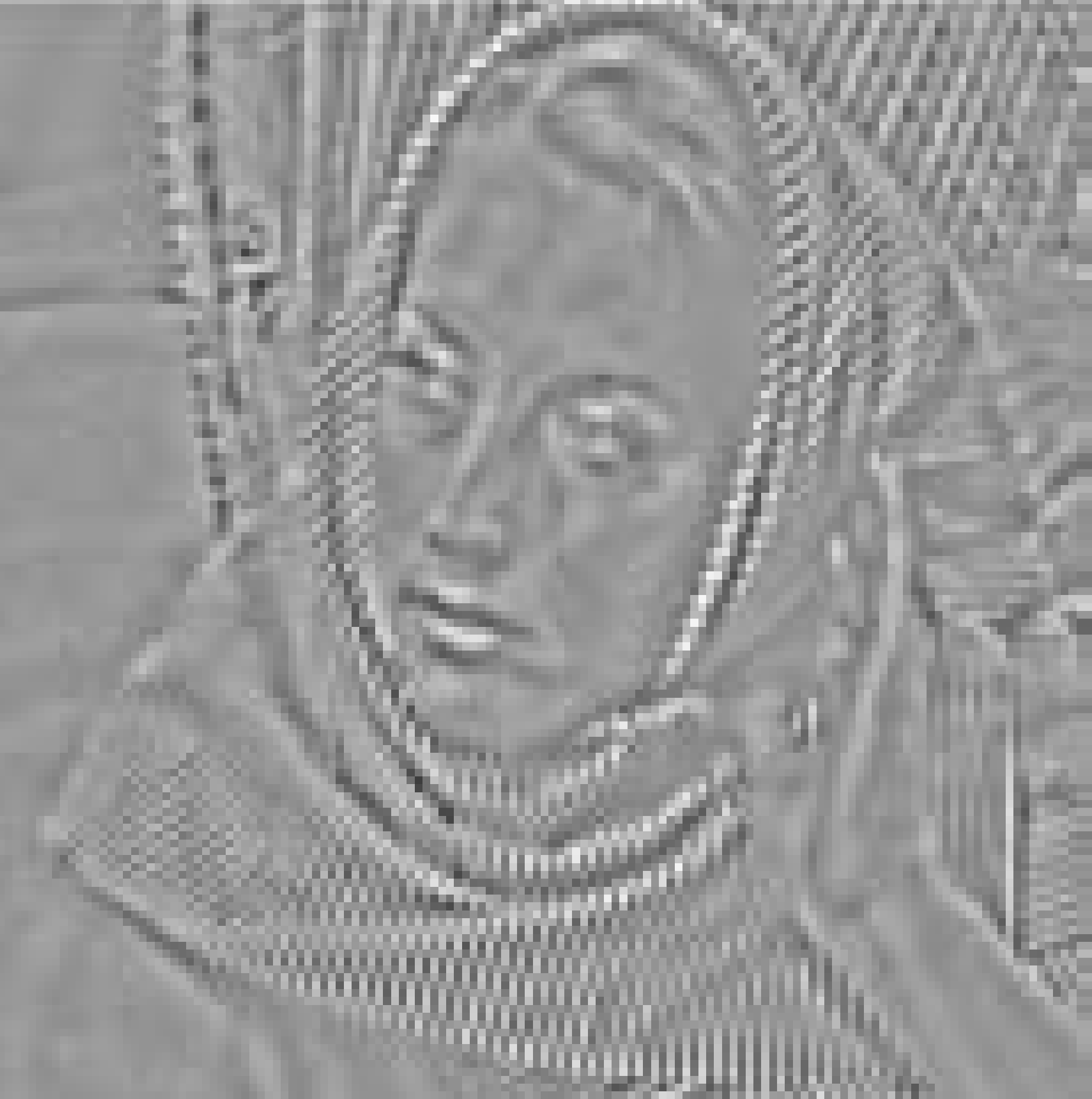}}  
  
  % G3PD:
  % delta = 0:
  \subfigure[DG3PD: $\Bu$, $\delta = 0$]{\includegraphics[width=0.22\textwidth,width=0.22\textwidth]{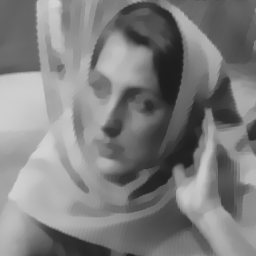}}  
  \subfigure[$\Bv$]{\includegraphics[width=0.22\textwidth,width=0.22\textwidth]{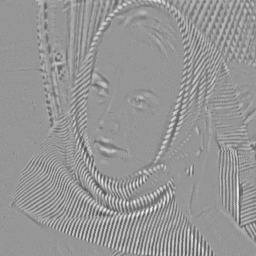}}
  \subfigure[$\Bv_\text{bin}$]{\includegraphics[width=0.22\textwidth,width=0.22\textwidth]{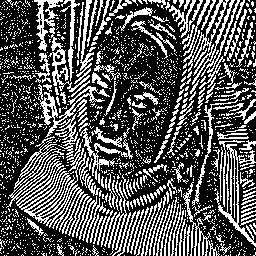}}
  \subfigure[$\Beps = \mathbf 0$]{\includegraphics[width=0.22\textwidth,width=0.22\textwidth]{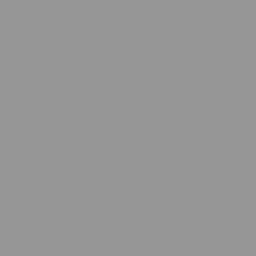}}    
  
  % delta = 10:
  \subfigure[DG3PD: $\Bu$, $\delta = 10$]{\includegraphics[width=0.22\textwidth,width=0.22\textwidth]{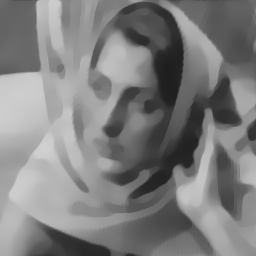}}  
  \subfigure[$\Bv$]{\includegraphics[width=0.22\textwidth,width=0.22\textwidth]{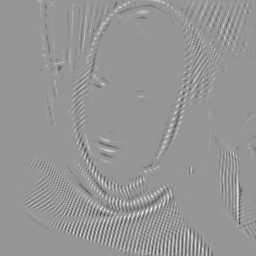}}
  \subfigure[$\Bv_\text{bin}$]{\includegraphics[width=0.22\textwidth,width=0.22\textwidth]{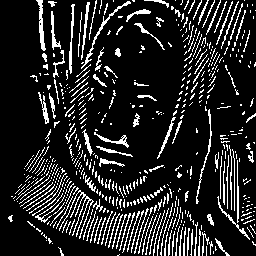}}
  \subfigure[$\Beps$]{\includegraphics[width=0.22\textwidth,width=0.22\textwidth]{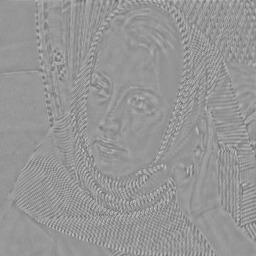}}      
  
\caption{The images in the first row depict the comparison of the cartoon $\Bu$ for different methods in the literature 
         on the original highlighted region (cf. Figure \ref{fig:OriginalImage}(c)). 
         The images in the second row show their corresponding texture $\Bv$. 
         Note that the ROF is reported in \cite{AujolGilboaChanOsher2006}.
         The images in the third row are obtained by the DG3PD model with $\delta = 0$
         and $\delta = 10$ in the fourth row (cf. Figure \ref{fig:DG3PD:barbara} for the whole image in these two cases).
        }
\label{fig:comparisonMethods:Barbara_crop1}
\end{center}
\end{figure}

% compare3:
\begin{figure}
\begin{center}  
  
  % ROF, VeseOsher (0.1573), StarckEladDonoho, TVG: 0.244
  % u:
  \subfigure[ROF]{\includegraphics[width=0.207\textwidth]{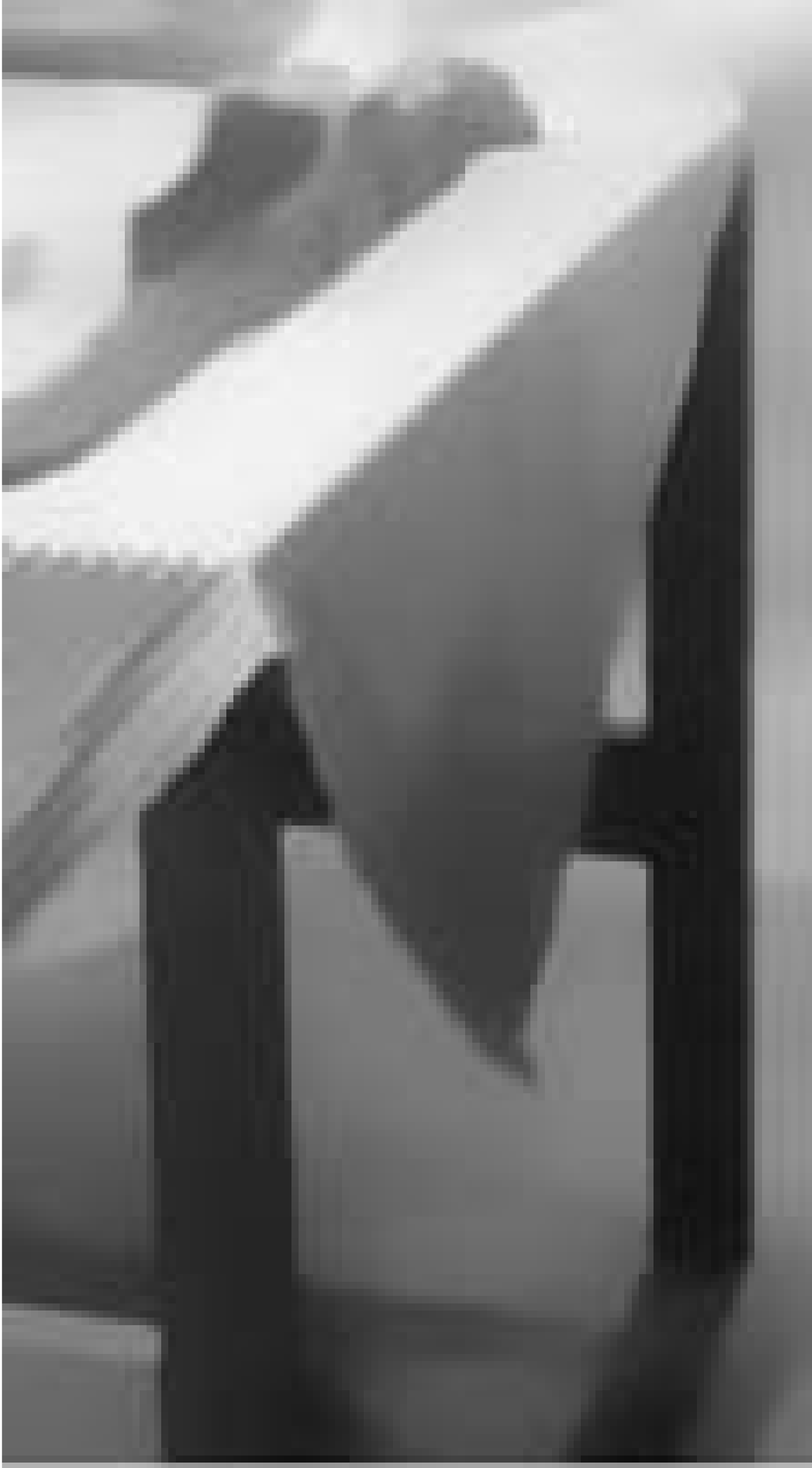}}
  \subfigure[VO]{\includegraphics[width=0.134\textwidth]{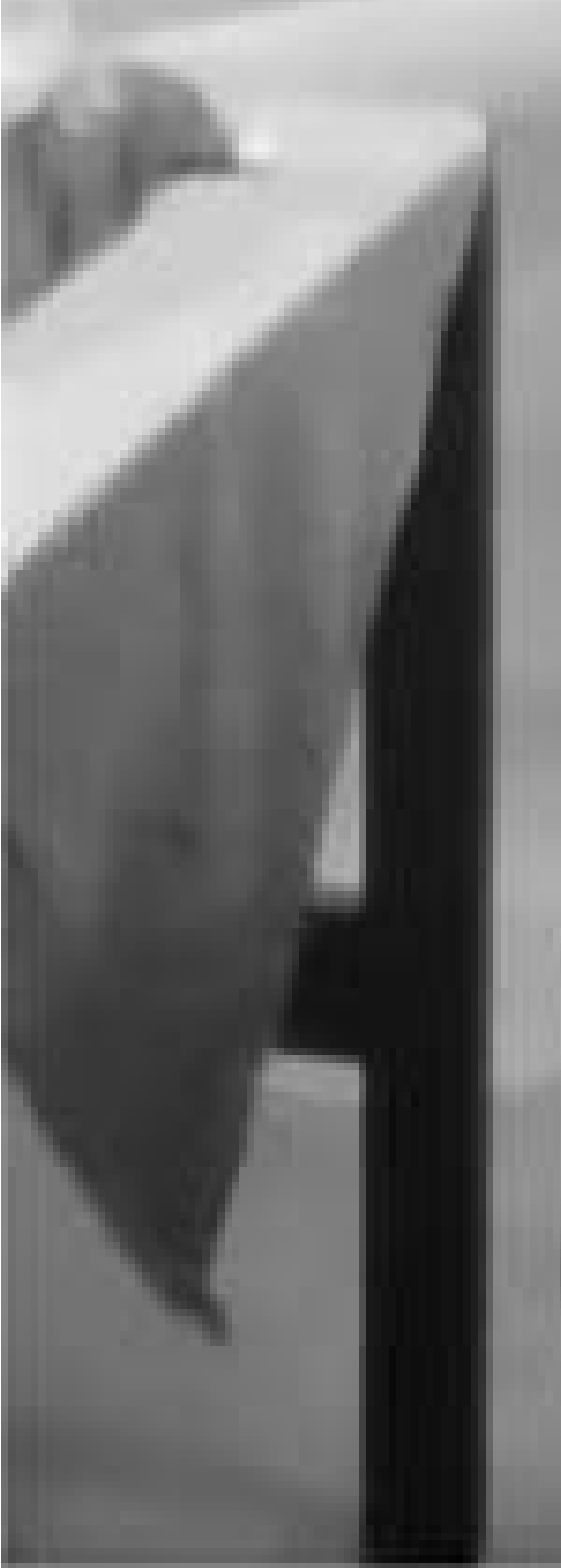}}
  \subfigure[SED]{\includegraphics[width=0.207\textwidth]{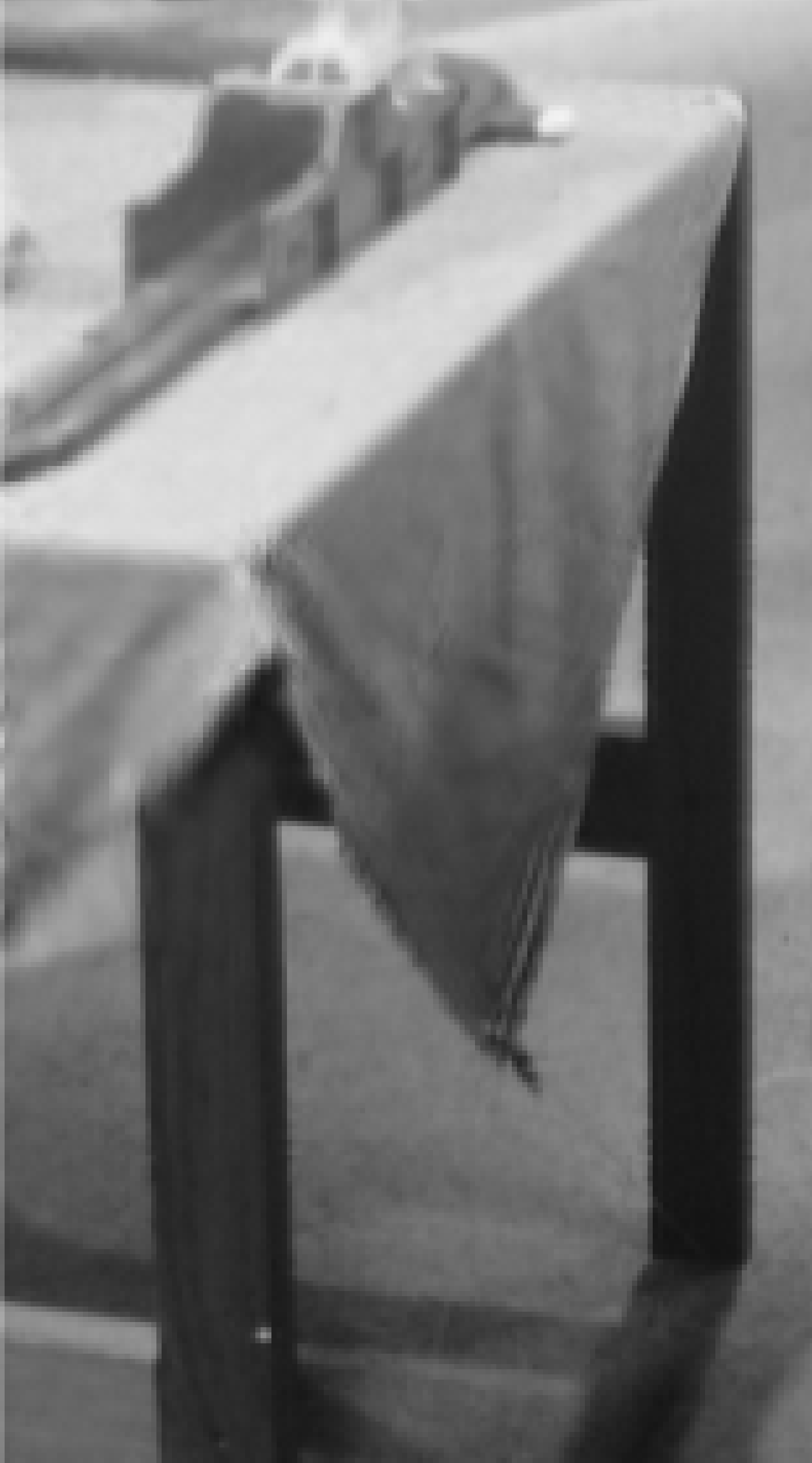}}
  \subfigure[TVG]{\includegraphics[width=0.207\textwidth]{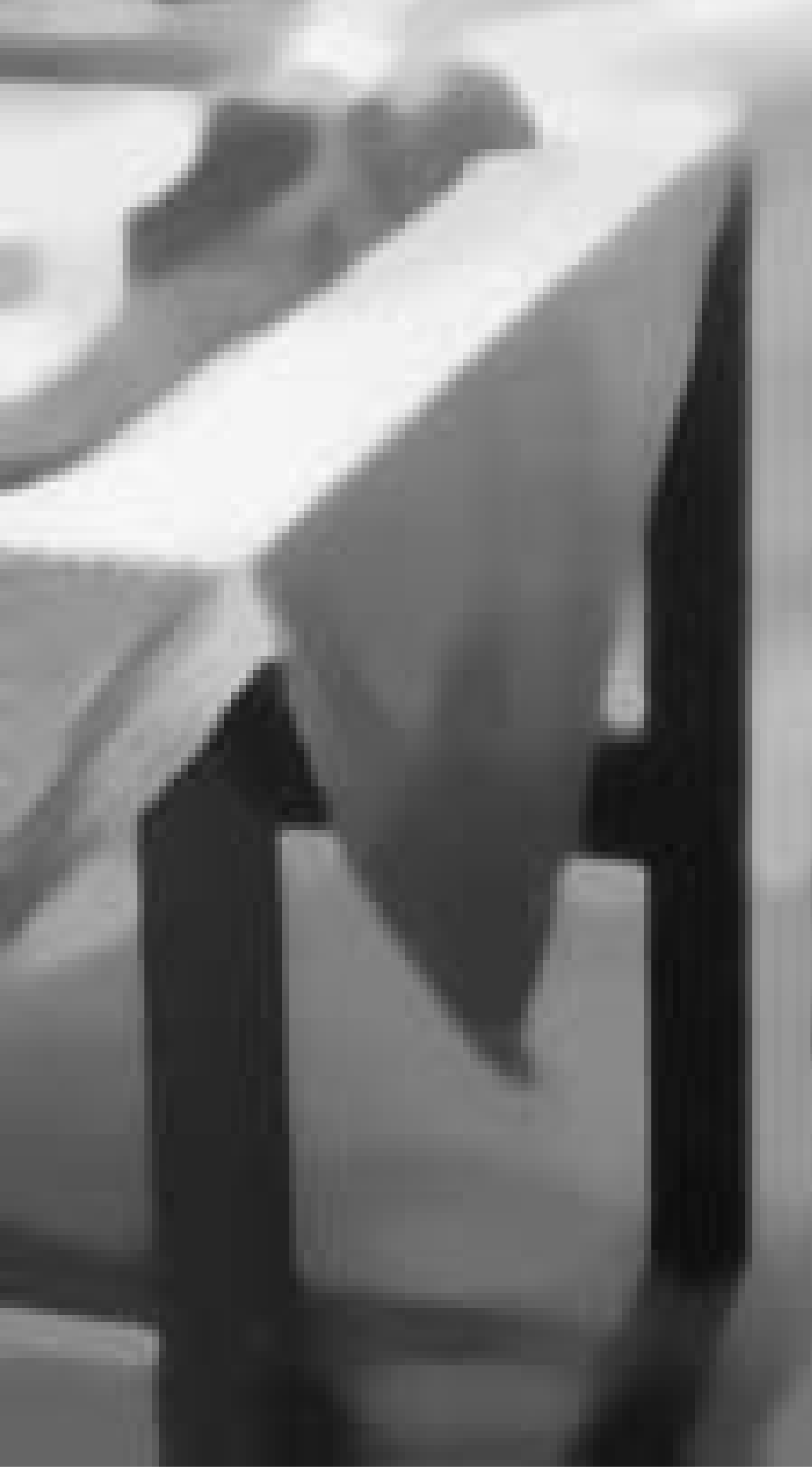}}
  \subfigure[DG3PD, $\delta = 10$]{\includegraphics[width=0.207\textwidth]{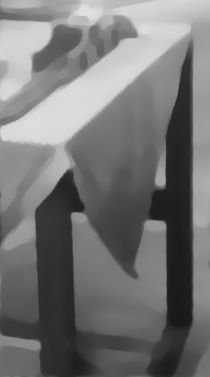}}  
  
  % v:
  \subfigure[]{\includegraphics[width=0.207\textwidth]{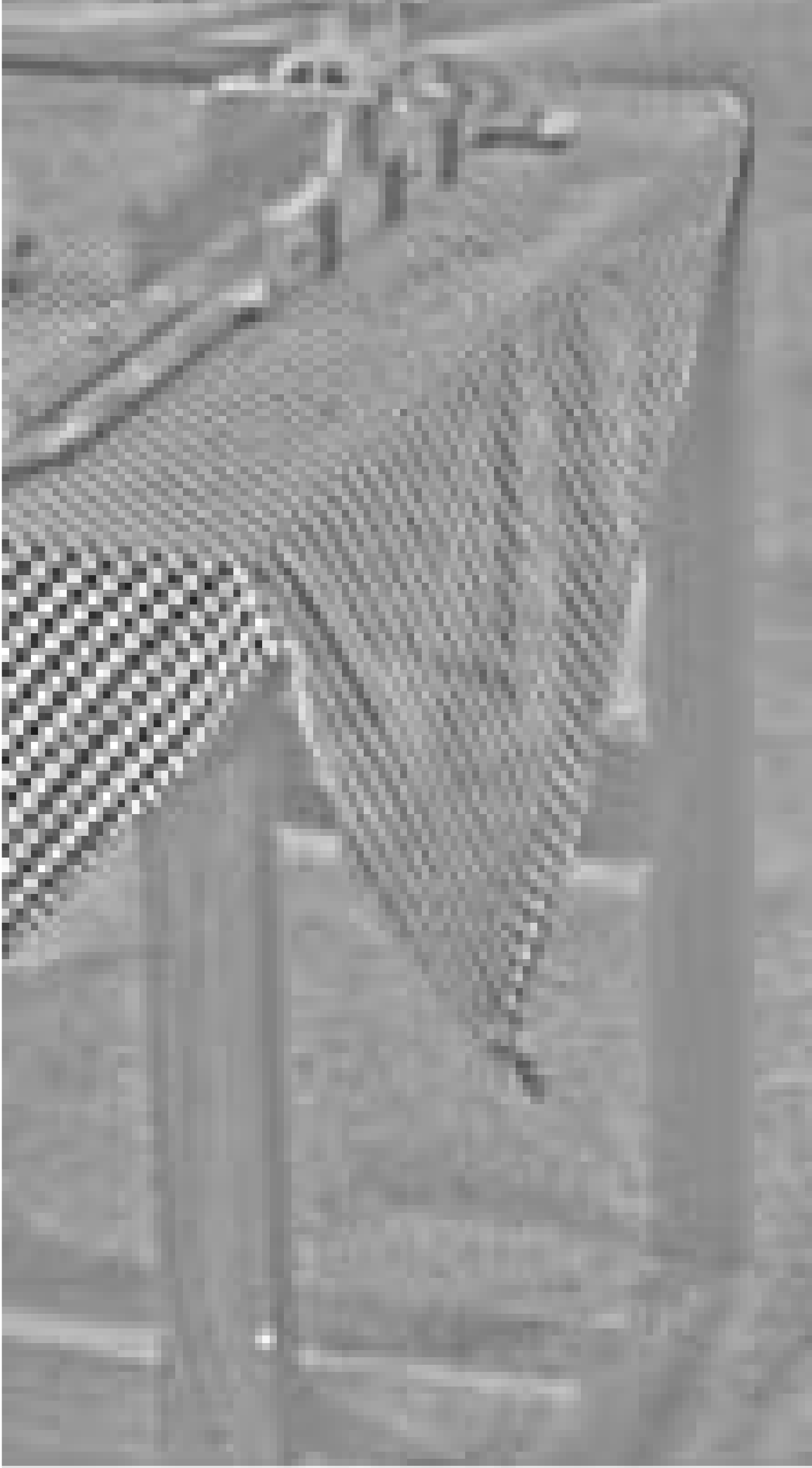}}  
  \subfigure[]{\includegraphics[width=0.134\textwidth]{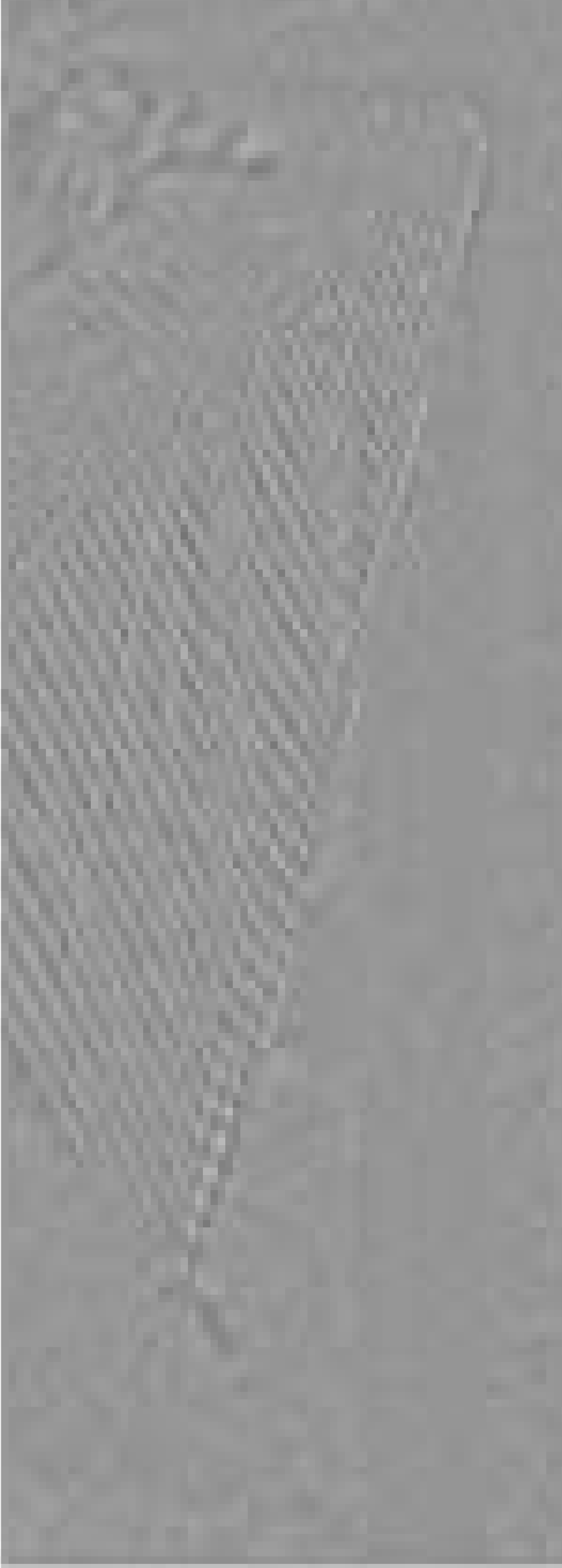}}
  \subfigure[]{\includegraphics[width=0.207\textwidth]{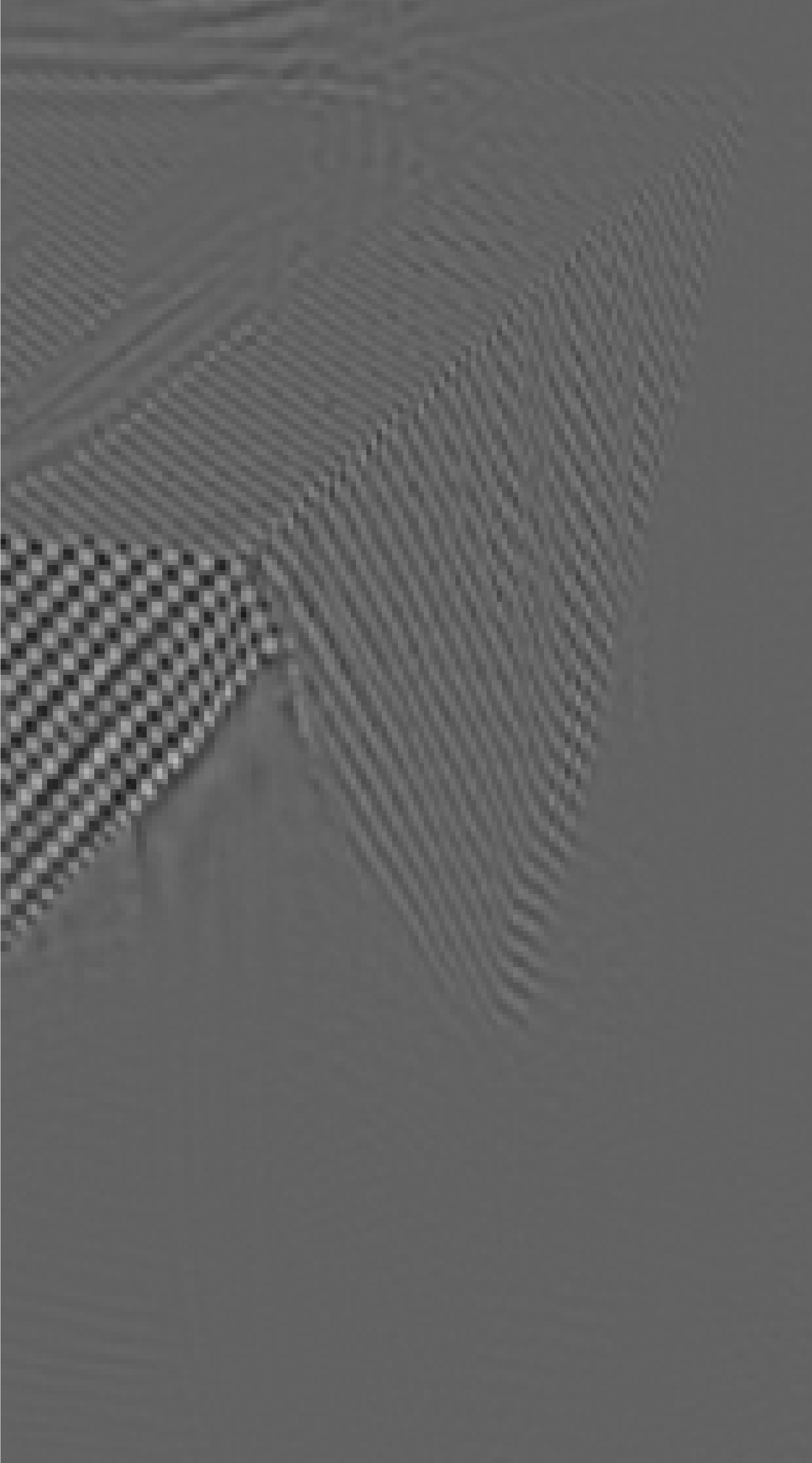}}
  \subfigure[]{\includegraphics[width=0.207\textwidth]{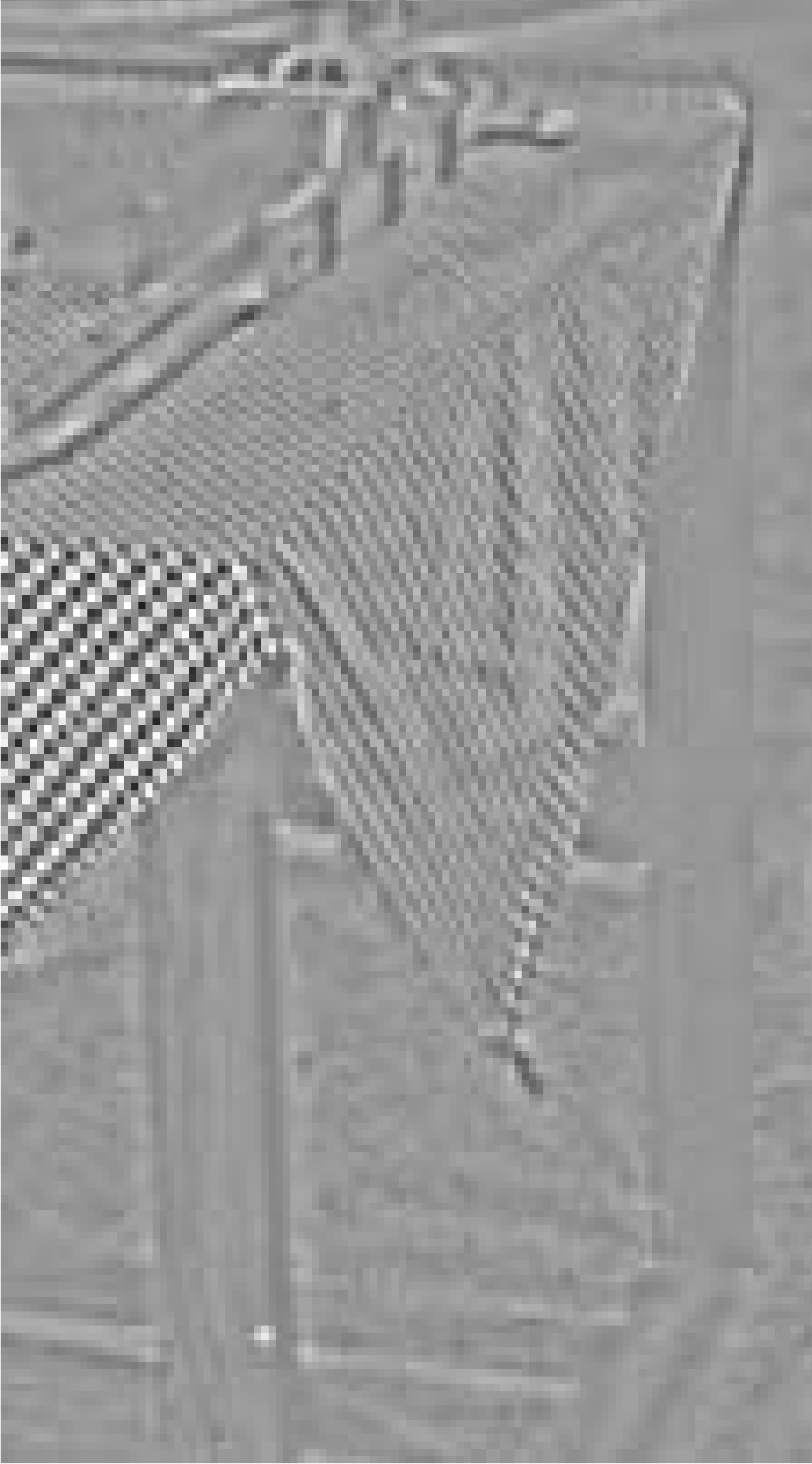}}  
  \subfigure[]{\includegraphics[width=0.207\textwidth]{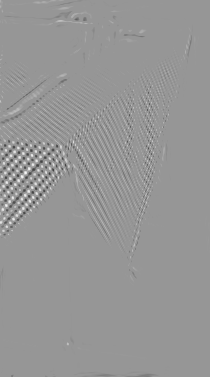}}
   
\caption{The images in the first row depict the comparison of the cartoon $\Bu$ for different methods 
         on the original highlighted region (cf. Figure \ref{fig:OriginalImage}(b)). 
         The images in the second row show their corresponding texture $\Bv$. 
        }
\label{fig:comparisonMethods:Barbara_crop2}
\end{center}
\end{figure}

% -----------------------

For better visibility of differences between the various models,
we show decomposition results for the ROF, VO, SED, TVG and DG3PD models 
for two magnified parts of the original image (cf. Figure \ref{fig:OriginalImage} (b, c))
in Figures \ref{fig:comparisonMethods:Barbara_crop1} and \ref{fig:comparisonMethods:Barbara_crop2}.

To the best of our knowledge, we observe two main differences between the compared models and 
the proposed DG3PD model:
\begin{itemize}
 \item Two-part decomposition instead of three-part decomposition.
 \item Quadratic penalty method (QPM) for solving the constrained minimization instead of ALM.
\end{itemize}

\textbf{Goal 1: cartoon $\Bu$.}
Regarding the cartoon $\Bu$, we observe that the VO and SED models still contain texture on the scarf 
(cf. Figure \ref{fig:comparisonMethods:Barbara_crop1} (b) and (c), respectively).
For the SED, the cartoon $\Bu$ is blurred and there are some small scale objects under the table 
(cf. Figure \ref{fig:comparisonMethods:Barbara_crop2} (c)).
Comparing all five methods, VO and SED are furthest away from achieving the first goal,
whereas ROF and TVG generate better cartoon images in terms of smoother surfaces without texture
and sharper boundaries. 
Recently, Schaeffer and Osher \cite{SchaefferOsher2013} suggested a two-part decomposition approach.
Note that the cartoon image by their decomposition which is shown in Figure 7 (a) of \cite{SchaefferOsher2013} 
also contains texture and does not meet the first goal.
The cartoon images produced by the DG3PD method come closest to the first goal,
cf. Figure \ref{fig:comparisonMethods:Barbara_crop1} (i,m) and Figure \ref{fig:comparisonMethods:Barbara_crop2} (e).

\textbf{Goal 2: texture $\Bv$.}
Concerning the texture $\Bv$, among the state-of-the-art methods, 
the decomposition by the SED model results in the sparsest texture
(cf. Figure \ref{fig:comparisonMethods:Barbara_crop1} (g) and Figure \ref{fig:comparisonMethods:Barbara_crop2} (h)),
while the texture images of the ROF, VO and TVG have more coefficients different from zero.
In addition, the texture component obtained by the ROF model also contains some geometry information
which should have been assigned to the cartoon component,
see Figure \ref{fig:comparisonMethods:Barbara_crop1} (e) 
and Figure \ref{fig:comparisonMethods:Barbara_crop2} (f).
The DG3PD model yields an even sparser texture than the SED model due to the $\norm{\Bv}_{\ell_1}$ 
in the minimization (\ref{eq:DTV-DG3part}), see the binarized versions with threshold ``0'' for visualization 
in Figure \ref{fig:comparisonMethods:Barbara_crop1} (o) or Figure \ref{fig:DG3PD:barbara} (g).

\begin{figure}
\begin{center}  
  
  % ROF, VeseOsher, G3PD (Iter = 20), G3PD (Iter = 60):
  \subfigure[$\Bf$]{\includegraphics[width=0.11\textwidth]{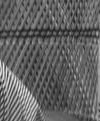}}  
  \subfigure[VO: $\Bu + \Bv$]{\includegraphics[width=0.11\textwidth]{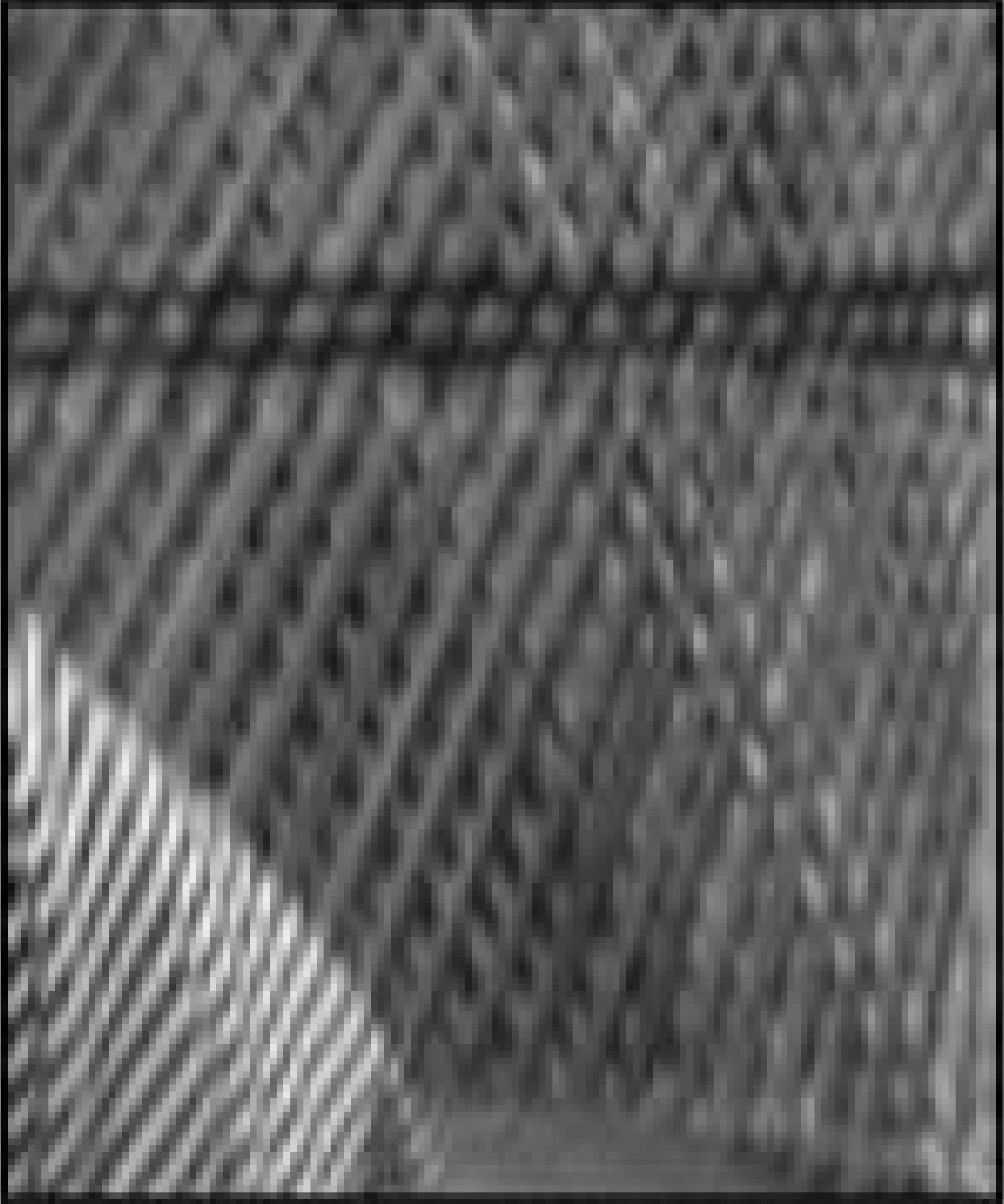}}
  \subfigure[DG3PD: $\Bu + \Bv + \Beps$]
  {\includegraphics[width=0.11\textwidth]{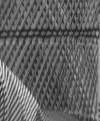}}     
   \subfigure[DG3PD: $\Bu + \Bv + \Beps$]
  {\includegraphics[width=0.11\textwidth]{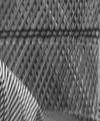}}         
  \subfigure[ROF: $\Bu \,, \lambda_\text{ROF} = 0.1$]{\includegraphics[width=0.11\textwidth]{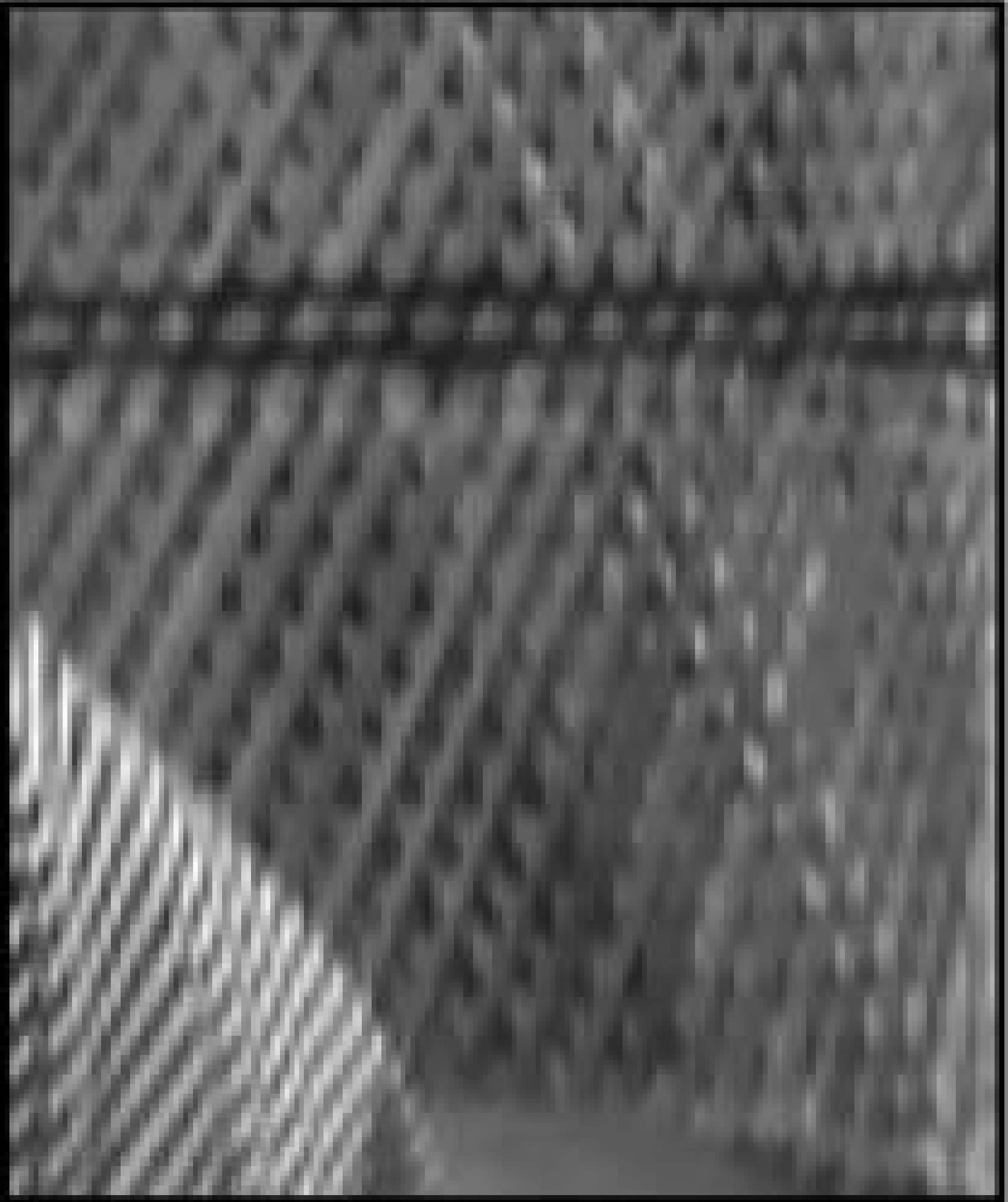}}  
  \subfigure[VO: $\Bu$]{\includegraphics[width=0.11\textwidth]{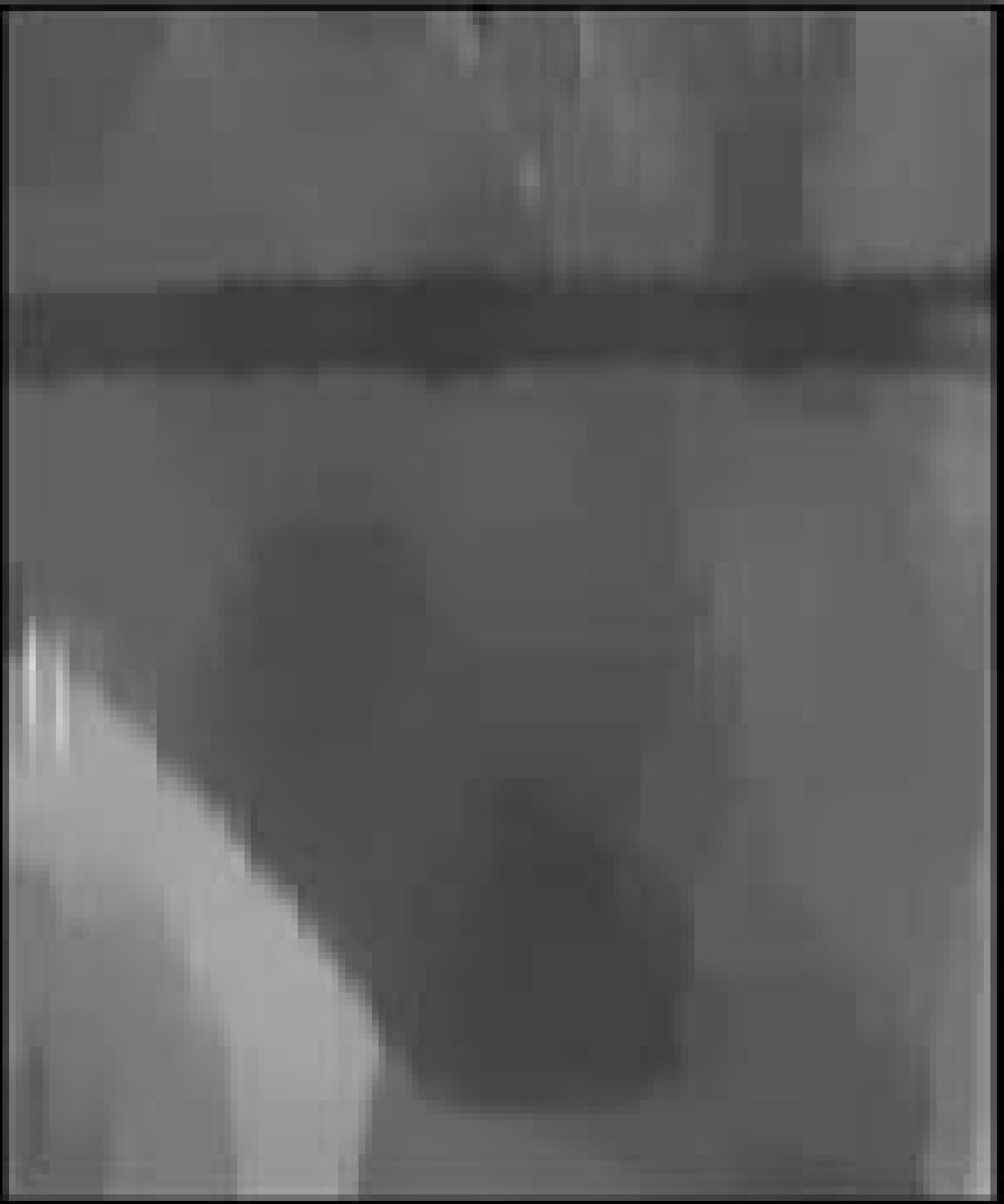}}  
  \subfigure[DG3PD: $\Bu$]
  {\includegraphics[width=0.11\textwidth]{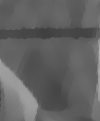}}    
  \subfigure[DG3PD: $\Bu$]
  {\includegraphics[width=0.11\textwidth]{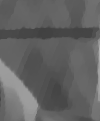}}  
  
  % v:  
  \subfigure[ROF: $150 + (\Bf - \Bu)$]{\includegraphics[width=0.11\textwidth]{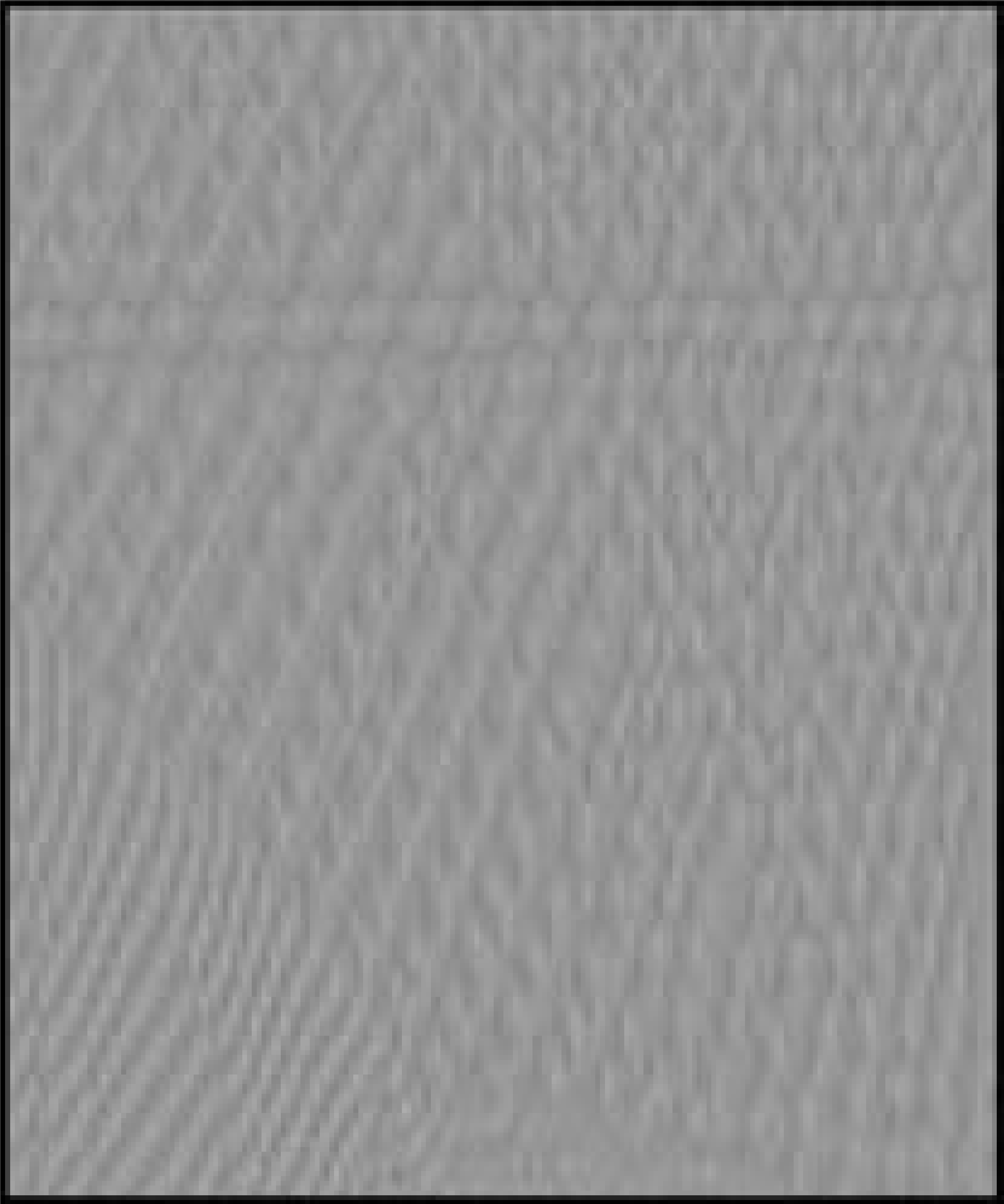}}
  \subfigure[VO: $150 + \Bv$]{\includegraphics[width=0.11\textwidth]{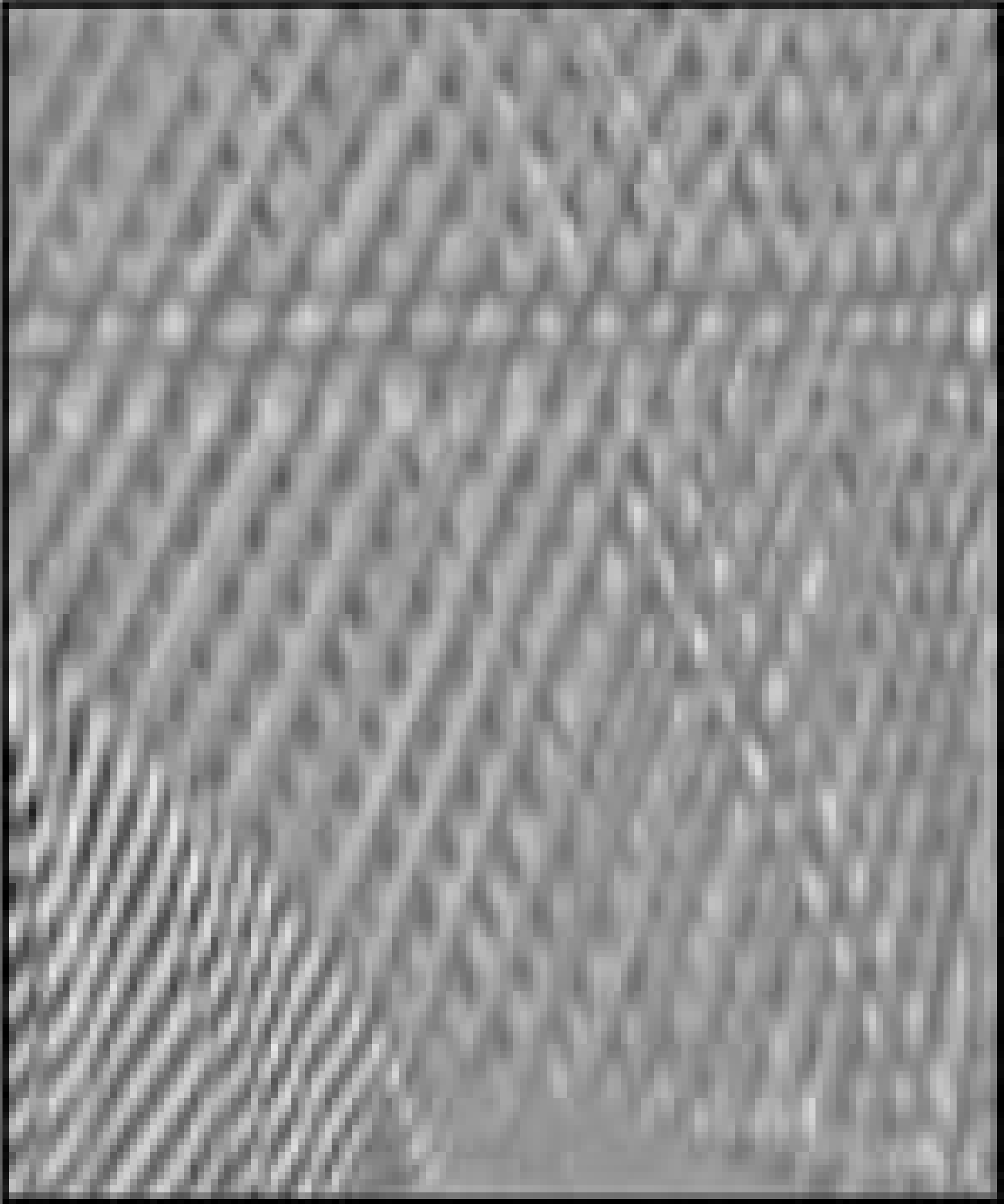}} 
  \subfigure[DG3PD: $150 + \Bv$]
  {\includegraphics[width=0.11\textwidth]{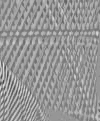}}    
  \subfigure[DG3PD: $150 + \Bv$]
  {\includegraphics[width=0.11\textwidth]{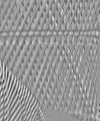}}    
    \subfigure[DG3PD: $\delta = 0$]
    {\includegraphics[width=0.11\textwidth,height=0.1331\textwidth]{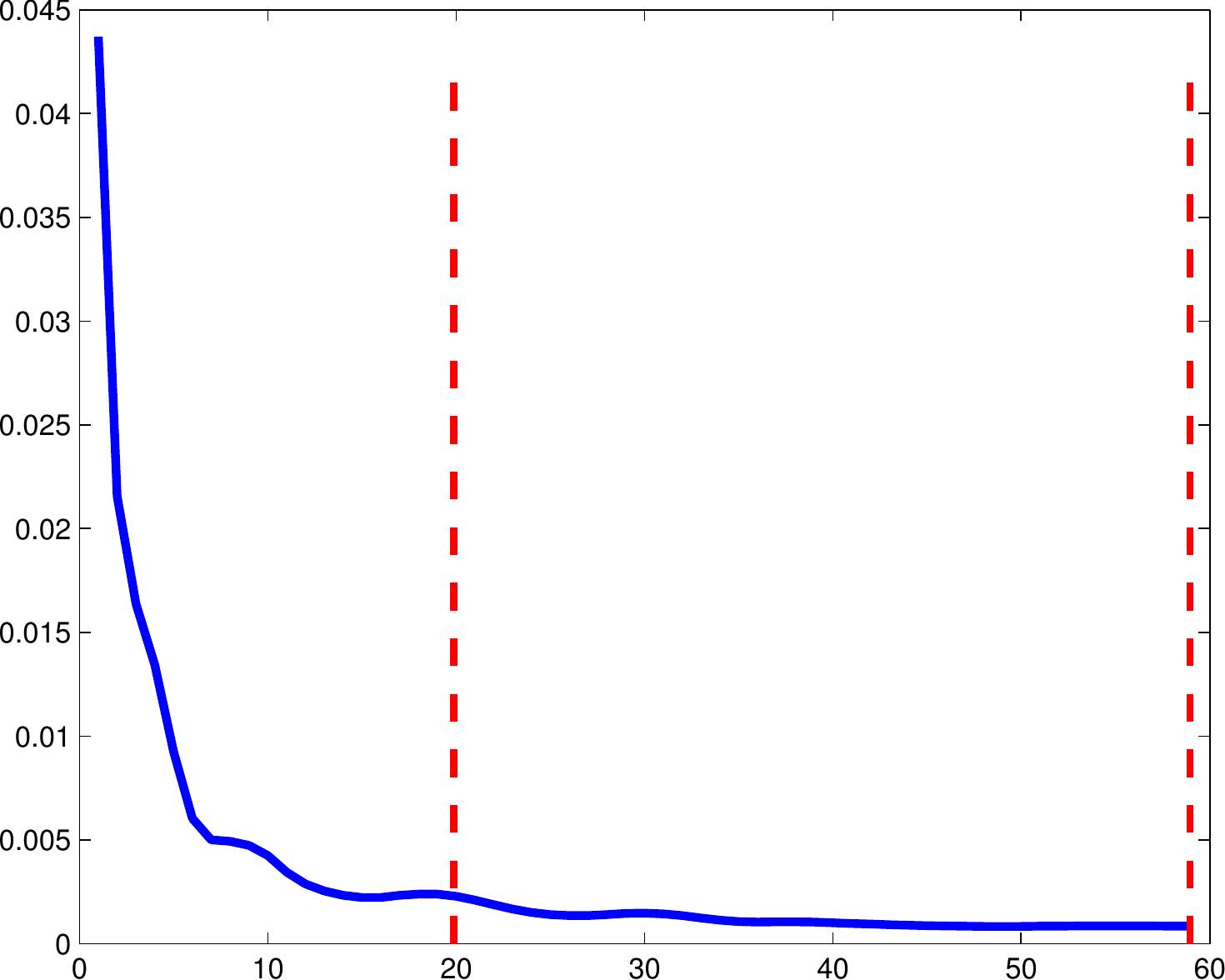}} 
    \subfigure[$150 + (\Bf - \Bu - \Bv)$]{\includegraphics[width=0.11\textwidth]{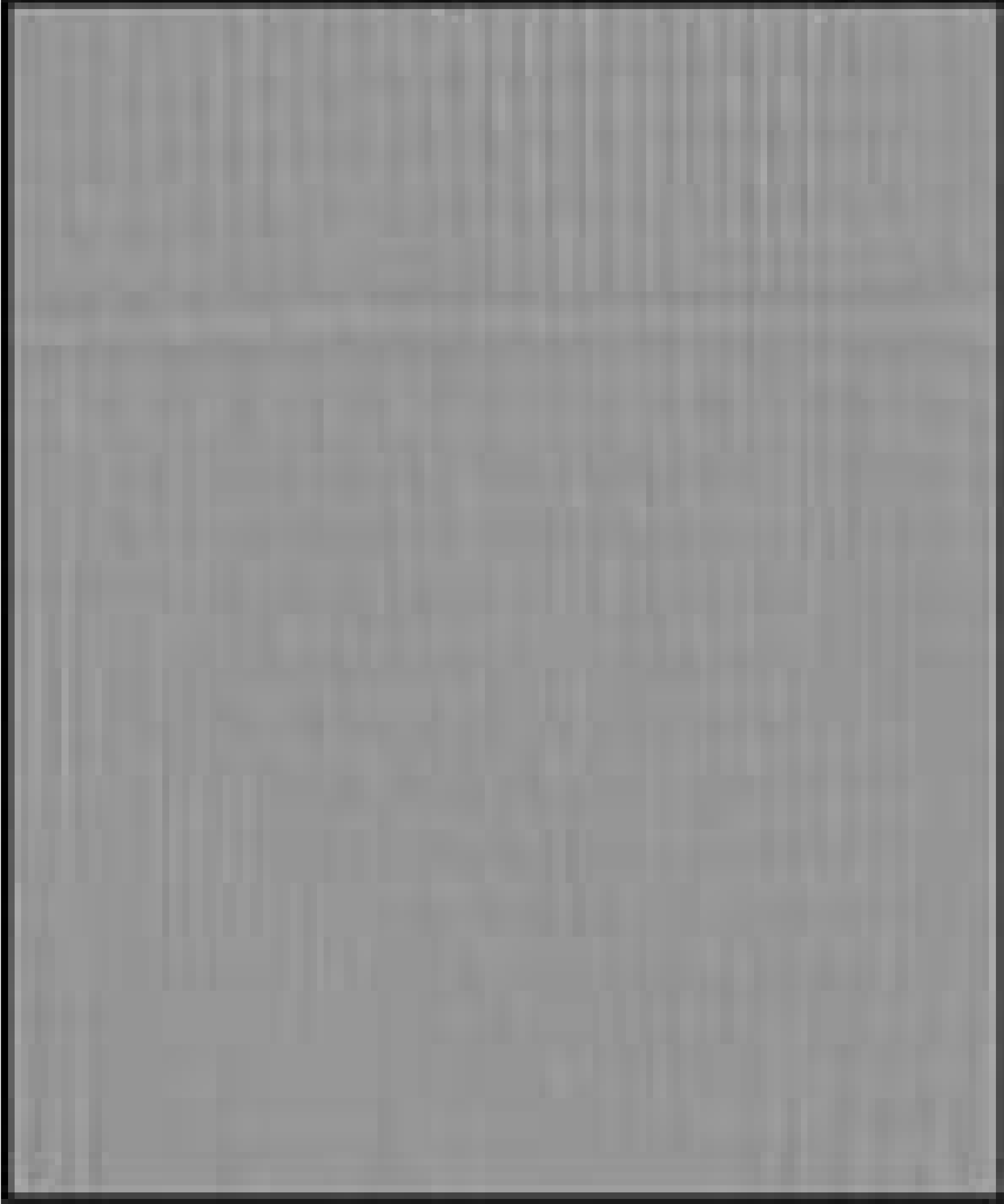}} % VO    
    \subfigure[$150 + (\Bf - \Bu - \Bv - \Beps)$]
    {\includegraphics[width=0.11\textwidth]{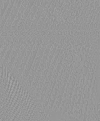}}  % DG3PD    
    \subfigure[$150 + (\Bf - \Bu - \Bv - \Beps)$]    % DG3PD
    {\includegraphics[width=0.11\textwidth]{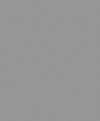}}      
  \caption{Comparison of QPM and ALM for the ROF, VO and DG3PD models:
           The images for ROF and VO models are obtained from \cite{VeseOsher2003}.
           The parameters for the DG3PD model are $\delta = 0$ and 20 iterations for the third and 60 for the fourth column,
					 and the other parameters are the same as in Figure \ref{fig:DG3PD:barbara}.           
           The relative error (y-axis) versus the number of iterations (x-axis) is illustrated in (m).
           The first row shows that the VO and the DG3PD models can achieve good reconstructed images, cf. (b, c, d),
           in comparison with the original magnified image (a).
           As mentioned in \cite{VeseOsher2003}, the error image from the VO model (n)
           contains much less geometry and texture than the one from the ROF model.
           However, the error image from our model is much further reduced in comparison to the VO model.
           After 20 iterations some pieces of information still remain in the error image, cf. (o). 
           As the number of iterations increases, the error numerically tends to $\mathbf 0$, cf. (p) after 60 iterations.
          }
\label{fig:comparison:Barbara:penaltyALM}
\end{center}
\end{figure}

\begin{figure}
\begin{center}  
  %\subfigure[$\Bf$]{\includegraphics[width=0.19\textwidth,height=0.19\textwidth]{tm20_1_1.png}}
  %\subfigure[$\Bu$]{\includegraphics[width=0.19\textwidth,height=0.19\textwidth]{tm20_1_1_20nonoise_u.png}}  
  %\subfigure[$150 + \Bv$]{\includegraphics[width=0.19\textwidth,height=0.19\textwidth]{tm20_1_1_20nonoise_v.png}}  
  %\subfigure[$\Bv_\text{bin} = \Bv > 0$]{\includegraphics[width=0.19\textwidth,height=0.19\textwidth]{tm20_1_1_20nonoise_vbin.png}}
  %\subfigure[$150 + \Beps$]{\includegraphics[width=0.19\textwidth,height=0.19\textwidth]{tm20_1_1_20nonoise_epsilon.png}}  
	
	\subfigure[$\Bf$]{\includegraphics[width=0.16\textwidth,height=0.16\textwidth]{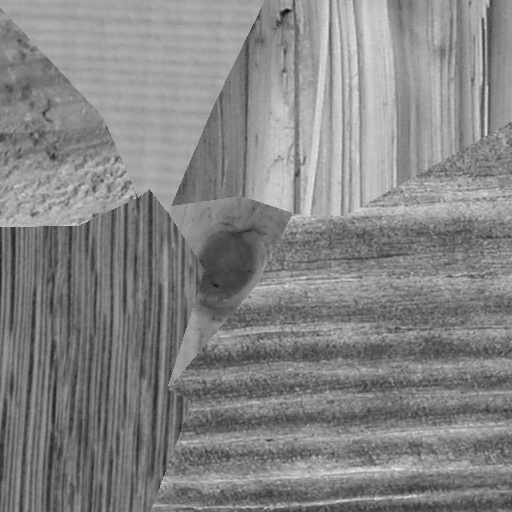}}
  \subfigure[$\Bu$]{\includegraphics[width=0.16\textwidth,height=0.16\textwidth]{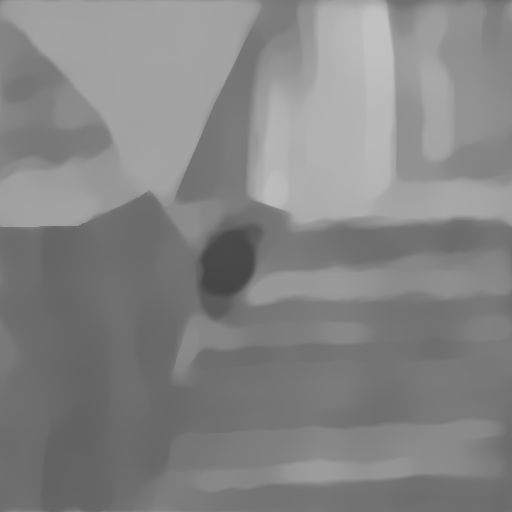}}  
  \subfigure[$150 + \Bv$]{\includegraphics[width=0.16\textwidth,height=0.16\textwidth]{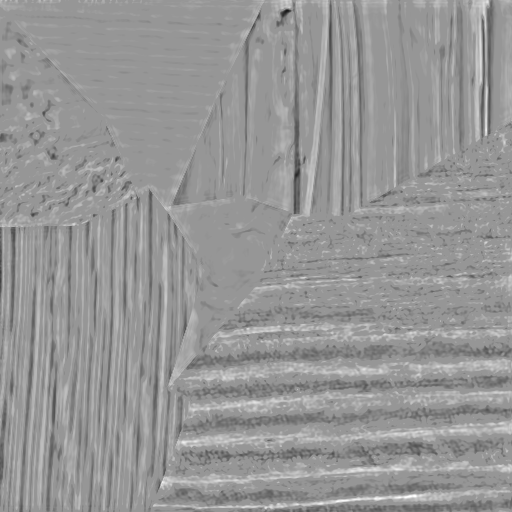}}  
  \subfigure[$\Bv_\text{bin} = \Bv > 0$]{\includegraphics[width=0.16\textwidth,height=0.16\textwidth]{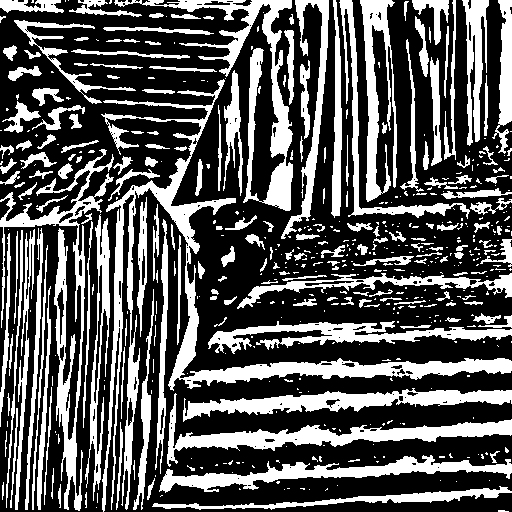}}
  \subfigure[$150 + \Beps$]{\includegraphics[width=0.16\textwidth,height=0.16\textwidth]{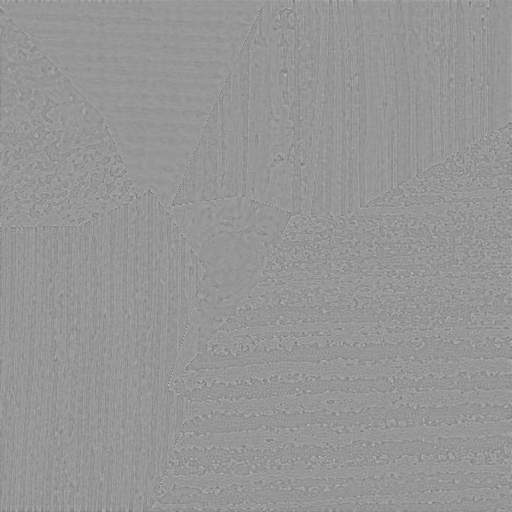}}   \\
  
  \subfigure[ALM]{\includegraphics[width=0.16\textwidth,height=0.16\textwidth]{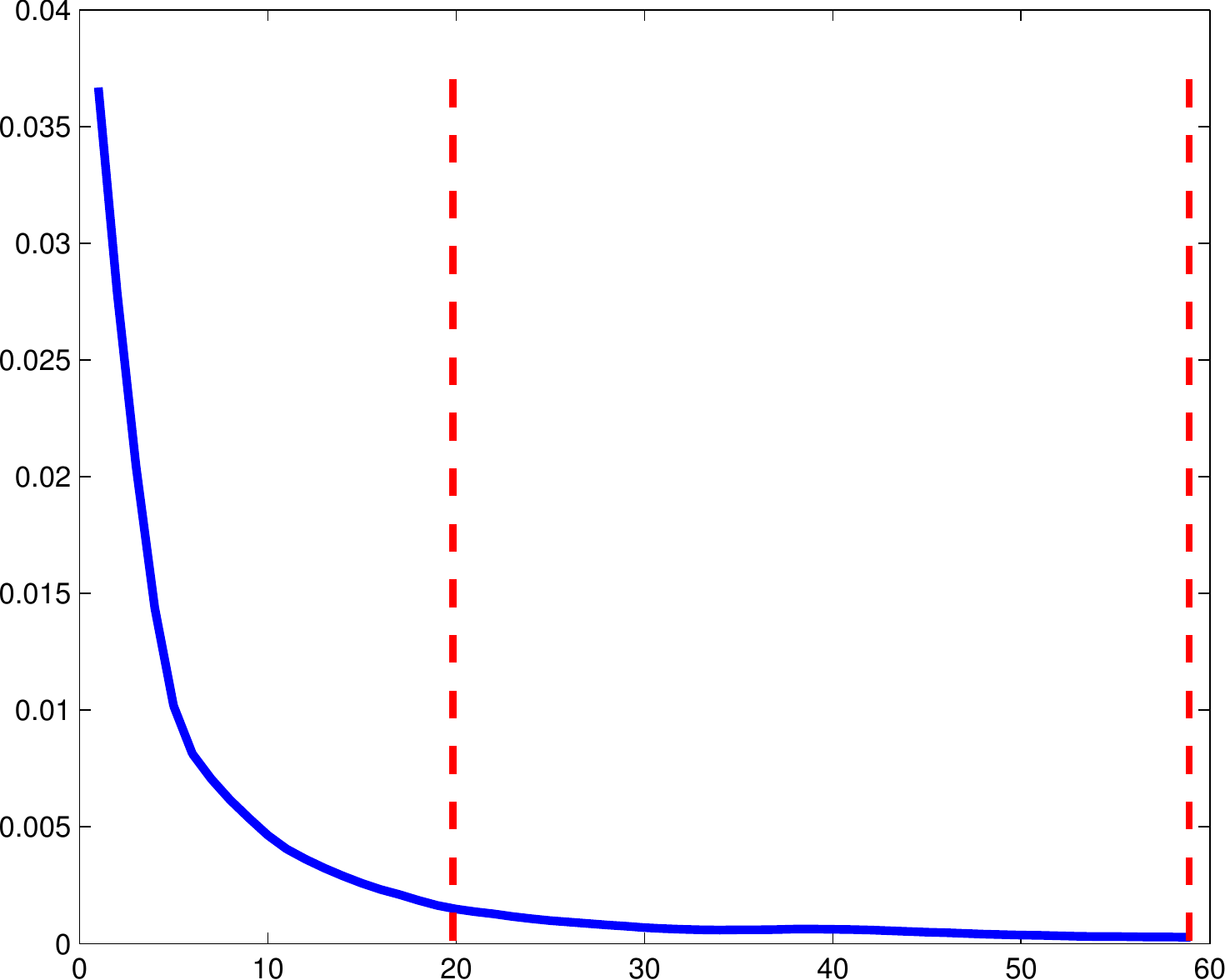}}
  \subfigure[$\Bu + \Bv + \Beps$]{\includegraphics[width=0.16\textwidth,height=0.16\textwidth]{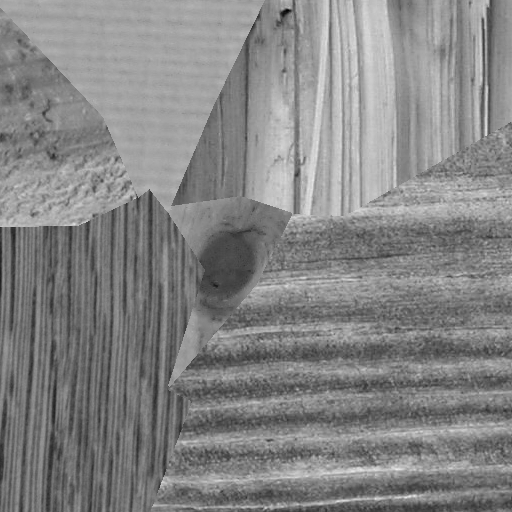}}  
  \subfigure[$150 + (\Bf - \Bu - \Bv - \Beps)$, Iter = 20]{\includegraphics[width=0.16\textwidth,height=0.16\textwidth]{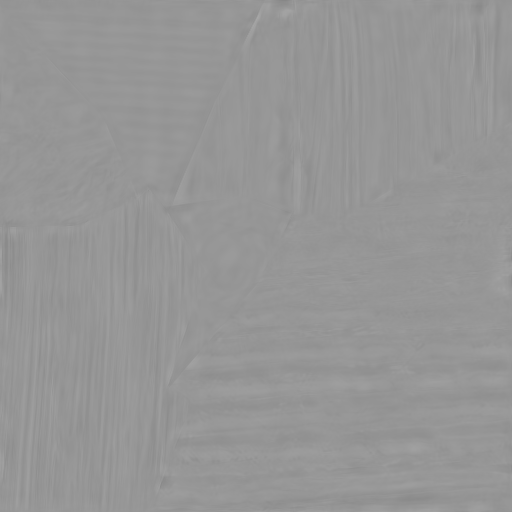}}
  \subfigure[$150 + (\Bf - \Bu - \Bv - \Beps)$, Iter = 60]{\includegraphics[width=0.16\textwidth,height=0.16\textwidth]{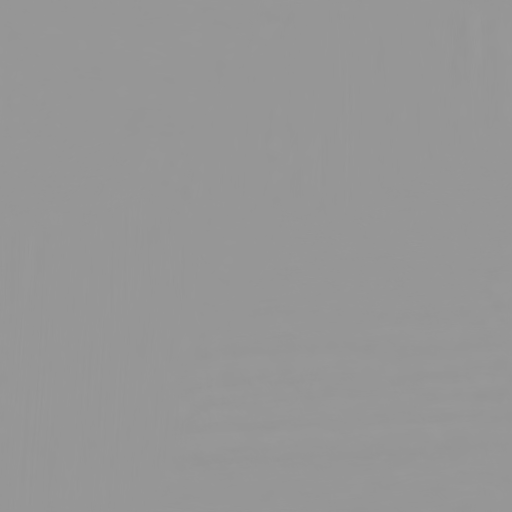}}
  \subfigure[$\Bu$]{\includegraphics[width=0.16\textwidth,height=0.16\textwidth]{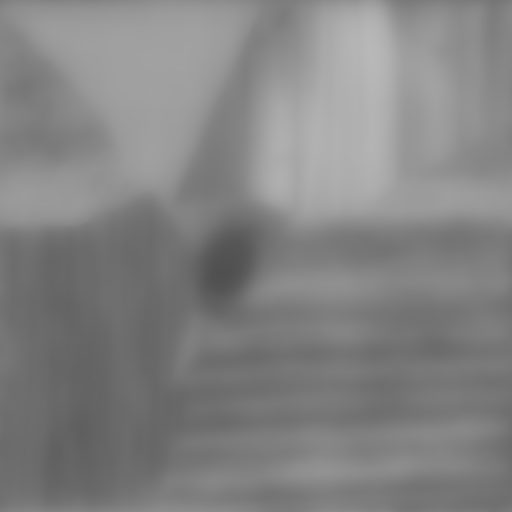}}  
  \subfigure[$150 + \Bv$]{\includegraphics[width=0.16\textwidth,height=0.16\textwidth]{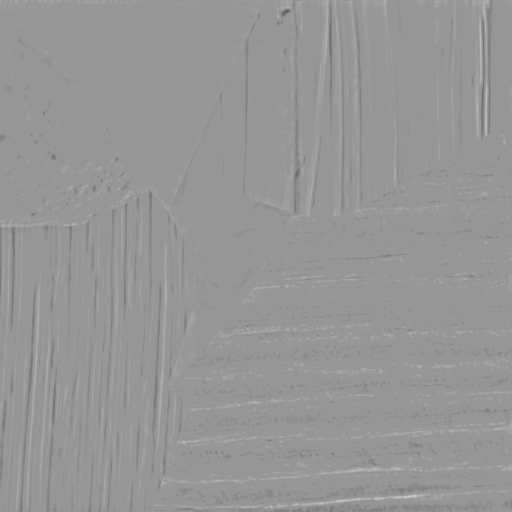}}  
	
  \subfigure[$\Bv_\text{bin} = \Bv > 0$]{\includegraphics[width=0.16\textwidth,height=0.16\textwidth]{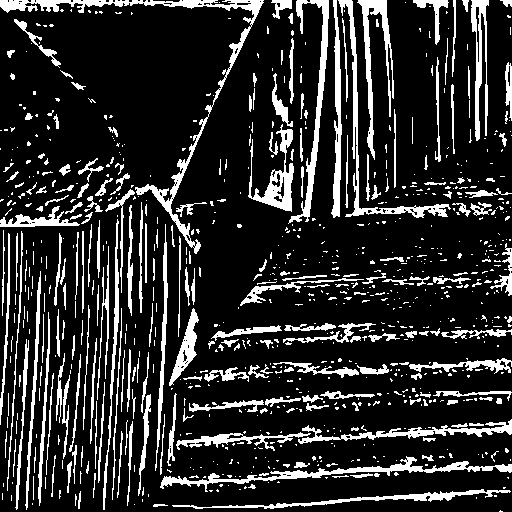}}
  \subfigure[$150 + \Beps$]{\includegraphics[width=0.16\textwidth,height=0.16\textwidth]{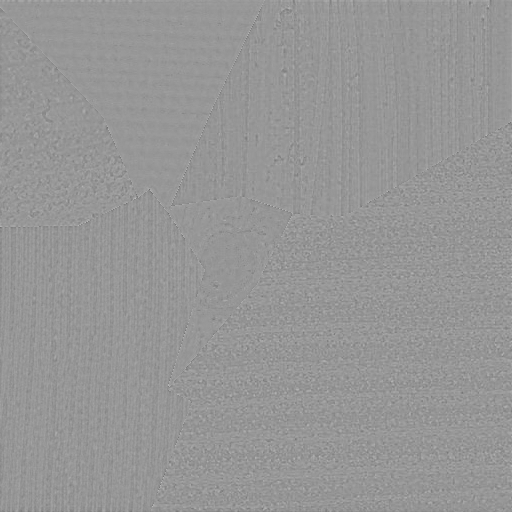}}    
  \subfigure[penalty]{\includegraphics[width=0.16\textwidth,height=0.16\textwidth]{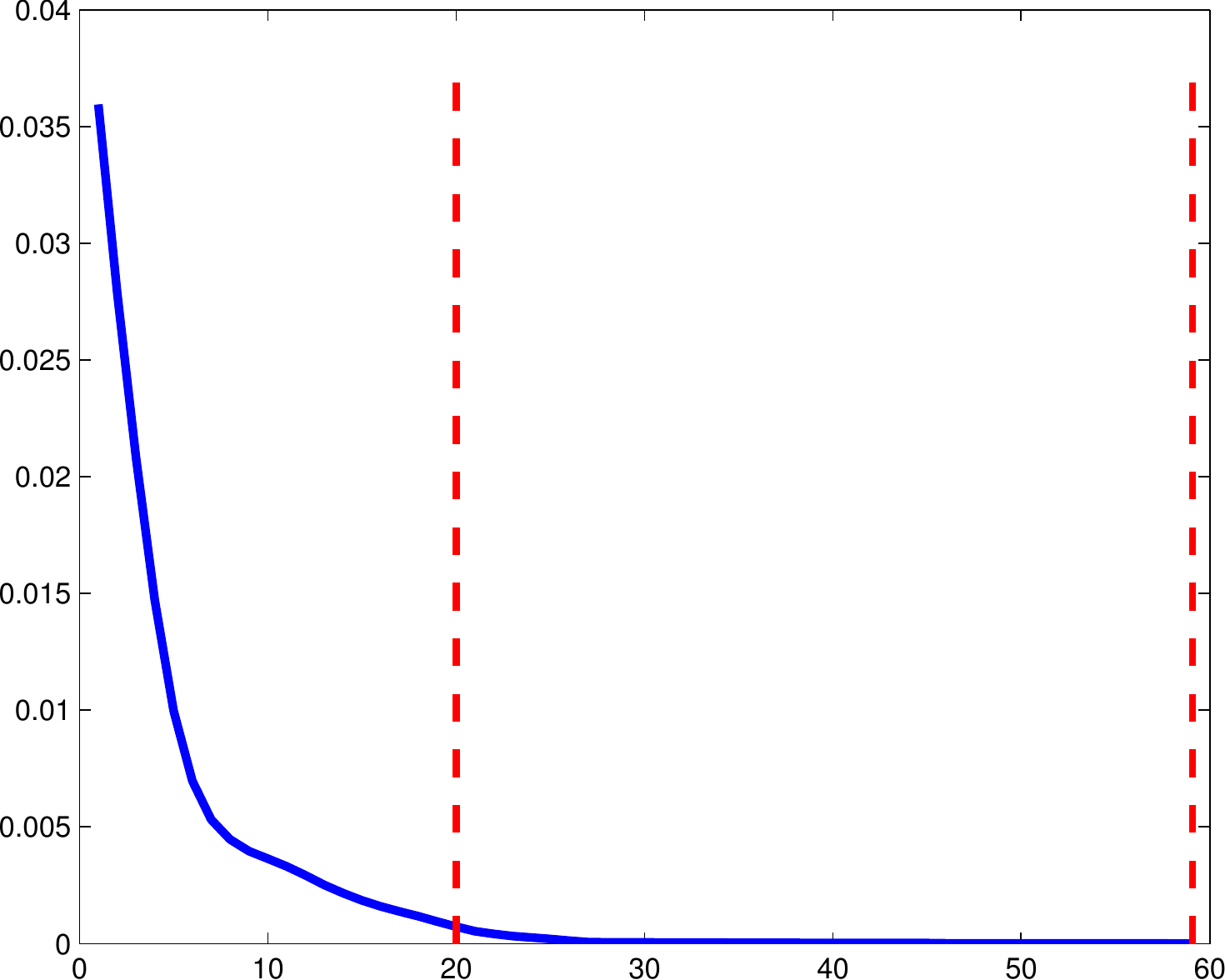}}
  \subfigure[$\Bu + \Bv + \Beps$]{\includegraphics[width=0.16\textwidth,height=0.16\textwidth]{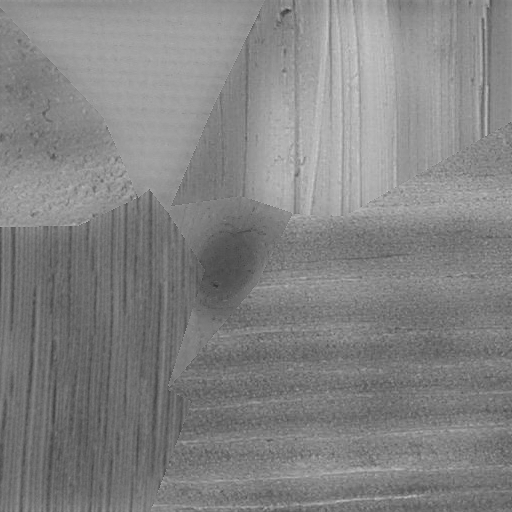}}  
  \subfigure[$150 + (\Bf - \Bu - \Bv - \Beps)$, Iter = 20]{\includegraphics[width=0.16\textwidth,height=0.16\textwidth]{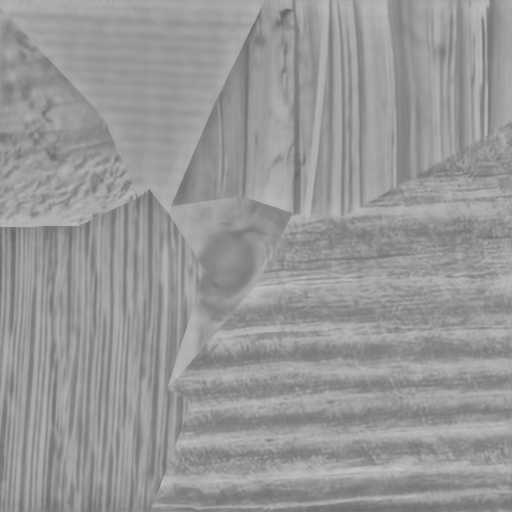}}
  \subfigure[$150 + (\Bf - \Bu - \Bv - \Beps)$, Iter = 60]{\includegraphics[width=0.16\textwidth,height=0.16\textwidth]{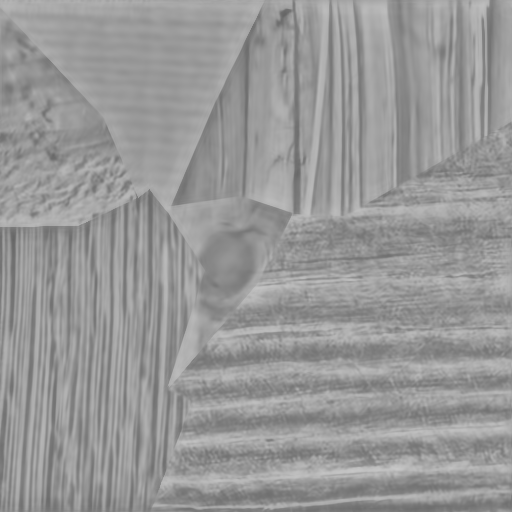}}  
  
\caption{The comparison of ALM (b-i) and QPM (j-q) for the DG3PD model 
         with the same parameters and 20 iterations:
         $\beta_4 = 0.025 \,, \theta = 0.9 \,, c_1 = 10 \,, c_2 = 1.3 \,, c_{\mu_1} = c_{\mu_2} = 0.03
          \,, \gamma = 1 \,, S = L = 32 \,, \delta = 5$.
         We observe the error image of QPM still contains geometry and texture information,
         cf. after 20 iterations in (p) and after 60 in (q).
         Using QPM for constrained minimization is similar to decomposing
         the original image into four parts, namely cartoon (j), texture (k), residual (m)
         and error (p). 
         The amount of information in the error image by QPM strongly depends on 
         the choice of parameters $(\beta_1 \,, \beta_2 \,, \beta_3 \,, \beta_4)$.
         However, for ALM, the updated Lagrange multipliers 
         $(\boldsymbol{\lambda_1 \,, \lambda_2 \,, \lambda_3 \,, \lambda_4})$
         compensate for the choice of $(\beta_1 \,, \beta_2 \,, \beta_3 \,, \beta_4)$.
         Thus, the error numerically tends to $\mathbf 0$ as the number of iterations increases,
         cf. the error image after 20 iterations in (h) and after 60 iterations in (i). 
         }
\label{fig:comparison:tm20_1_1:penaltyALM}
\end{center}
\end{figure}

\textbf{Goal 3: reconstruction by summation of all components.}
Figures \ref{fig:comparison:Barbara:penaltyALM} and \ref{fig:comparison:tm20_1_1:penaltyALM}
illustrate the effects of QPM and ALM.
The decomposition by the ROF model results in a relatively large error $(\Bf - \Bu)$
which contains geometry and texture information, 
see \cite{VeseOsher2003} and Figure \ref{fig:comparison:Barbara:penaltyALM} (i).
In the VO model, the error $(\Bf - \Bu - \Bv)$ is reduced in comparison to the ROF model,
but some information still remains in the error image, see Figure \ref{fig:comparison:Barbara:penaltyALM} (n).
In case of the DG3PD model with the ALM based approach for solving the constrained minimization,
the error $(\Bf - \Bu - \Bv - \Beps)$ is significantly reduced and 
numerically the error tends to $\mathbf 0$ as the number of iterations increases.
For a comparison to ROF and VO using the same detail,
see Figure \ref{fig:comparison:Barbara:penaltyALM} (o) for a visualization of the error after 20 iterations 
and (p) after 60 iterations.
The error for the whole image after 20 and 60 iterations is displayed 
in Figure \ref{fig:DG3PD:barbara} (j) and (k), respectively.
To the best of our knowledge, this effect can be explained by using ALM for solving the constrained minimization
instead of QPM.
For more details about QPM and ALM, we refer the reader to 
\cite{Courant1943},
\cite{BoydParikhChuPeleatoEckstein2011}, and
Chapter 3 in \cite{FortinGlowinski1983}.

\begin{figure}
\begin{center} 
	\subfigure[AC $\Bu$]{\includegraphics[width=0.244\textwidth]{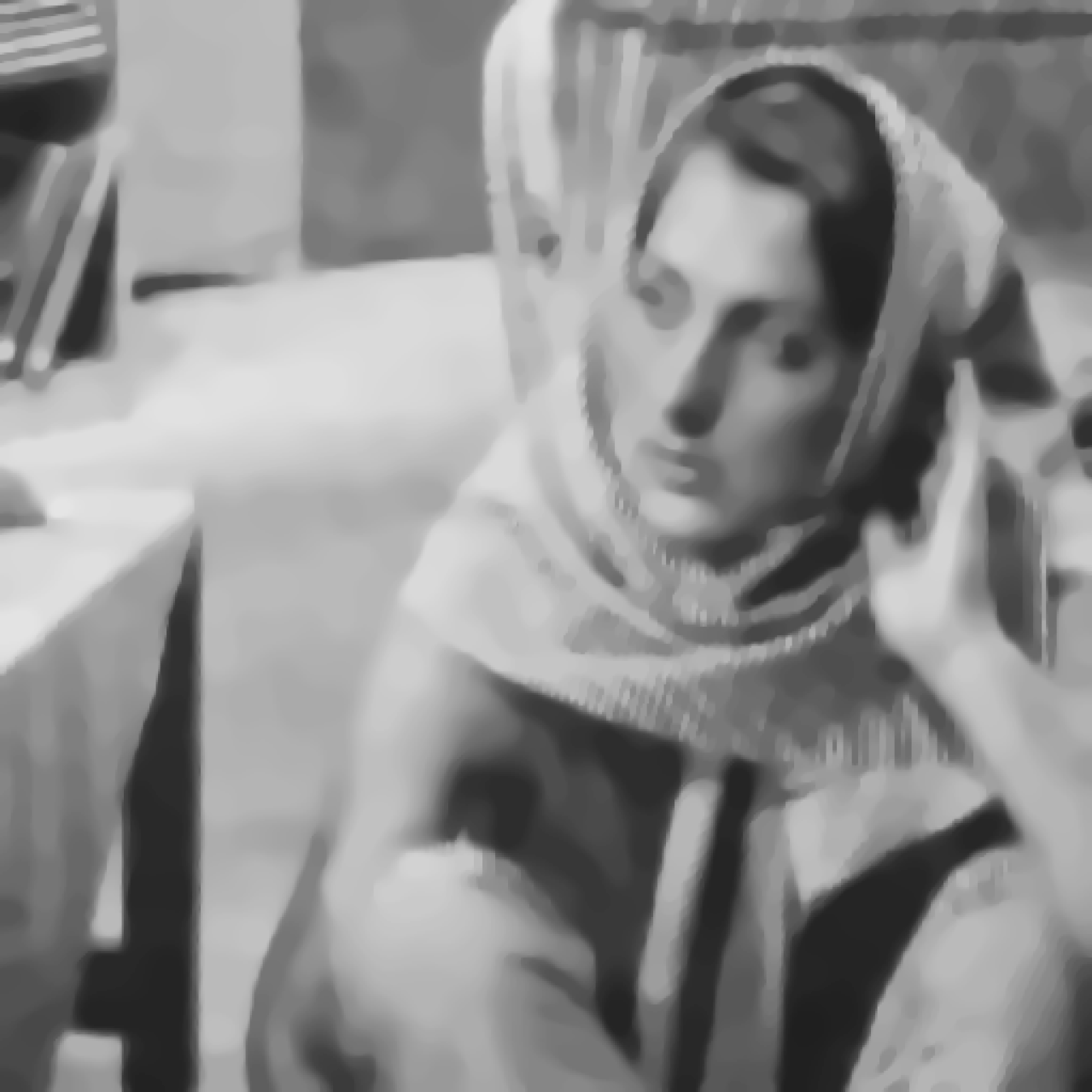}}
	\subfigure[AC $\Bv$]{\includegraphics[width=0.244\textwidth]{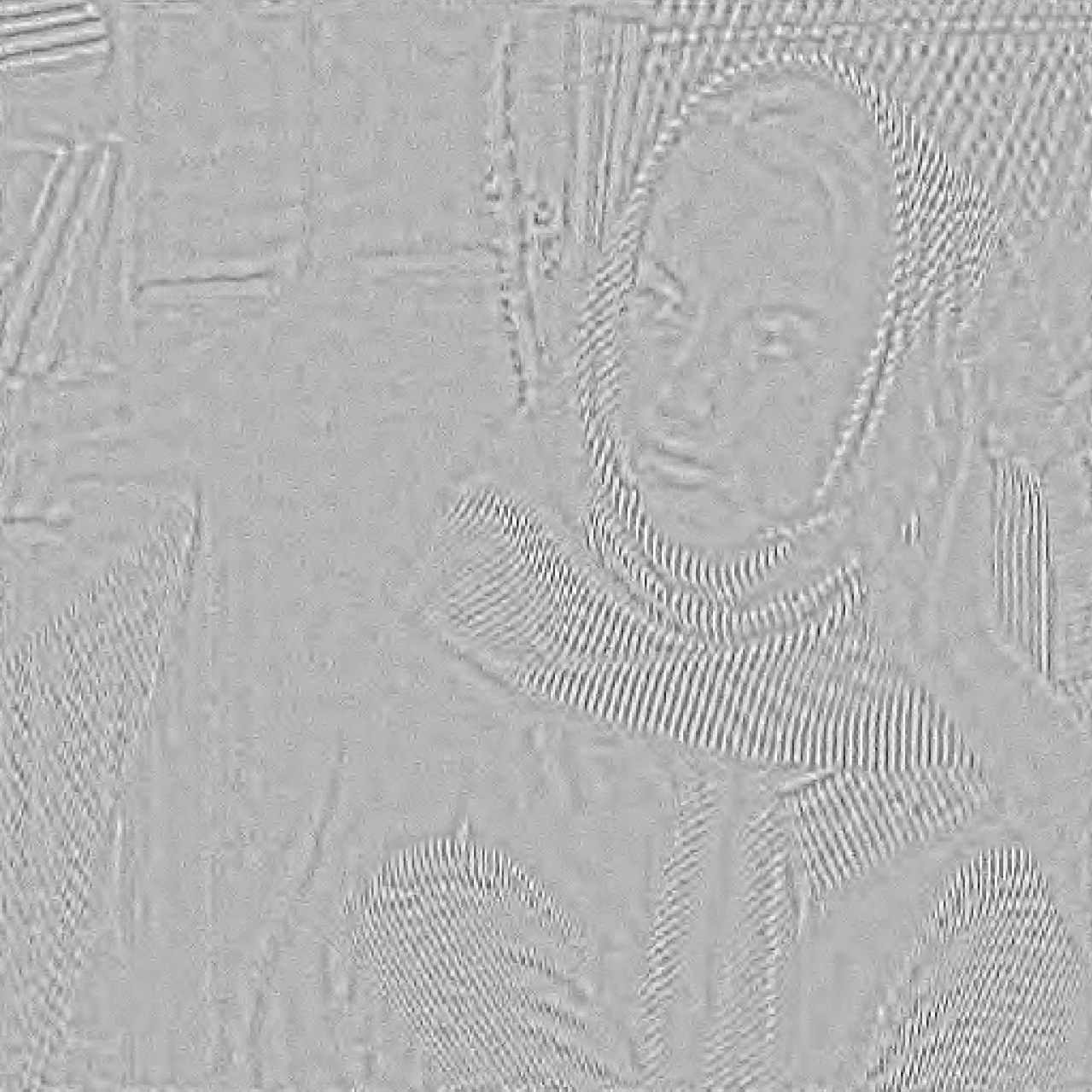}}
	\subfigure[AC $\Beps$]{\includegraphics[width=0.244\textwidth]{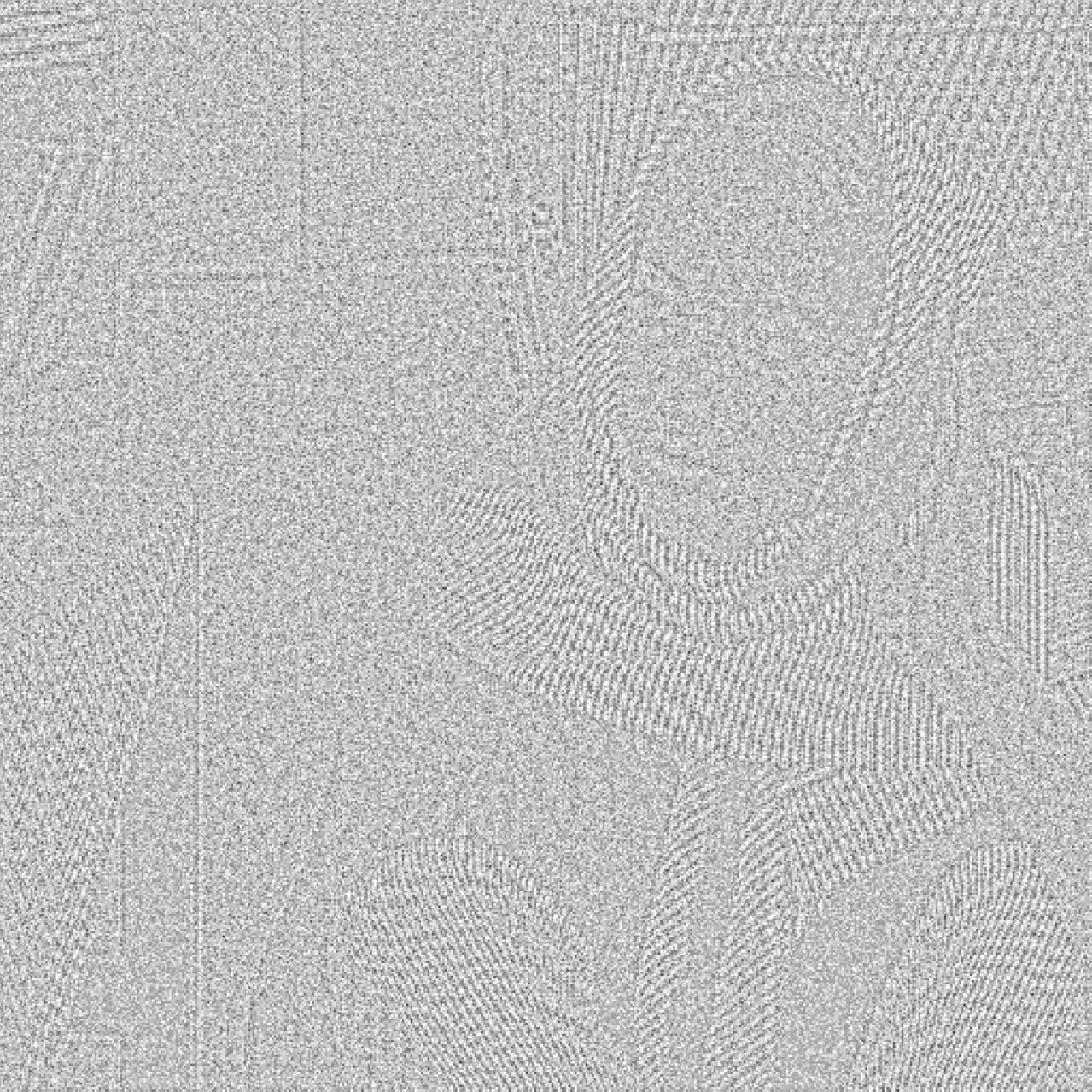}} 
	\subfigure[]{\includegraphics[width=0.244\textwidth]{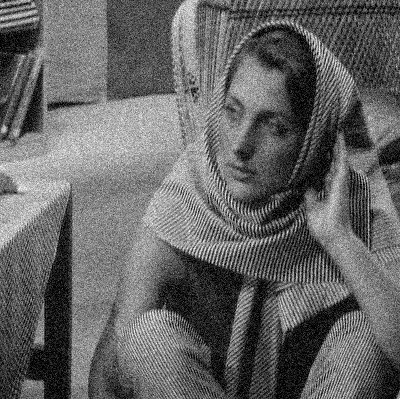}} \\
	\subfigure[DG3PD $\Bu$]{\includegraphics[width=0.244\textwidth]{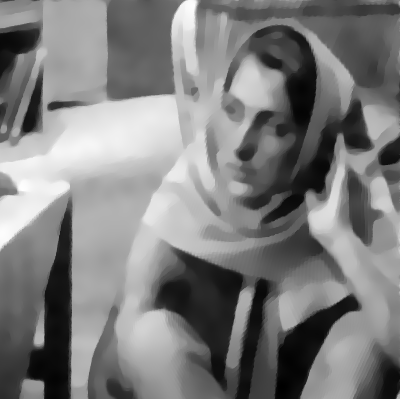}}
	\subfigure[DG3PD $\Bv$]{\includegraphics[width=0.244\textwidth]{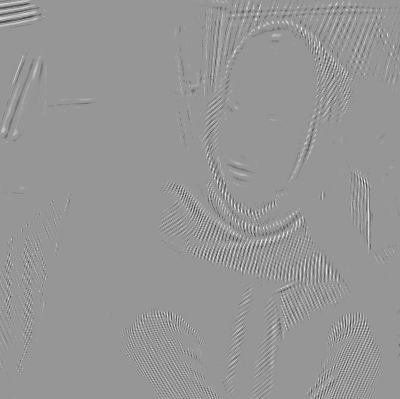}}	
	\subfigure[DG3PD $\Beps$]{\includegraphics[width=0.244\textwidth]{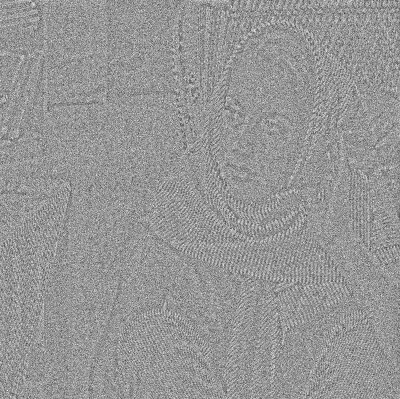}} 
	\subfigure[DG3PD $\Bv_\text{bin}$]{\includegraphics[width=0.244\textwidth]{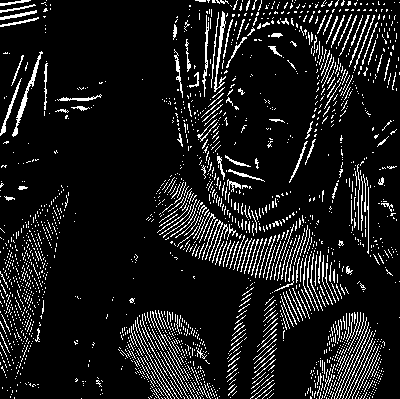}} \\
	\subfigure[QQ plot]{\includegraphics[width=0.33\textwidth]{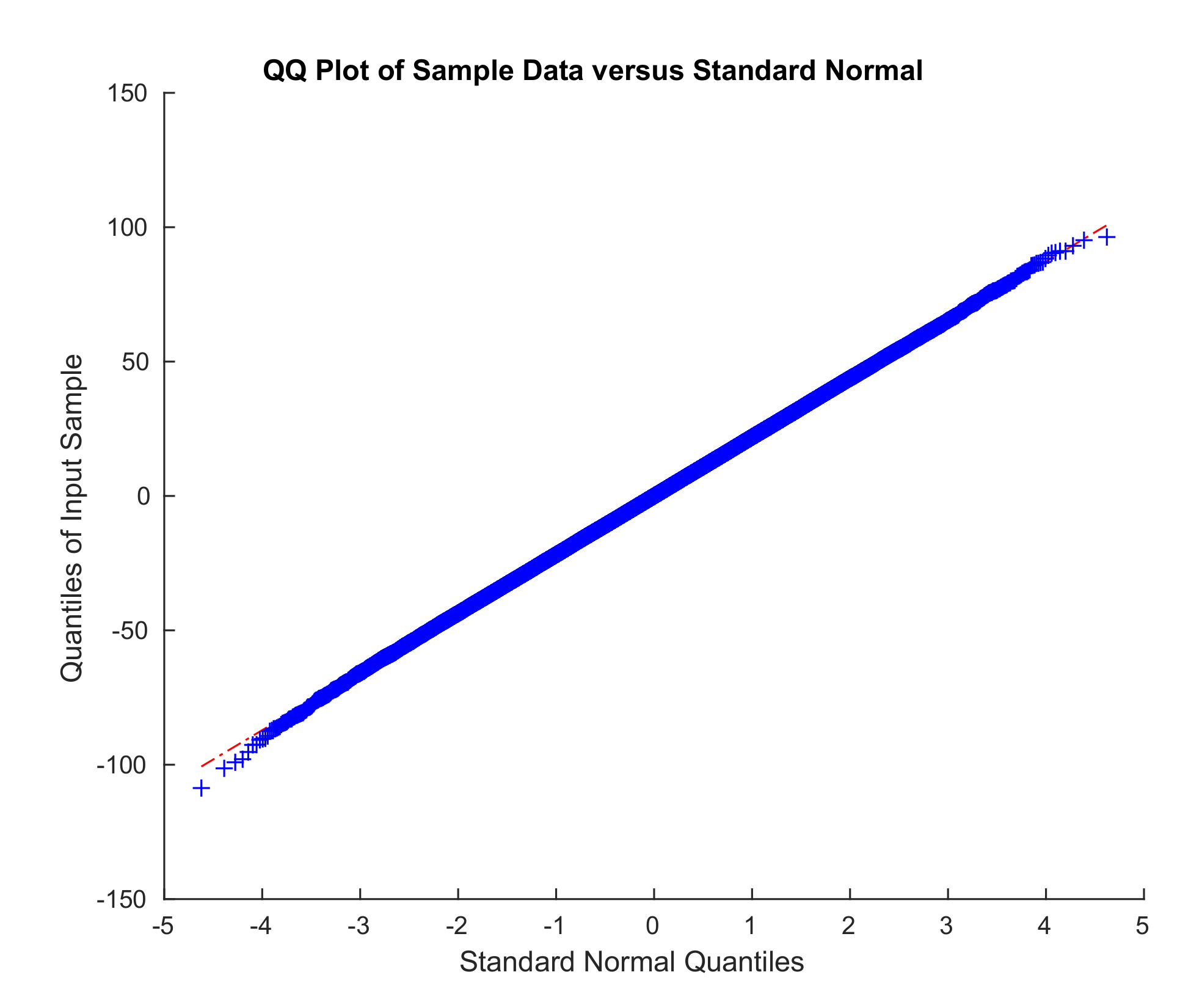}}
\caption{The Barbara image (d) with additive Gaussian white noise ($\sigma = 20$)  
         is decomposed by the Aujol and Chambolle model (a-c)
         and the DG3PD model (e-h) with $\delta = 16$ and the other parameters are the same as in Figure \ref{fig:DG3PD:barbara}. 
	       Comparing (a) and (e) we observe that the cartoon image $\Bu$ obtained by the DG3PD model (e)
	       has a smoother surface and sharper edges than (a). 
	       Comparing the texture images $\Bv$ we note that (f) is smoother and sparser than (b).
	       In order to highlight the sparseness of the DG3PD texture, 
	       all positive coefficients are visualized as white pixels in (h).
				 For visualization, we add 150 to the value of the residual $\boldsymbol \epsilon$ in (g).
         The residual in (g) still contains some texture, but mainly the Gaussian noise which is obviously shown in
         the QQ plot (i). There are some differences at the end of the tail in (i) probably due to the remaining texture and
         the numerical simulation of Gaussian noise.
	}
\label{fig:comparisonAC}
\end{center}
\end{figure}

\textbf{Comparison with Aujol and Chambolle.}
Figure \ref{fig:comparisonAC} illustrates a situation in which the image suffers from i.i.d. Gaussian noise $\mathcal N(0 \,, \sigma)$
with $\sigma = 20$, and compares DG3PD with the AC model \cite{AujolChambolle2005} for three-part decomposition.
It shows that under ``heavy'' noise, our DG3PD model still meets the criteria for cartoon $\Bu$ 
and texture $\Bv$, i.e.
\begin{itemize}
 \item Our cartoon $\Bu$ contains smooth surfaces with sharp edges and no texture, cf. Figure \ref{fig:comparisonAC} (e).
       However, the cartoon $\Bu$ from the AC model is blurry with texture on the scarf, cf. Figure \ref{fig:comparisonAC} (a). 
 \item Our texture $\Bv$ is sparse and smooth, cf. Figure \ref{fig:comparisonAC} (f) and its binarization (h).
       However, the texture from the AC model is not sparse, see Figure \ref{fig:comparisonAC} (b).
\end{itemize}
However,  there is a limitation for both methods: the noise image $\Beps$ contains some pieces of information 
due to the value of $\delta$
which defines the level of the noise. 
Similar to \cite{AujolChambolle2005}, we modify the classical threshold for curvelet coefficients, 
i.e. $\sigma \sqrt{2 \log\abs{\mathcal K}}$, with a weighting parameter $\eta$ as follows
$\delta = \eta \sigma \sqrt{2 \log\abs{\mathcal K}}$ and $\abs{\mathcal K}$ is total number of curvelet coefficients.

\textbf{Summary.}
We observe that the DG3PD method meets all three requirements much more closely 
than the other methods for images without noise, like the original Barbara image.
And in particular, for images with additive noise, 
the DG3PD method still achieves all three goals
as shown in the comparison with Aujol and Chambolle.

%\clearpage

% compare4: penalty-ALM for barbara (crop3)
%\addtolength{\voffset}{-25mm}

%\addtolength{\voffset}{-8mm}

%\addtolength{\voffset}{33mm}

\section{Applications} \label{sec:applications}

Here, we limit ourselves to consider three important applications of the DG3PD method: 
feature extraction, denoising and image compression.

\subsection{Feature Extraction} \label{sec:featextract}

\begin{figure}
\begin{center}
  \subfigure[original with estimated ROI]{\includegraphics[width=0.19\textwidth]{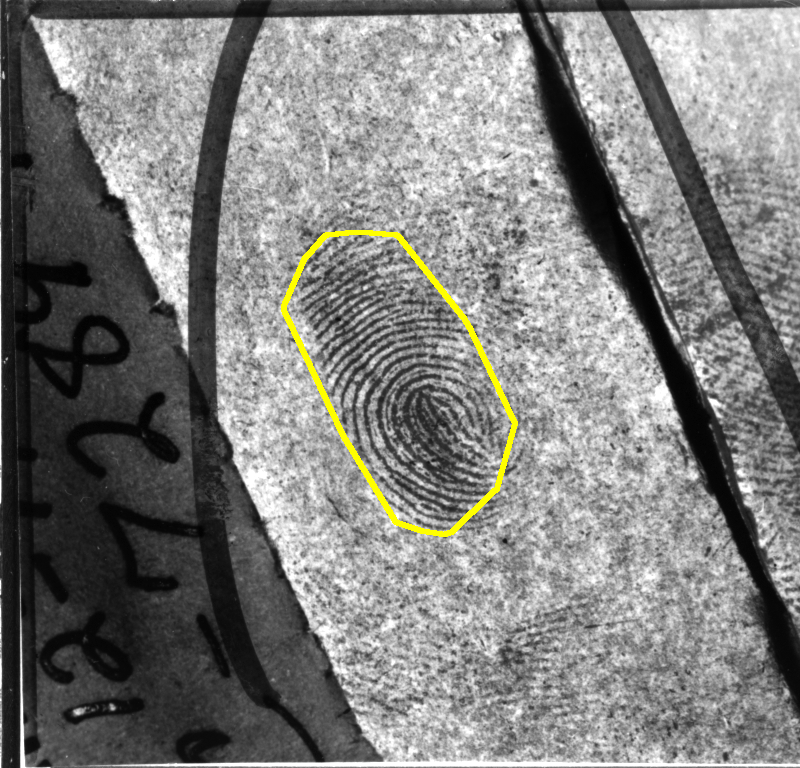}}
	\subfigure[$\Bu$]{\includegraphics[width=0.19\textwidth]{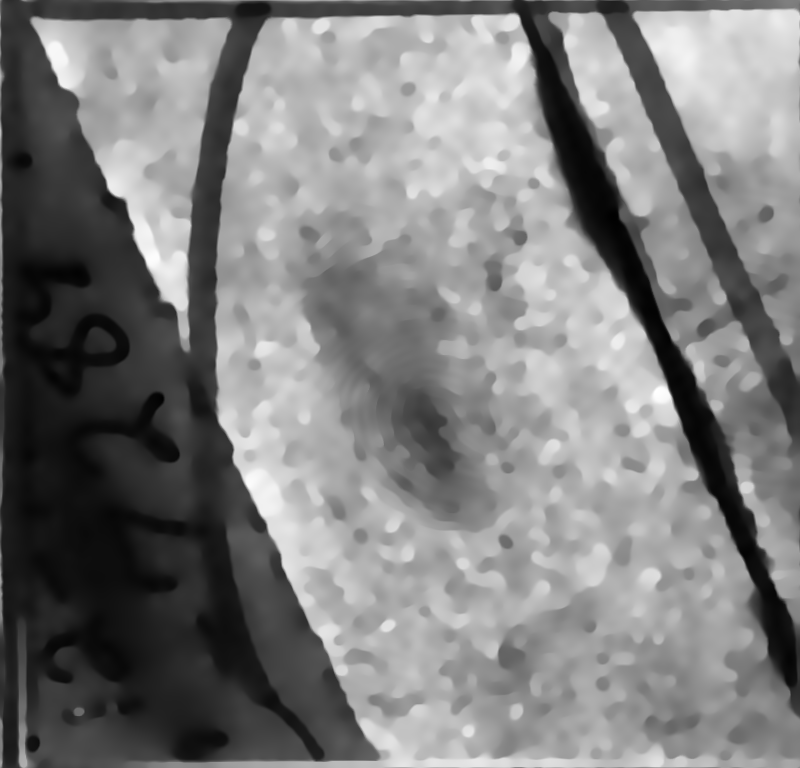}}
	\subfigure[$\Bv$]{\includegraphics[width=0.19\textwidth]{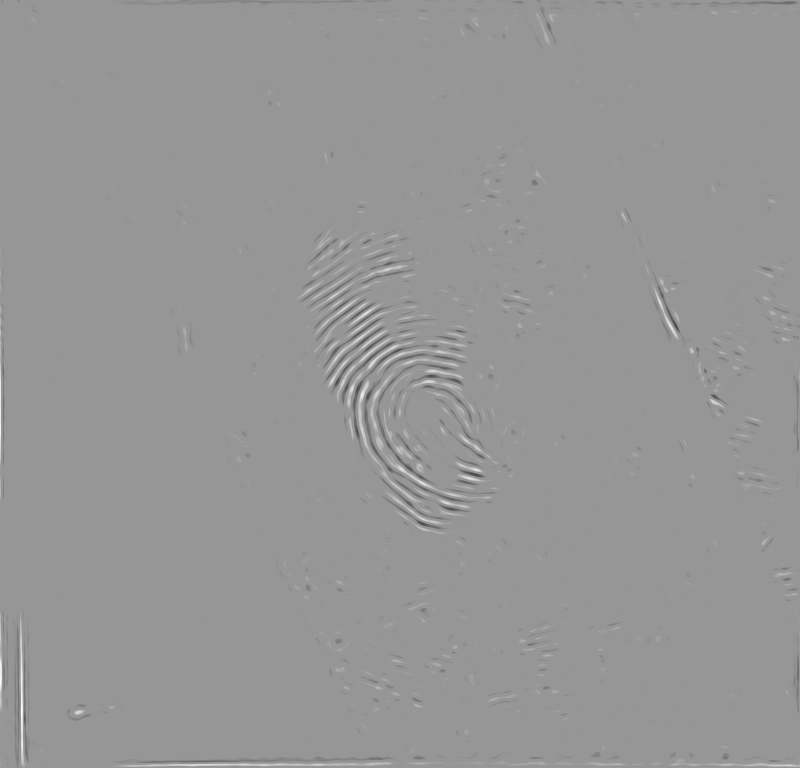}}
	\subfigure[$\Bv_\text{bin}$]{\includegraphics[width=0.19\textwidth]{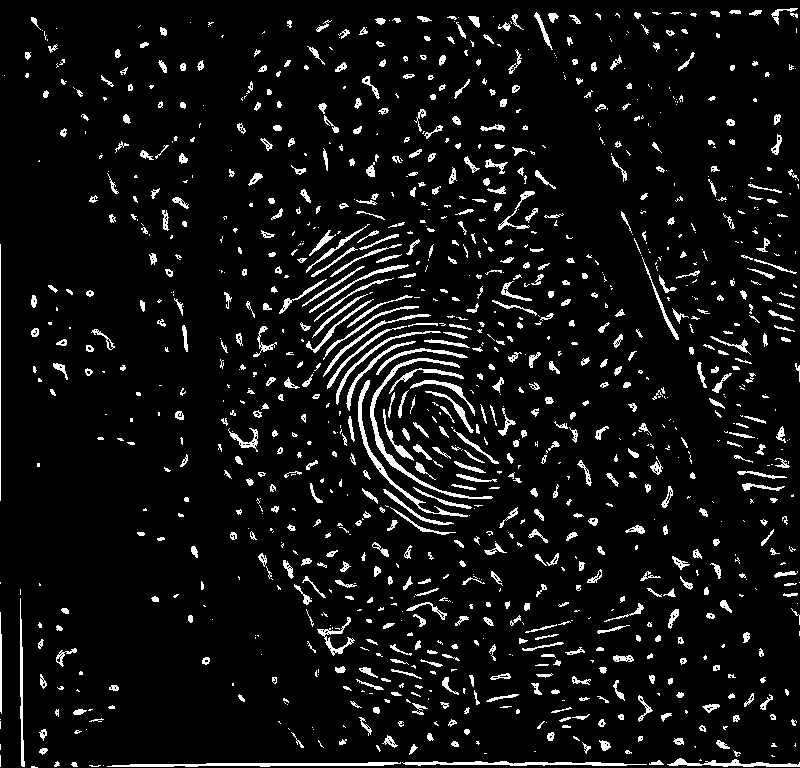}}
	\subfigure[$\Beps$]{\includegraphics[width=0.19\textwidth]{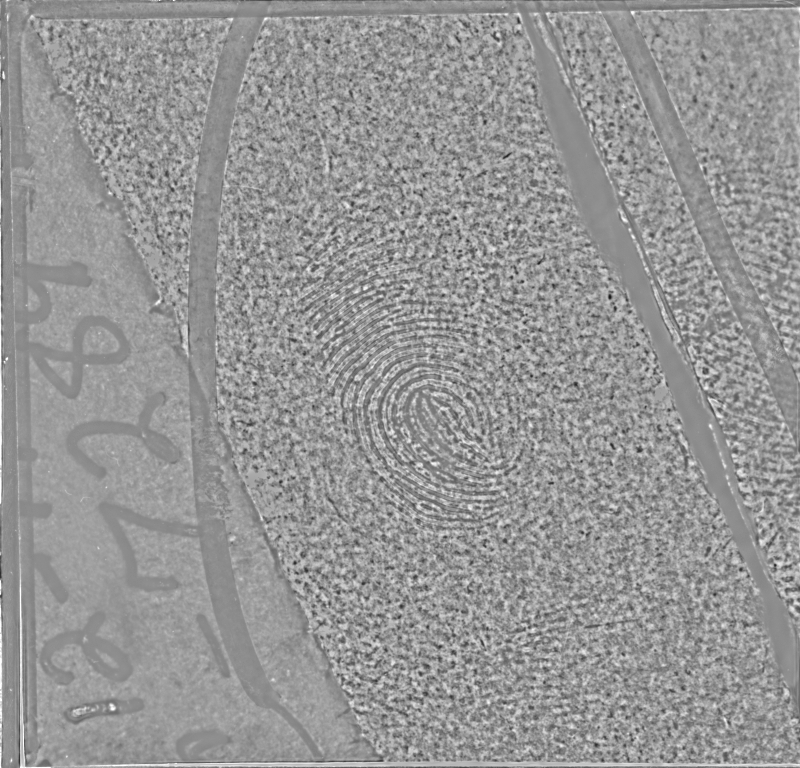}}
\end{center}
\caption{DG3PD decomposition of a latent fingerprint for segmentation (a) with $\delta = 60$.
         The ROI is obtained from the binarized texture $\Bv_\text{bin}$
				 and morphological postprocessing as described in~\cite{ThaiHuckemannGottschlich2015}.}
\label{fig:latentHowTo}
\end{figure}

\begin{figure}
\begin{center}
  \subfigure[]{\includegraphics[width=0.24\textwidth]{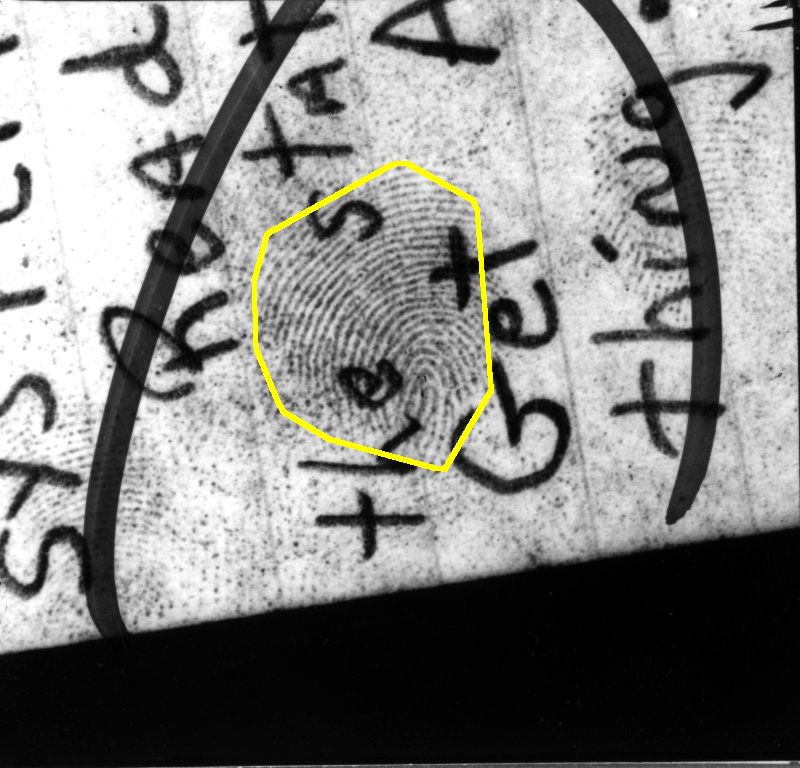}}
	\subfigure[]{\includegraphics[width=0.24\textwidth]{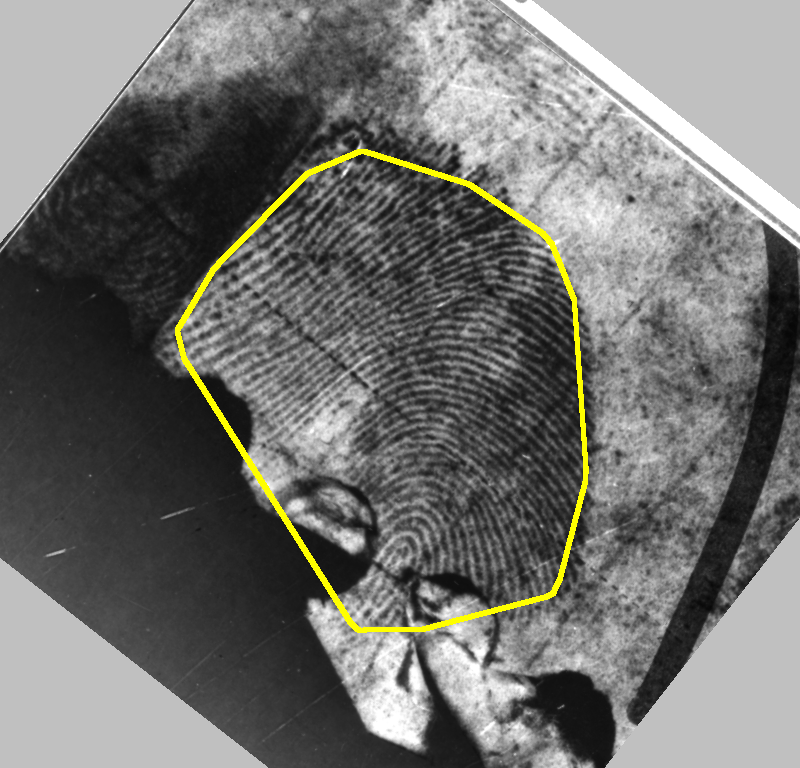}}
	\subfigure[]{\includegraphics[width=0.24\textwidth]{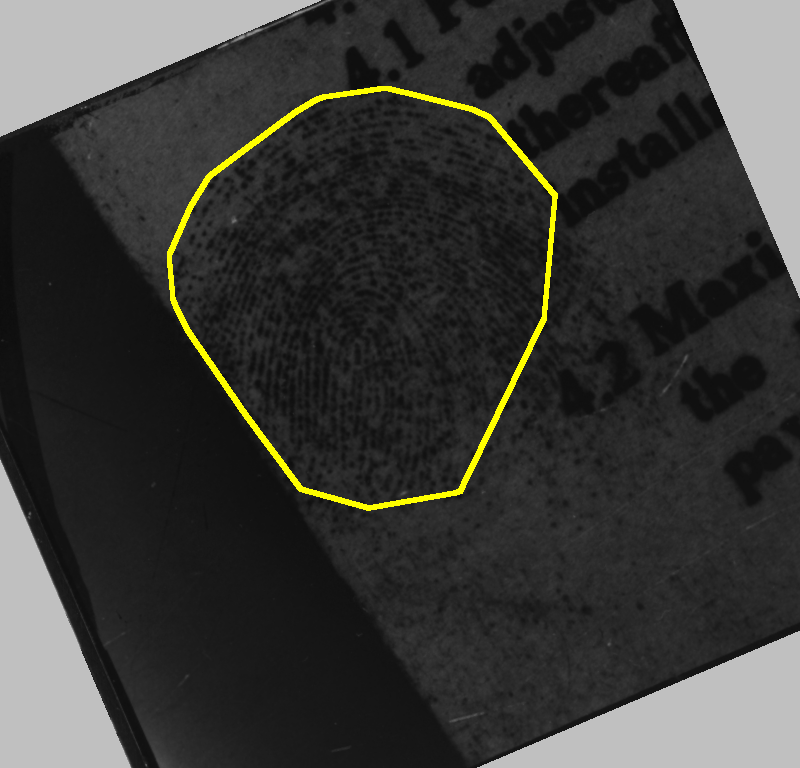}} 
	\subfigure[]{\includegraphics[width=0.24\textwidth]{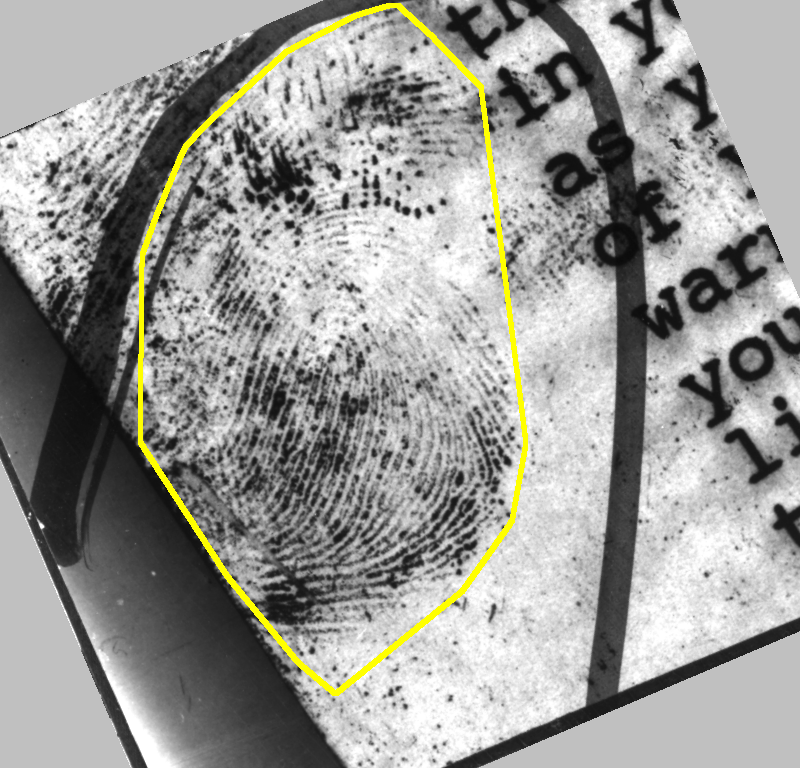}} 
\end{center}
\caption{Latent fingerprint images from NIST SD27. 
         The boundary of the foreground estimated by the DG3PD method is drawn in yellow.
				 }
\label{fig:latentSegmentation}
\end{figure}

Depending on the specific field of application, 
the cartoon or texture, or both can be viewed as feature images.
For the application of DG3PD to fingerprints, we are especially interested in the texture image $\mathbf v$ 
as a feature for subsequent processing steps like segmentation, 
orientation field estimation \cite{GottschlichMihailescuMunk2009}
and ridge frequency estimation \cite{Gottschlich2012} 
and fingerprint image enhancement \cite{Gottschlich2012,GottschlichSchoenlieb2012}.
The first of these processing steps
is to separate the foreground from the background \cite{ThaiHuckemannGottschlich2015,ThaiGottschlich2015G3PD}.
The foreground area (or region-of-interest) contains the relevant information 
for a fingerprint comparison.
Segmentation is still a challenging problem for latent fingerprints \cite{SankaranVatsaSingh2014}
which are very low-quality fingerprints lifted from crime scenes. 
Both the foreground and background area can contain 'noise' on all scales,
from small objects or dirt on the surface, to written or printed characters (on paper)
and large scale objects like an arc drawn by the forensic examiner.
Standard fingerprint segmentation methods cannot cope with this variety of noise,
whereas the texture image by decomposition with the DG3PD method crops out to be an excellent feature for 
estimating the region-of-interest.
Figure \ref{fig:latentHowTo} depicts a detailed example of the latent fingerprint segmentation by the DG3PD decomposition.
In Figure \ref{fig:latentSegmentation}, 
we show further examples of segmentation results 
obtained using the texture image extracted by the DG3PD method 
and morphological postprocessing as described in \cite{ThaiHuckemannGottschlich2015,ThaiGottschlich2015G3PD}.

\subsection{Denoising} 

The DG3PD model can be used for denoising images with texture,
because noise and small scale objects are moved into the residual image $\boldsymbol \epsilon$ 
during the decomposition of $\mathbf f$
due to the supremum norm of the curvelet coefficients of $\boldsymbol \epsilon$ .
Therefore, the image $\mathbf f_\text{denoised} = \mathbf u + \mathbf v$ can be regarded as a denoised version of $\mathbf f$
and the degree of denoising can be steered by the choice of parameters,
especially $\delta$.
For $\delta = 0$ which is equal to two-part decomposition, 
we obtain the original image again. 
As we increase $\delta$, more noise is driven into $\boldsymbol \epsilon$ and thereby removed from $\mathbf f_\text{denoised}$.
Denoising images with texture, in particular with texture parts on different scales, 
is a relevant problem which we plan to address in future works.

\subsection{Compression} \label{sec:compression}

Based on the DG3PD model, we propose a novel approach to image compression.
The core idea is to perform image decomposition by the DG3PD method first, 
and subsequently to compress the three component images by three different algorithms, 
each particularly suited for compressing the specific type of image.
This scheme can be used for lossy as well as lossless compression.

\subsubsection{Cartoon Image Compression} 

As stated in our definition of goals, the cartoon image consists of geometric objects with 
a very smooth or piecewise constant surface and sharp edges. 
This special kind of images is high compressible and a very effective approach is based on diffusion.
Anisotropic diffusion \cite{Weickert1998} is useful for many purposes in image processing
e.g. fingerprint image enhancement by oriented diffusion filtering \cite{GottschlichSchoenlieb2012}.

The basic idea of diffusion based compression is store information for only a few sparse locations 
which encode the edges of the cartoon image. 
The surface areas are inpainted using a linear or nonlinear diffusion process.
Please note that the cartoon image obtained by the DG3PD method is much better suited 
for this type of compression due to the property of sharper edges between geometric objects 
in comparison to the cartoon images of the other decomposition approaches.
Moreover, some difficulties and drawbacks of diffusion based compression for arbitrary images 
do not apply to this special case. 
In general, it is a challenging question 
how and where to select locations for diffusion seed points.
In our case, this task is easily solvable,
because of the sharp edges between in homogeneous regions in the DG3PD cartoon image.
This allows for an extremely sparse selection of locations on corners and edges.

Image compression with edge-enhancing anisotropic diffusion (EED) 
has been studied by Galic \textit{et al.} \cite{GalicWeickertWelkBruhnBelyaevSeidel2008}
and has been improved by Schmaltz \textit{et al.} \cite{SchmaltzPeterMainbergerEbelWeickertBruhn2014}.
Compression of cartoon-like images with homogeneous diffusion has been analyzed 
by Mainberger \textit{et al.} \cite{MainbergerBruhnWeickertForchhammer2011}.
%Unfortunately, for none of these methods exists a publicly available implementation.
%We plan to address this deficiency in our future work.
A viable alternative to diffusion based compression of cartoon images
is a dictionary based approach \cite{EladAharon2006} in which the dictionary is optimized for cartoon images.
Another very promising possibility to compress the DG3PD cartoon component is the usage of linear splines
over adaptive triangulations which has been proposed in the work of Demaret \textit{et al.} \cite{DemaretDynIske2006}.

\subsubsection{Texture Image Compression} 

Tailor-made solutions are available for texture image compression,
and especially for compressing oscillating patterns like e.g. fingerprints.

Larkin and Fletcher \cite{LarkinFletcher2007}
achieved a compression rate of 1:239 for a fingerprint image
using amplitude and frequency modulated (AM-FM) functions.
They decompose a fingerprint image into four parts
and this idea can be applied to the texture image $\Bv$ 
obtained by the DG3PD method:

$\Bv \approx \boldsymbol a + [ \boldsymbol b \cdot cos( \boldsymbol \Psi_C + \boldsymbol \Psi_S ) ] $ 

\begin{figure}
\begin{center} 
  \subfigure[Barbara $\Bu$]{\includegraphics[width=0.21\textwidth]{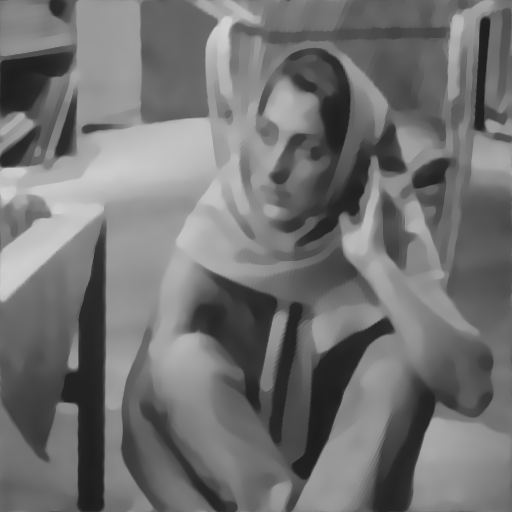}}
	\subfigure[2000]{\includegraphics[width=0.21\textwidth]{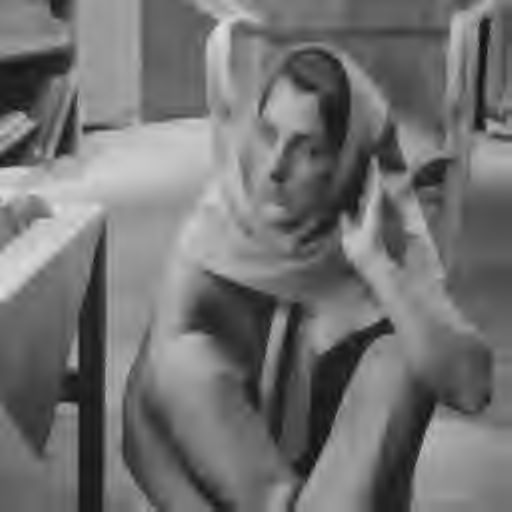}}
	\subfigure[5000]{\includegraphics[width=0.21\textwidth]{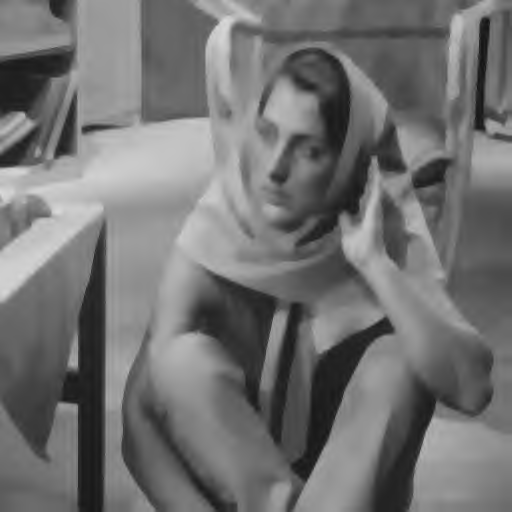}} 
	\subfigure[8000]{\includegraphics[width=0.21\textwidth]{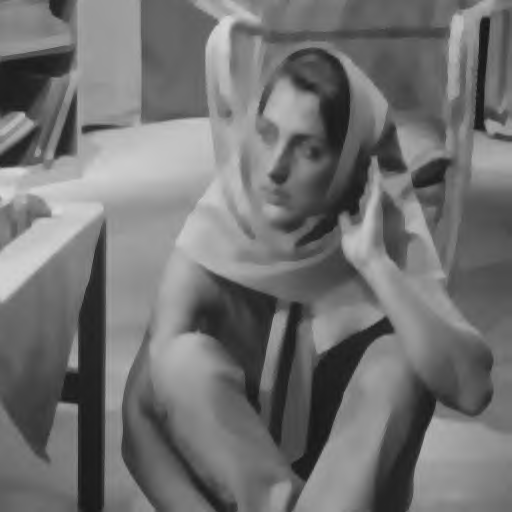}}
	\subfigure[Barbara $\Bv$]{\includegraphics[width=0.21\textwidth]{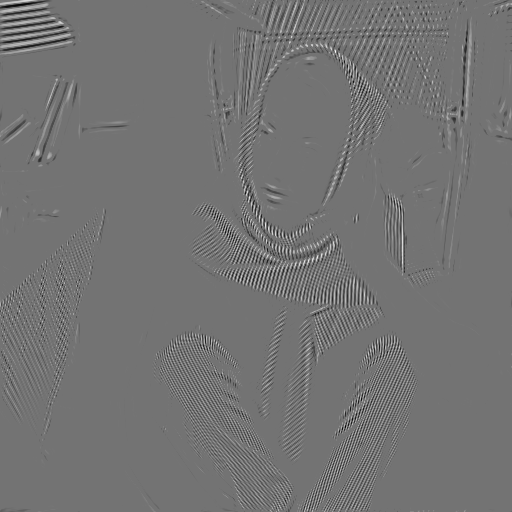}}
	\subfigure[CR 1:15]{\includegraphics[width=0.21\textwidth]{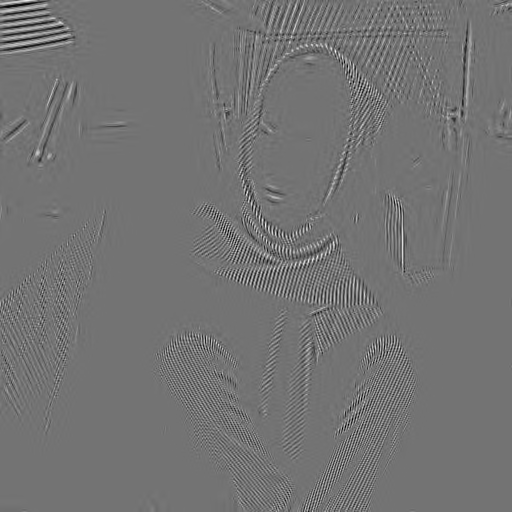}}
	\subfigure[CR 1:55]{\includegraphics[width=0.21\textwidth]{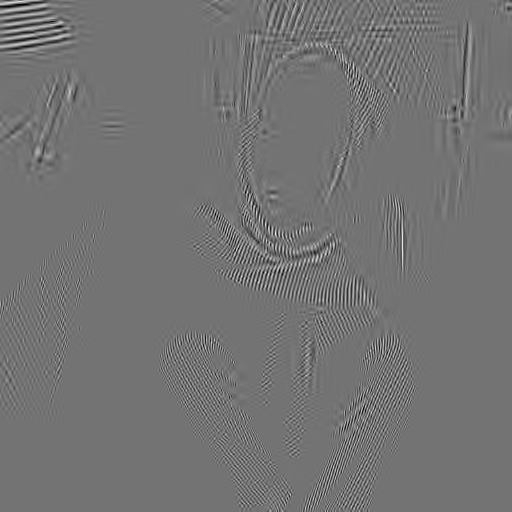}} 
	\subfigure[CR 1:85]{\includegraphics[width=0.21\textwidth]{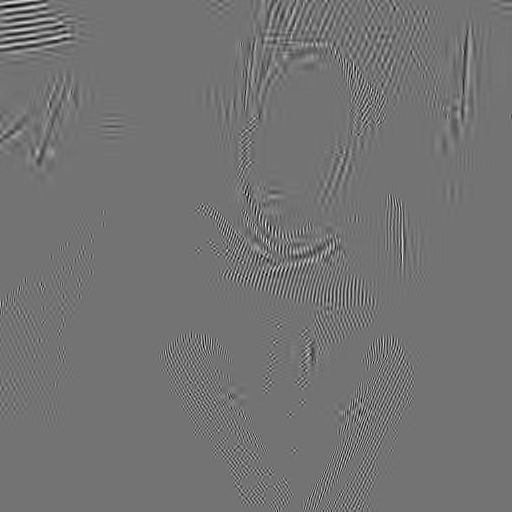}} 
\caption{The cartoon component $\Bu$ (a) and texture component $\Bv$ (e) of the Barbara image obtained by DG3PD.
	       Images (b-d) display the reconstruction of $\Bu$ from the 2000, 5000, and 8000 largest 
				 Wavelet coefficients (from a total of 262,144 coefficients).
	       (f-h) show the decompressed images after compression of $\Bv$ 
				 by WSQ at compression rates between 1:15 and 1:85.
			   }
\label{fig:componentCompression}
\end{center}
\end{figure}

Each of the four components is again highly compressible and can be stored with only a few bytes (see Figure 5 in  \cite{LarkinFletcher2007}).
This is remarkable and we would like offer another perspective on the AM-FM model.
Storing a minutiae template can be viewed as a lossy form of fingerprint compression.
The minutiae of a fingerprint are locations where ridges (dark lines) end or bifurcate.
and a template stores the locations and directions of these minutiae.
Several algorithms have been proposed for reconstructing 
the orientation field (OF) from a minutia template \cite{OehlmannHuckemannGottschlich2015}.
The continuous phase $\boldsymbol \Psi_C$ can be derived from the unwrapped reconstructed OF 
and the spiral phase $\boldsymbol \Psi_S$ directly constructed from the minutiae template.
Choosing appropriate values for  $\boldsymbol a$ and $\boldsymbol b$ leads to a fingerprint image.
A survey of further methods for reconstructing fingerprints from their minutiae template is given in \cite{GottschlichHuckemann2014}.
An alternative way of lossy fingerprint image compression is 
wavelet scalar quantization (WSQ) \cite{WSQ}
which has been a compression standard for fingerprints used by the Federal Bureau of Investigation in the United States.
See Figure \ref{fig:componentCompression} (f-h) for application example of WSQ to the texture of the Barbara image.

A third, very good compression possibility is dictionary learning \cite{EladAharon2006}
with optimization of the dictionary for the texture component $\Bv$.
For fingerprint images, this problem has recently been studied by Shao \textit{et al.} \cite{ShaoWuYongGuoLiu2014}.

\subsubsection{Residual Image Compression} 

For image compression using DG3PD, we propose the following steps in this order:
First, image decomposition $\mathbf f = \mathbf u + \mathbf v + \boldsymbol \epsilon$.
Second, a tailor-made, lossy, high compression of the cartoon component $\mathbf u$ 
and the texture component $\mathbf v$.
Third, decompressing $\mathbf u$ and $\mathbf v$
in order to compute the compression residual image $\mathbf s = \mathbf f - \mathbf u_d - \mathbf v_d$,
where $\mathbf u_d$ is the cartoon image and $\mathbf v_d$ the texture image after decompression.
Fourth, compression of $\mathbf s$.

In step two and four, the term ``compression'' denotes the whole process including coefficient quantization
and symbol encoding (see \cite{Salomon2007Compression} for scalar quantization, Huffman coding, LZ77, LZW 
and many other standard techniques).

Let be $\mathbf e_u = \mathbf u - \mathbf u_d$ 
the difference between the cartoon component before and after compression, i.e. the compression error,
and $\mathbf e_v = \mathbf v - \mathbf v_d$,
then we can rewrite $\mathbf r = \mathbf f - \mathbf u_d - \mathbf v_d = \mathbf u + \mathbf v + \boldsymbol \epsilon - \mathbf u_d - \mathbf v_d = \boldsymbol \epsilon + \mathbf e_u + \mathbf e_v$. 
Hence, the residual image $\mathbf s$ computed in step four contains the residual component $\boldsymbol \epsilon$ 
plus the compression errors of the other two components. 
Now, lossless compression can be achieved by lossless compression of $\mathbf s$.
If the goal is lossy compression with a certain target quality or target compression rate,
this can be achieved by adapting the lossy compression of $\mathbf s$ accordingly.
See Figure \ref{fig:componentCompression} for the effects of different compression rates
on the decompressed cartoon $\mathbf u_d$ and the decompressed texture $\mathbf v_d$.

An additional advantage of decompression beginning with $\mathbf u_d$, followed by $\mathbf v_d$
and finally $\mathbf s_d$ is the fast generation of a preview image
which mimics the effects of interlacing.
In a scenario with limited bandwidth for data transmission, 
e.g. sending an image to a mobile phone, the user can be shown a preview 
based on the compressed, transmitted and decompressed $\mathbf u$ image.
During the transmission of the compressed  $\mathbf v$ 
and  $\mathbf s$, the user can decide whether to continue or abort the transmission.

\section{Conclusion} \label{sec:conclusion}

The DG3PD model is a novel method for three-part image decomposition. 
We have shown that the DG3PD method achieves the goals defined in the introduction
much better than other relevant image decomposition approaches. 
The DG3PD model lays the groundwork for applications such as image compression, denoising
and feature extraction for challenging tasks such as latent fingerprint processing.
We follow in the footsteps of Aujol and Chambolle \cite{AujolChambolle2005} 
who pioneered three-part decomposition and DG3PD generalizes their approach.
We believe that three-part decomposition is the 
way forward to address many important problems in image processing and computer vision.
Buades \textit{et al.} \cite{BuadesLeMorelVese2010} asked in 2010: 
``Can images be decomposed into the sum of a geometric part and a textural part?''
Our answer to that question is: no if an image contains other parts than cartoon and texture, i.e. noise or small scale objects.
Consider e.g. the noisy Barbara image in Figure~\ref{fig:comparisonAC}~(d).
If the sum of the cartoon and texture images shall reconstruct the input image $\mathbf f$,
a two-part decomposition has to assign the noise parts either to the cartoon or to the texture component.
In principle, not even the best two-part decomposition model can fully achieve both goals 
regarding the desired properties of the cartoon and texture component simultaneously. 
The solution is that noise and small scale objects 
which do not belong to the cartoon or texture have to be allotted to a third component.

In our future work we intend to optimize the DG3PD method for specific applications,
especially image compression and latent fingerprint processing. 
Issues for improvement include the data-driven, automatic parameter selection 
and the convergence rate (can the same decomposition be achieved in fewer iterations?)
Furthermore, we plan to explore and evaluate specialized compression approaches for cartoon, texture and residual images.

\section*{Acknowledgements}

D.H Thai is supported by the National Science Foundation 
under Grant DMS-1127914 to the Statistical and Applied Mathematical Sciences Institute.
C. Gottschlich gratefully acknowledges the support of the 
Felix-Bernstein-Institute for Mathematical Statistics in the Biosciences 
and the Niedersachsen Vorab of the Volkswagen Foundation.

\section*{Appendix} \label{sec:appendix}

%% --------------------------------------
%% --------------------------------------------------
 \begin{figure}[b]
 \begin{center}     
   % ROF, VeseOsher, StarckEladDonoho, A2BC:
   % u:
   \subfigure[ROF: $\Bu$]{\includegraphics[width=0.21\textwidth,height=0.21\textwidth]{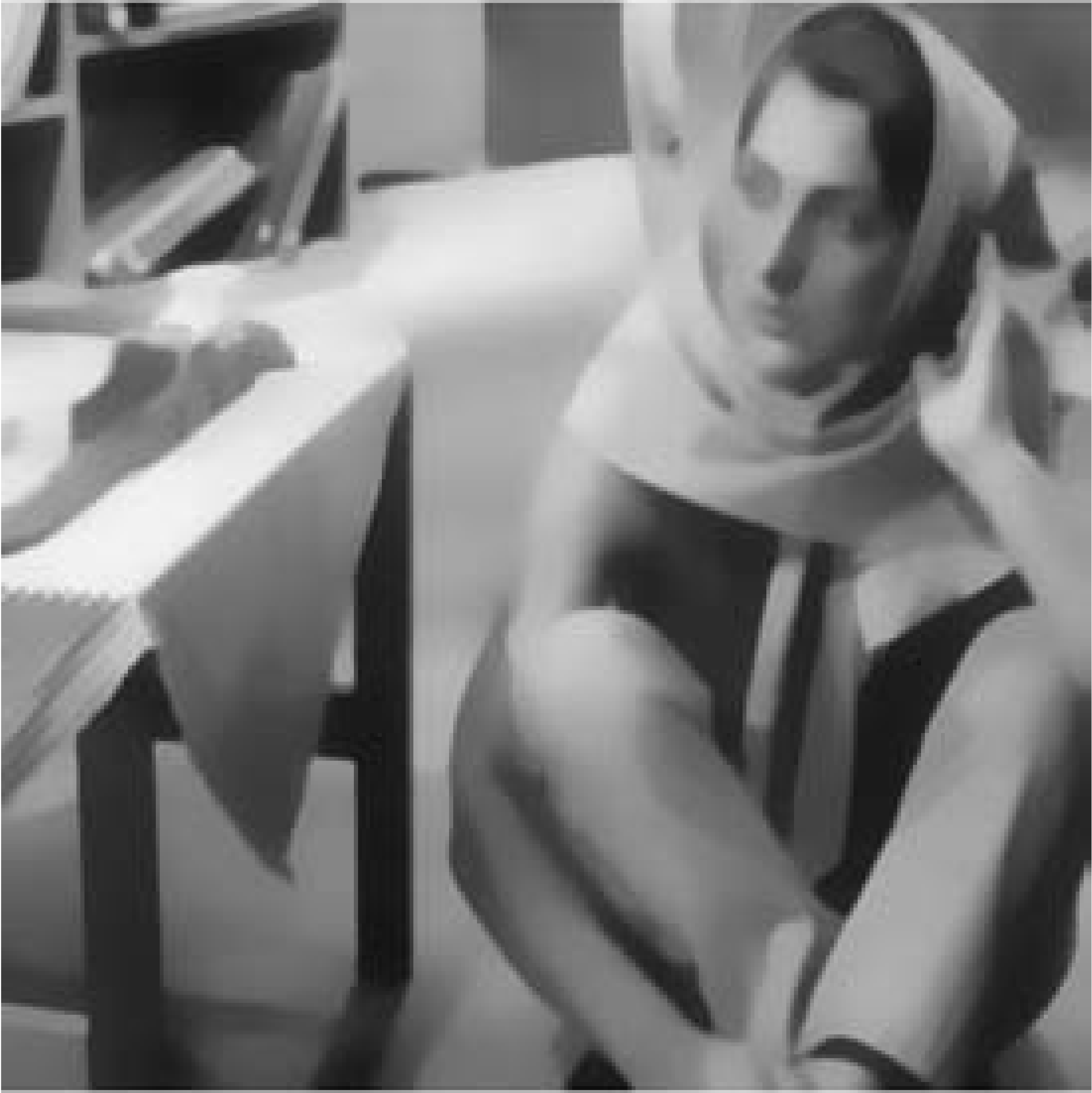}}
   \subfigure[VO: $\Bu$]{\includegraphics[width=0.21\textwidth,height=0.21\textwidth]{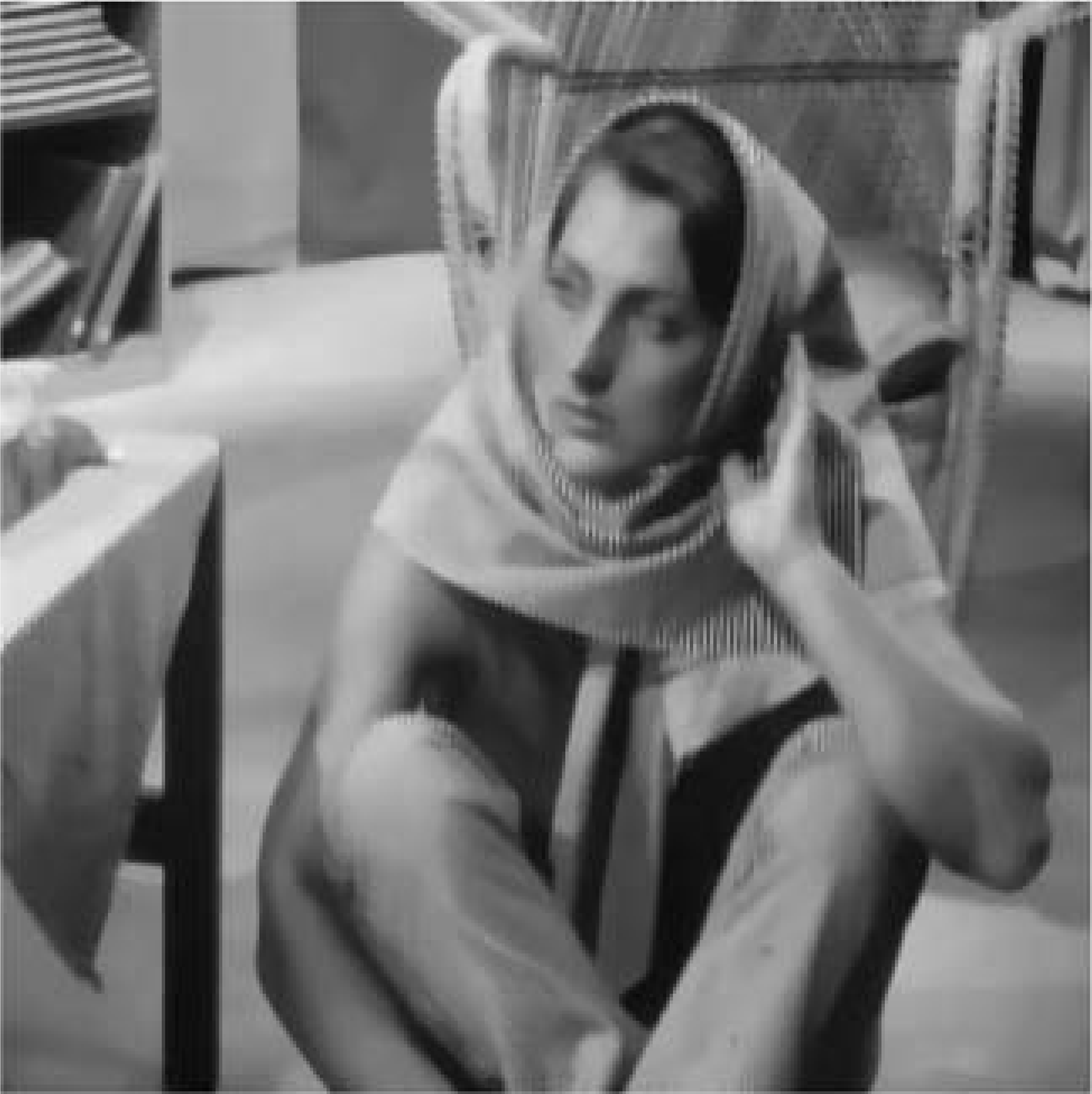}}
   \subfigure[SED: $\Bu$]{\includegraphics[width=0.21\textwidth,height=0.21\textwidth]{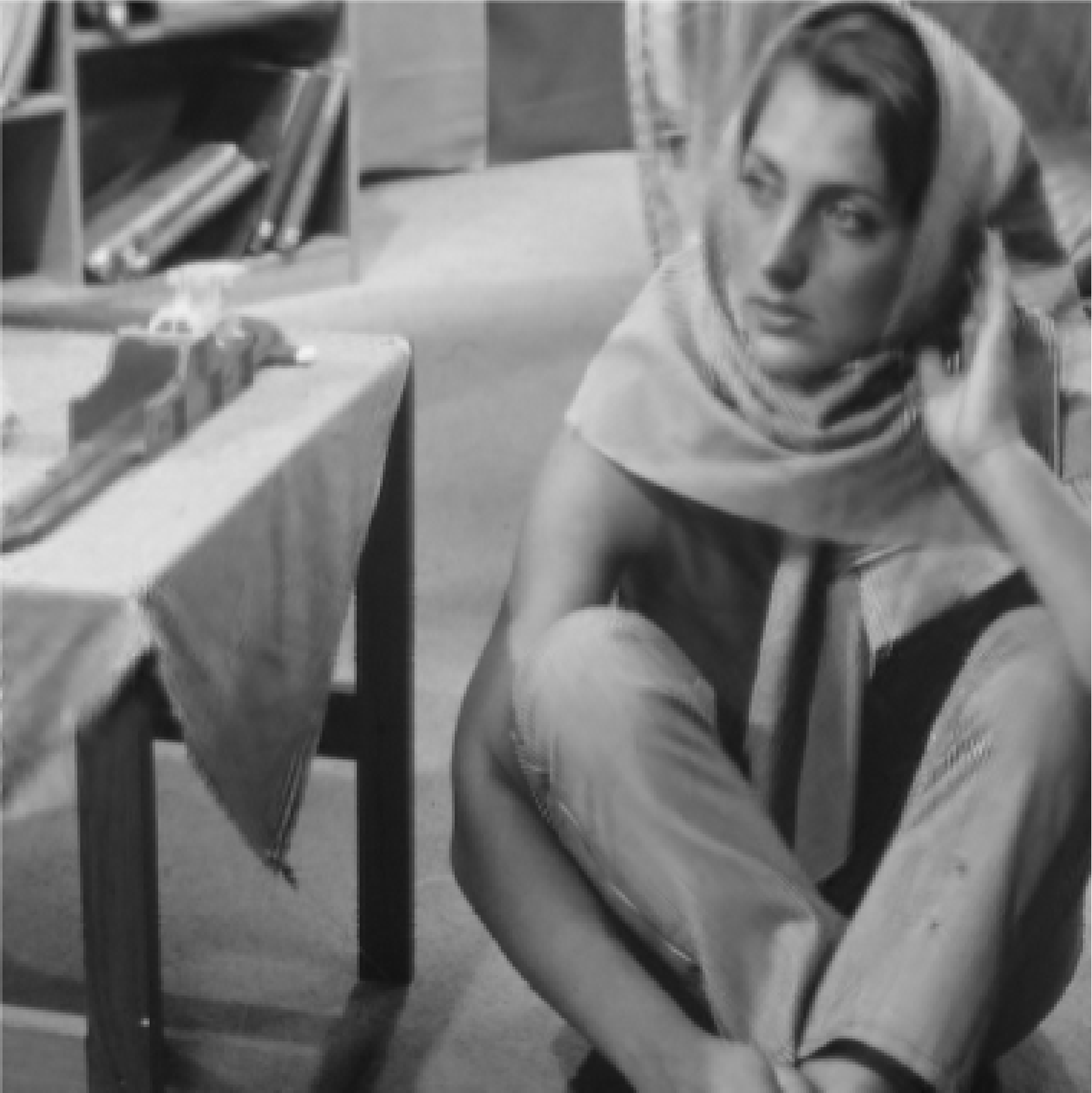}}
   \subfigure[TVG: $\Bu$]{\includegraphics[width=0.21\textwidth,height=0.21\textwidth]{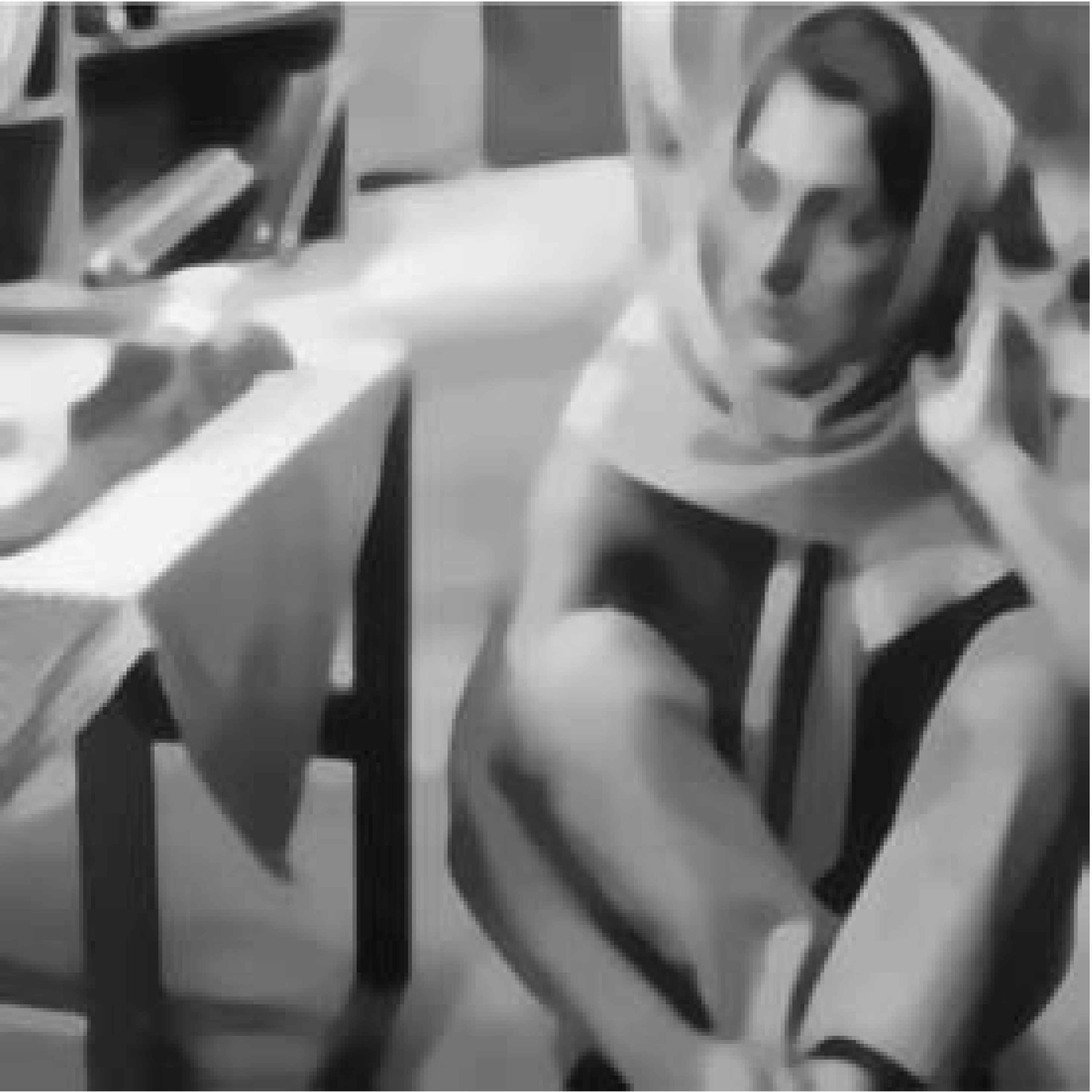}}   
   % v:
   \subfigure[ROF: $\Bv$]{\includegraphics[width=0.21\textwidth,height=0.21\textwidth]{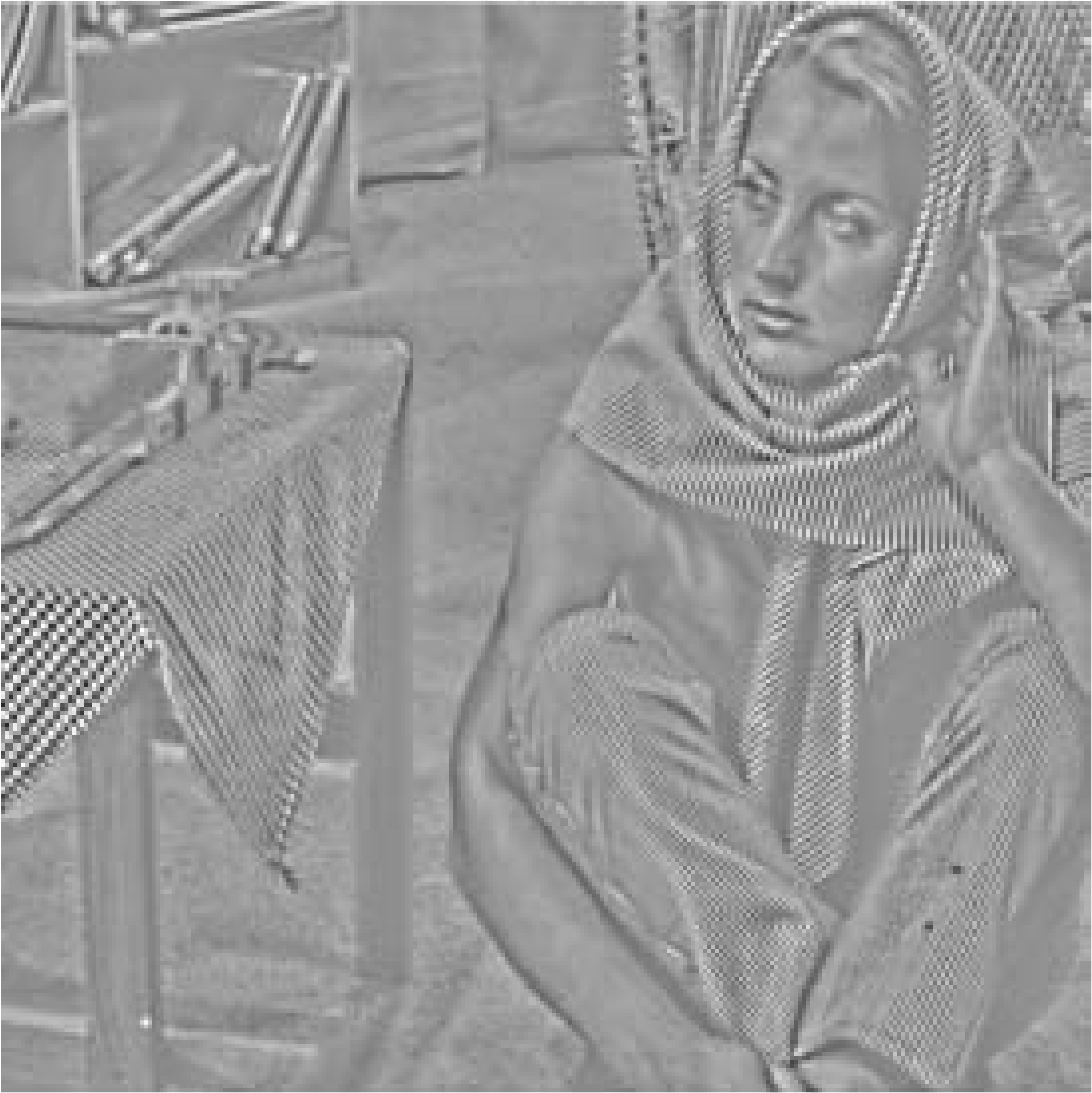}}  
   \subfigure[VO: $\Bv$]{\includegraphics[width=0.21\textwidth,height=0.21\textwidth]{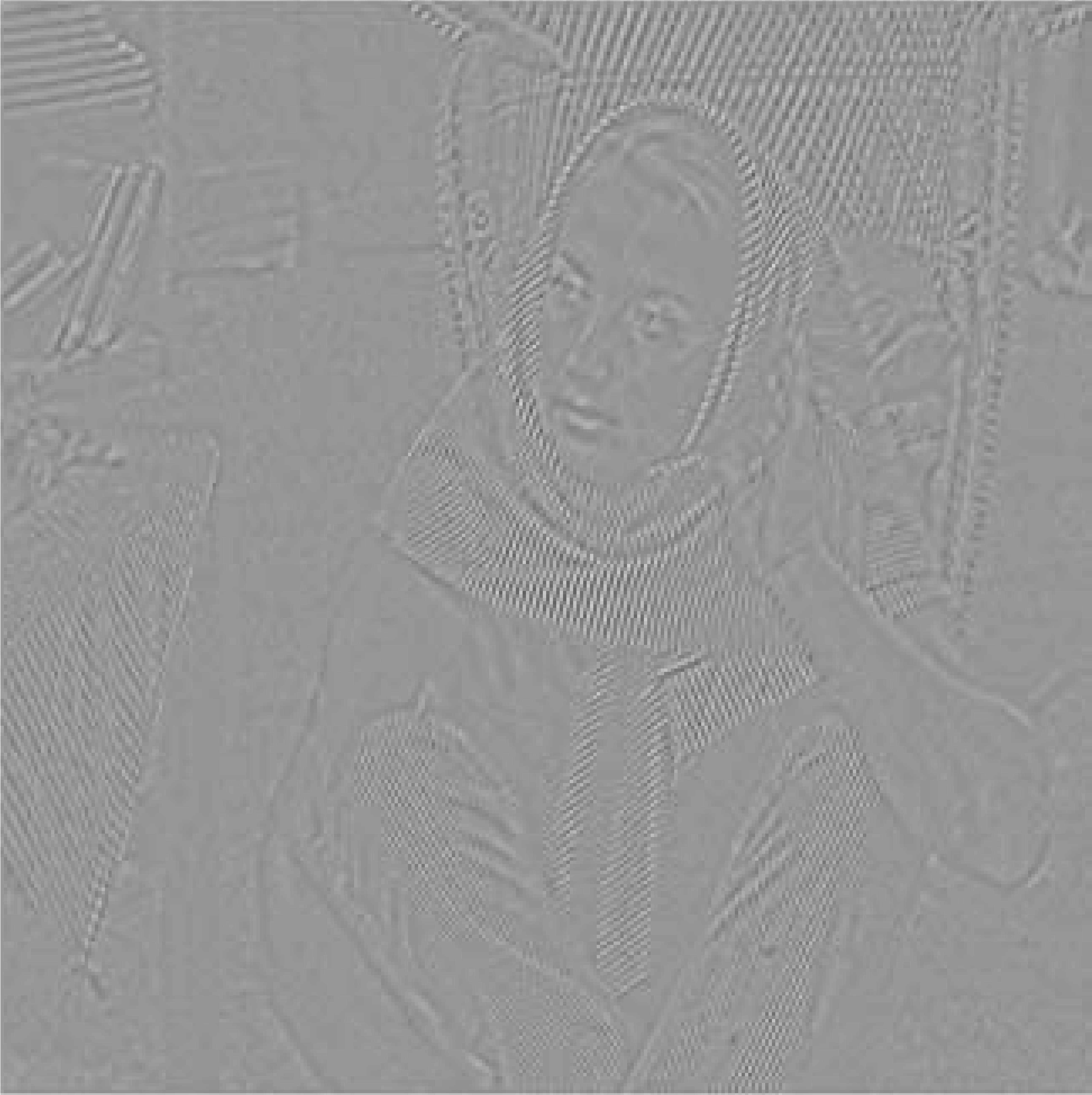}}
   \subfigure[SED: $\Bv$]{\includegraphics[width=0.21\textwidth,height=0.21\textwidth]{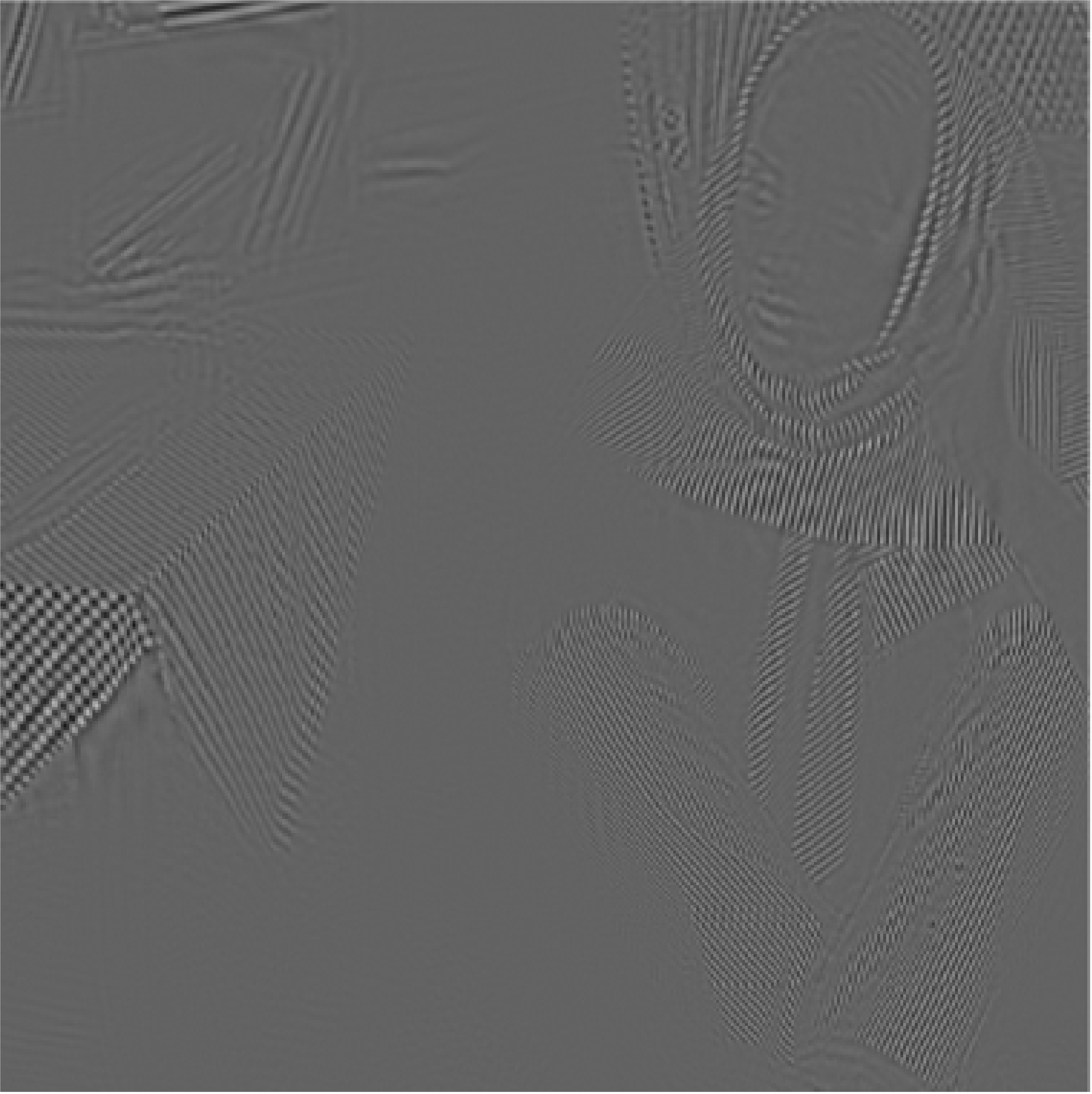}}
   \subfigure[TVG: $\Bv$]{\includegraphics[width=0.21\textwidth,height=0.21\textwidth]{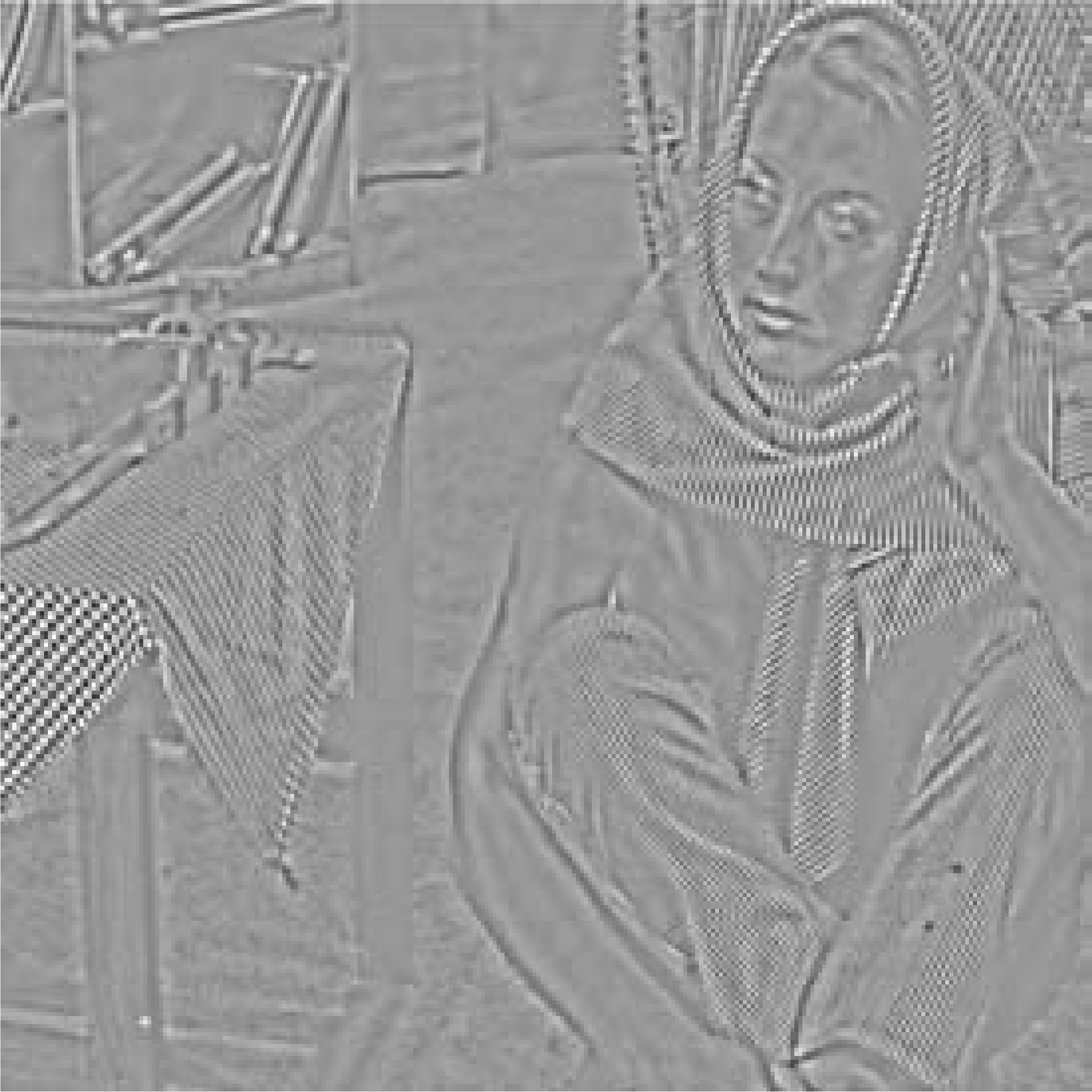}}     
   % G3PD:
   % delta = 0:
   \subfigure[DG3PD: $\Bu$, $\delta = 0$]{\includegraphics[width=0.21\textwidth,height=0.21\textwidth]{barbarabeta4_0-04_c1_1_S_9_L_9_Z_1_I_1_M_1_N_1_xi_0_alpha_0_T_2_delta_0nonoise_u.png}}  
   \subfigure[$\Bv$]{\includegraphics[width=0.21\textwidth,height=0.21\textwidth]{barbarabeta4_0-04_c1_1_S_9_L_9_Z_1_I_1_M_1_N_1_xi_0_alpha_0_T_2_delta_0nonoise_v.png}}
   \subfigure[$\Bv_\text{bin}$]{\includegraphics[width=0.21\textwidth,height=0.21\textwidth]{barbarabeta4_0-04_c1_1_S_9_L_9_Z_1_I_1_M_1_N_1_xi_0_alpha_0_T_2_delta_0nonoise_vbin.png}}
   \subfigure[$\Beps$]{\includegraphics[width=0.21\textwidth,height=0.21\textwidth]{barbarabeta4_0-04_c1_1_S_9_L_9_Z_1_I_1_M_1_N_1_xi_0_alpha_0_T_2_delta_0nonoise_epsilon.png}}       
   % delta = 10:
   \subfigure[DG3PD: $\Bu$, $\delta = 10$]{\includegraphics[width=0.21\textwidth,height=0.21\textwidth]{barbarabeta4_0-04_c1_1_S_9_L_9_Z_1_I_1_M_1_N_1_xi_0_alpha_0_T_2_delta_10nonoise_u.png}}  
   \subfigure[$\Bv$]{\includegraphics[width=0.21\textwidth,height=0.21\textwidth]{barbarabeta4_0-04_c1_1_S_9_L_9_Z_1_I_1_M_1_N_1_xi_0_alpha_0_T_2_delta_10nonoise_v.png}}
   \subfigure[$\Bv_\text{bin}$]{\includegraphics[width=0.21\textwidth,height=0.21\textwidth]{barbarabeta4_0-04_c1_1_S_9_L_9_Z_1_I_1_M_1_N_1_xi_0_alpha_0_T_2_delta_10nonoise_vbin.png}}
   \subfigure[$\Beps$]{\includegraphics[width=0.21\textwidth,height=0.21\textwidth]{barbarabeta4_0-04_c1_1_S_9_L_9_Z_1_I_1_M_1_N_1_xi_0_alpha_0_T_2_delta_10nonoise_epsilon.png}}         
 \caption{Comparison of decomposition results by the ROF \cite{RudinOsherFatemi1992}, VO \cite{VeseOsher2003}, 
          SED \cite{StarckEladDonoho2005}, TVG \cite{AujolGilboaChanOsher2006} and DG3PD models.
          Parameters for DG3PD are $\beta_4 = 0.04 \,, \theta = 0.9 \,, c_1 = 1 \,, c_2 = 1.3 \,,
          c_{\mu_1} = c_{\mu_2} = 0.03 \,, \gamma = 1 \,, S = L = 9$.
         }
 \label{fig:comparisonWholeBarbara}
 \end{center}
 \end{figure}
 % -------------------------------------------------

\begin{algorithm}
\label{alg:DG3PD}
\caption{The Discrete DG3PD Model}
\begin{algorithmic}
\small
% Initialization: $\Bu^{(0)} = \big[\mathbf{r}_{\mathbf l}^{(0)}\big]_{l=0}^{L-1} = \big[\mathbf{g}_{\mathbf s}^{(0)}\big]_{s=0}^{S-1} =
%                  \big[\mathbf{w}_{\mathbf s}^{(0)}\big]_{s=0}^{S-1} = \mathbf 0$
 \STATE{
  {\bfseries Initialization:}
  $\Bu^{(0)} = \Bf \,, \Bv^{(0)} = \Beps^{(0)} = \big[ \Br_l^{(0)} \big]_{l=0}^{L-1} = \big[ \Bw_s^{(0)} \big]_{s=0}^{S-1} = \big[ \Bg_s^{(0)} \big]_{s=0}^{S-1}
   = \big[\boldsymbol{\lambda}_{\mathbf{1}l}^{(0)}\big]_{l=0}^{L-1} = \big[\boldsymbol{\lambda}_{\mathbf{2}a}^{(0)}\big]_{a=0}^{S-1}
   = \boldsymbol{\lambda}_{\mathbf 3}^{(0)} = \boldsymbol{\lambda}_{\mathbf 4}^{(0)} =\boldsymbol 0$.
 } 
 \STATE{ } 
 \FOR{$t = 1 \,, \ldots \,, T $}
 \STATE
{
  {\bfseries 1. Compute}
%   $(\mathbf{u}^{(t)} \,, \mathbf{r}_{\mathbf 0}^{(t)} \,, \ldots \,, \mathbf{r}_{\mathbf{L-1}}^{(t)} ) \in X^{L+1}$:
   $\Big( \big[ \Br_b^{(t)} \big]_{b=0}^{L-1} \,, \big[ \mathbf{w}_a^{(t)} \big]_{a=0}^{S-1} \,,
    \big[ \mathbf{g}_a^{(t)} \big]_{a=0}^{S-1} \,, \Bv^{(t)} \,, \Bu^{(t)} \,, \Beps^{(t)} \Big) \in X^{L+2S+3}
   $:
   \begin{align*}
    \Br_b^{(t)} &~=~ \Shrink \Big( \cos\big( \frac{\pi b}{L} \big) \Bu^{(t-1)} \BDnT + \sin\big( \frac{\pi b}{L} \big) \BDm \Bu^{(t-1)}
    - \frac{\boldsymbol{\lambda}_{\boldsymbol{1} b}^{(t-1)}}{\beta_1} \,, \frac{1}{\beta_1} \Big)
    \,,~ \quad b = 0 \,, \ldots \,, L-1
    \\
    \mathbf{w}_a^{(t)} &~=~
    \Shrink \Big( \mathbf{t}_{\mathbf{w}_a} ~:=~ \mathbf{g}_a^{(t-1)} -
    \frac{ \boldsymbol{\lambda}_{\boldsymbol{2}a}^{(t-1)} }{\beta_2} \,, \frac{\mu_1}{\beta_2} \Big)
    \,,~ \quad a = 0 \,, \ldots \,, S-1
    \\
    \mathbf{g}_a^{(t)} &~=~ \RE \bigg[ \mathcal F^{-1} \Big\{ \mathcal A^{(t)}(\Bz) \cdot \mathcal B^{(t)}(\Bz) \Big\} \bigg]
    \,,~ a = 0 \,, \ldots \,, S-1
    \\
    \Bv^{(t)} &= \Shrink \bigg( \mathbf{t_v} ~:=~ \frac{\beta_3}{\beta_3 + \beta_4}
    \bigg( \sum_{s=0}^{S-1} \Big[ \cos\big( \frac{\pi s}{S} \big) \Bg_s^{(t)} \BDnT
    + \sin\big( \frac{\pi s}{S} \big) \BDm \Bg_s^{(t)} \Big] - \frac{\boldsymbol{\lambda}_{\boldsymbol 3}^{(t-1)}}{\beta_3} \bigg)
    \\& \qquad \qquad \qquad \qquad
    + \frac{\beta_4}{\beta_3 + \beta_4} \bigg( \Bf - \Bu^{(t-1)} - \Beps^{(t-1)} + \frac{\boldsymbol{\lambda}_{\boldsymbol 4}^{(t-1)}}{\beta_4} \bigg)
    \,,~ \frac{\mu_2}{\beta_3 + \beta_4}
    \bigg)   
    \\
    \Bu^{(t)} &~=~ \RE \bigg[ \mathcal F^{-1} \Big\{ \mathcal X^{(t)}(\Bz) \cdot \mathcal Y^{(t)}(\Bz) \Big\} \bigg]
    \\
    \Beps^{(t)} &~=~ \Big( \Bf - \Bu^{(t)} - \Bv^{(t)} + \frac{\boldsymbol{\lambda}_{\boldsymbol 4}^{(t-1)}}{\beta_4} \Big) ~-~
    \CST \big( \Bf - \Bu^{(t)} - \Bv^{(t)} + \frac{\boldsymbol{\lambda}_{\boldsymbol 4}^{(t-1)}}{\beta_4} \,, \delta \big)
   \end{align*} 
  {\bfseries 2. Update}
  $\Big( \big[\boldsymbol{\lambda}_{\mathbf{1}b}^{(t)}\big]_{b=0}^{L-1} \,, \big[\boldsymbol{\lambda}_{\mathbf{2}a}^{(t)}\big]_{a=0}^{S-1}
  \,, \boldsymbol{\lambda}_{\boldsymbol 3}^{(t)} \,, \boldsymbol{\lambda}_{\boldsymbol 4}^{(t)} \Big) \in X^{L+S+2}$:
  \begin{align*}
   \boldsymbol{\lambda}_{\mathbf{1}b}^{(t)} &~=~ \boldsymbol{\lambda}_{\mathbf{1}b}^{(t-1)}
   ~+~ \gamma \beta_1 \Big( \mathbf{r}_b^{(t)} - \cos\big( \frac{\pi b}{L} \big) \Bu^{(t)} \BDnT - \sin\big( \frac{\pi b}{L} \big) \BDm \Bu^{(t)}  \Big) 
   \,, \quad b = 0 \,, \ldots \,, L-1
   \\
   \boldsymbol{\lambda}_{\mathbf{2} a}^{(t)} &~=~ \boldsymbol{\lambda}_{\mathbf{2} a}^{(t-1)}
   ~+~ \gamma \beta_2 \Big( \mathbf{w}_a^{(t)} - \mathbf{g}_a^{(t)} \Big) 
   \,, \quad a = 0 \,, \ldots \,, S-1
   \\
   \boldsymbol{\lambda}_{\mathbf{3}}^{(t)} &~=~ \boldsymbol{\lambda}_{\mathbf{3}}^{(t-1)}
   ~+~ \gamma \beta_3 \Big( \Bv^{(t)} - \sum_{s=0}^{S-1} \big[ \cos\big(\frac{\pi s}{S} \big) \mathbf{g}_s^{(t)} \BDnT
   + \sin\big( \frac{\pi s}{S} \big) \BDm \mathbf{g}_s^{(t)} \big] \Big) 
   \\
   \boldsymbol{\lambda}_{\mathbf{4}}^{(t)} &~=~ \boldsymbol{\lambda}_{\mathbf{4}}^{(t-1)}
   ~+~ \gamma \beta_4 \big( \Bf - \Bu^{(t)} - \Bv^{(t)} - \Beps^{(t)} \big) 
  \end{align*}
 }
\ENDFOR
\end{algorithmic}
\end{algorithm}

\begin{algorithm}
\label{alg:DG3PD_Part2}
\begin{algorithmic}
\STATE
\begin{align*}
 &\mathcal A(\Bz) ~=~ \Bigg[ \beta_2 \mathbf{1_{mn}} + \beta_3 \Big[ \sin \frac{\pi a}{S} (\BzoI - 1) + \cos \frac{\pi a}{S} (\BztI - 1) \Big]
 \Big[ \sin \frac{\pi a}{S} (\Bzo - 1) + \cos \frac{\pi a}{S} (\Bzt - 1) \Big]
 \Bigg]^{-1}
 \,,
 \\
 &\mathcal B(\Bz) ~=~
 \beta_2 \Big[ W_a(\Bz)  + \frac{\Lambda_{2a}(\Bz) }{\beta_2} \Big]
 ~+~ \beta_3 \Big[ \sin\big( \frac{\pi a}{S} \big) (\BzoI -1) + \cos\big( \frac{\pi a}{S} \big) (\BztI-1) \Big] \times
 \\&
 \bigg[ V(\Bz) - \sum_{s=[0\,,S-1] \backslash \{a\} }
 \Big[ \cos\big( \frac{\pi s}{S} \big) (\Bzt-1)
 + \sin\big( \frac{\pi s}{S} \big) (\Bzo-1) \Big] G_s(\Bz)
 + \frac{\Lambda_3(\Bz)}{\beta_3} \bigg] \,,
 \\
 &\mathcal X(\Bz) =
 \Bigg[ \beta_4 \mathbf{1_{mn}} + \beta_1 \sum_{l=0}^{L-1} \Big[ \sin\big( \frac{\pi l}{L} \big) (\BzoI - 1) 
 + \cos\big( \frac{\pi l}{L} \big) (\BztI -1) \Big]
 \Big[ \sin\big( \frac{\pi l}{L} \big) (\Bzo -1) + \cos\big( \frac{\pi l}{L} \big) (\Bzt -1) \Big]
 \Bigg]^{-1} \,,
 \\
 &\mathcal Y(\Bz) =
 \beta_4 \Big[ F(\Bz) - V(\Bz) - \mathcal{E}(\Bz) + \frac{\Lambda_4(\Bz)}{\beta_4} \Big]
 + \beta_1 \sum_{l=0}^{L-1} \Big[ \sin \big( \frac{\pi l}{L} \big) (\BzoI-1) + \cos \big( \frac{\pi l}{L} \big) (\BztI -1) \Big]
 \Big[ R_l(\Bz) + \frac{\Lambda_{1l}(\Bz)}{\beta_1} \Big] .
\end{align*}
{\bfseries Choice of Parameters}
\begin{align*}
 \mu_1 &= c_{\mu_1} \beta_2 \cdot \max_{\Bk \in \Omega} \big( \abs{t_{\Bw_a}[\Bk]} \big) \,,~
 \mu_2 = c_{\mu_2} (\beta_3 + \beta_4) \cdot \max_{\Bk \in \Omega} \big( \abs{t_\Bv[\Bk]} \big)
 \text{  and  }
 \beta_2 = c_2 \beta_3 \,, \beta_3 = \frac{\theta}{1 - \theta} \beta_4 \,, \beta_1 = c_1 \beta_4.
\end{align*}
\end{algorithmic}
\end{algorithm}

\end{document}